\documentclass[10pt,twocolumn,letterpaper]{article}

\usepackage[pagenumbers]{cvpr} %
\usepackage{graphicx}
\usepackage{amsmath}
\usepackage{amssymb}
\usepackage{booktabs}
\usepackage[pagebackref,breaklinks,colorlinks]{hyperref}

\usepackage{times}
\usepackage{epsfig}
\usepackage{amsmath}
\usepackage{amssymb}
\usepackage{booktabs}
\usepackage{makecell}
\usepackage{multirow}
\usepackage{tabularx}
\usepackage{bbm}
\usepackage{caption} 
\usepackage{subcaption}
\usepackage{animate}
\usepackage{pifont}
\usepackage{adjustbox}
\usepackage[dvipsnames]{xcolor}
\usepackage{enumitem}
\makeatletter
\@namedef{ver@everyshi.sty}{}
\makeatother
\usepackage{pgfplots}
\makeatletter
\@namedef{ver@everyshi.sty}{}
\makeatother
\usepackage{tikz}
\usetikzlibrary{spy}

\newcommand{\ours}{vMAP}
\newcolumntype{L}[1]{>{\raggedright\let\newline\\\arraybackslash\hspace{0pt}}m{#1}}
\newcolumntype{C}[1]{>{\centering\let\newline\\\arraybackslash\hspace{0pt}}m{#1}}
\newcolumntype{R}[1]{>{\raggedleft\let\newline\\\arraybackslash\hspace{0pt}}m{#1}}
\newcolumntype{Y}{>{\centering\arraybackslash}X}

\makeatletter
\renewcommand{\paragraph}{%
  \@startsection{paragraph}{4}%
  {\z@}{1.1ex \@plus 1ex \@minus .2ex}{-1em}%
  {\normalfont\normalsize\bfseries}%
}
\makeatother

\def\cvprPaperID{6807} %
\def\confName{CVPR}
\def\confYear{2023}

\begin{document}

\title{vMAP: Vectorised Object Mapping for Neural Field SLAM}

\author{Xin Kong\quad Shikun Liu\quad Marwan Taher \quad Andrew J. Davison \\
Dyson Robotics Lab, Imperial College London\\
{\tt\small \{x.kong21, shikun.liu17, m.taher, a.davison\}@imperial.ac.uk}
}

\twocolumn[{%
\renewcommand\twocolumn[1][]{#1}%
\maketitle
\begin{minipage}[b]{.7\textwidth}
\includegraphics[width=\textwidth]{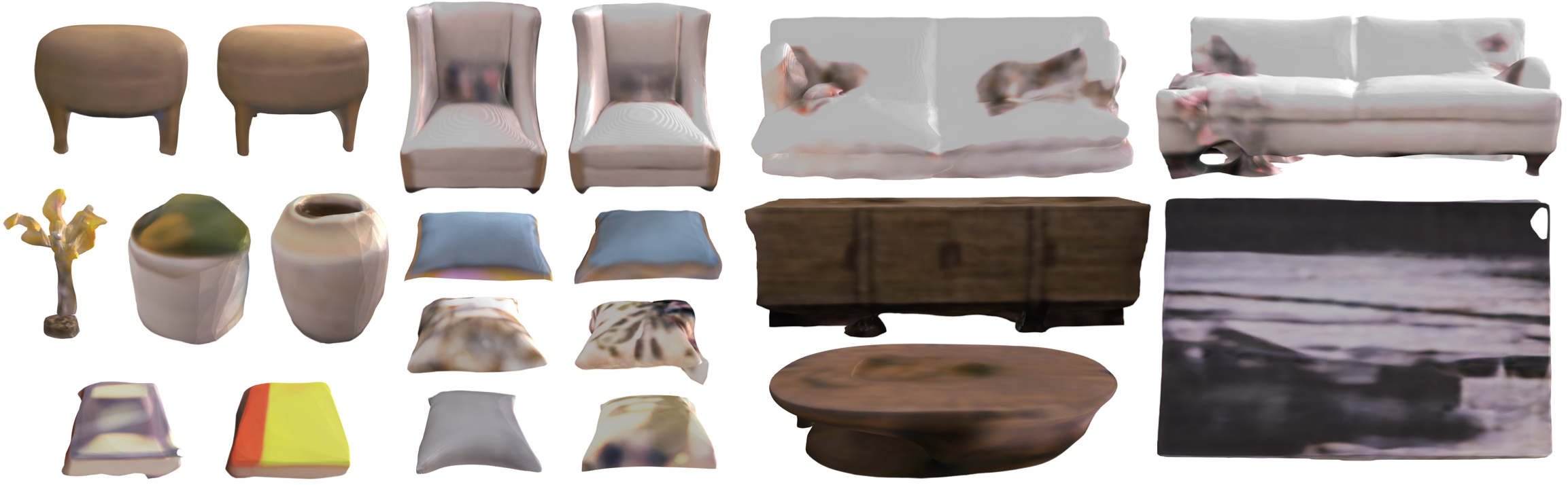}
\end{minipage}\hfill
\begin{minipage}[b]{.29\textwidth}
\centering
\includegraphics[width=\textwidth]{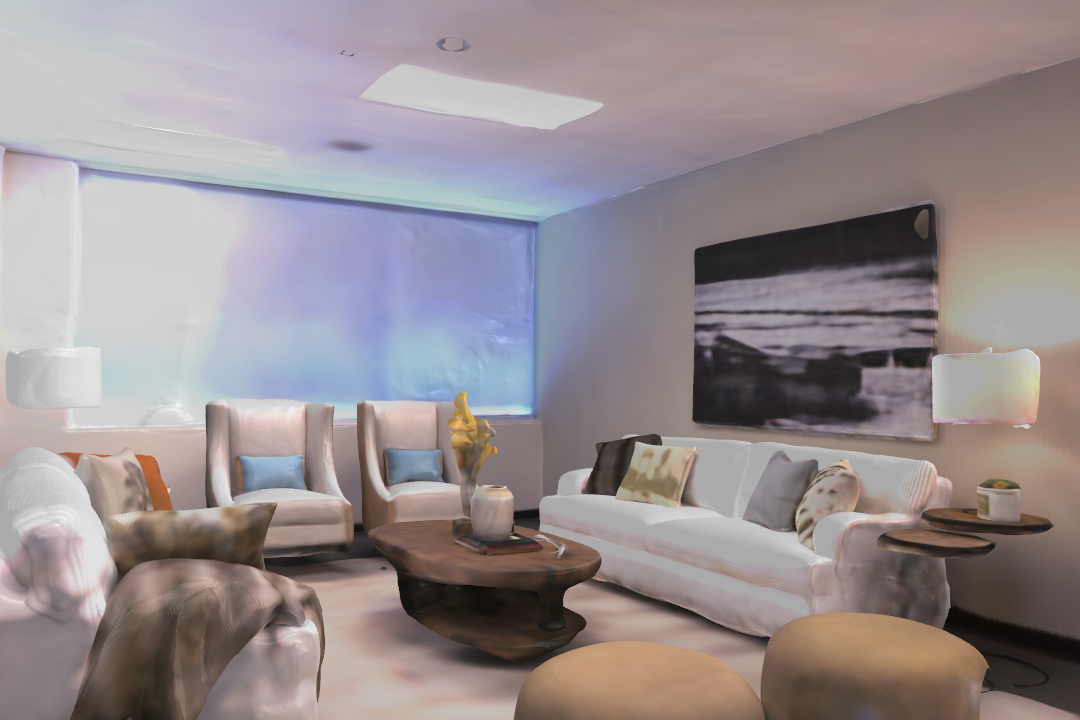}
\vspace{-2mm}
\end{minipage}
\captionof{figure}{vMAP automatically builds an object-level scene model from a real-time RGB-D input stream. Each object is represented by a separate MLP neural field model, all optimised in parallel via vectorised training. We use no 3D shape priors, but the MLP representation encourages object reconstruction to be watertight and complete, even when objects are partially observed or are heavily occluded in the input images. See for instance the separate reconstructions of the armchairs, sofas and cushions, which were mutually occluding each other, in this example from Replica.
}
\vspace{3mm}
}]

\begin{abstract}
We present \ours{}, an object-level dense SLAM system using neural field representations. Each object is represented by a small MLP, enabling efficient, watertight object modelling without the need for 3D priors.

As an RGB-D camera browses a scene with no prior information, \ours{} detects object instances on-the-fly, and dynamically adds them to its map. Specifically, thanks to the power of vectorised training, \ours{} can optimise as many as 50 individual objects in a single scene, with an extremely efficient training speed of 5Hz map update. We experimentally demonstrate significantly improved scene-level and object-level reconstruction quality compared to prior neural field SLAM systems. Project page: \url{https://kxhit.github.io/vMAP}.
\end{abstract}

\vspace{-4mm}
\section{Introduction}
For robotics and other interactive vision applications, an object-level model is arguably semantically optimal, with scene entities represented in a separated, composable way, but also efficiently focusing resources on what is important in an environment.

The key question in building an object-level mapping system is what level of prior information is known about the objects in a scene in order to segment, classify and reconstruct them. 
If no 3D object priors are available, 
then usually only the directly observed parts of objects can be reconstructed, leading to holes and missing parts\cite{zhou2013dense, dai2017bundlefusion}. 
Prior object information such as CAD models or category-level shape space models enable full object shape estimation from partial views, but only for the subset of objects in a scene for which these models are available.

In this paper, we present a new approach which applies to the case where no 3D priors are available but still often enables watertight object reconstruction in realistic real-time scene scanning.
Our system, \ours{}, builds on the attractive properties shown  by neural fields as a real-time scene representation~\cite{Sucar:etal:ICCV2021}, with efficient and complete representation of shape,
but now reconstructs a separate tiny MLP model of each object. The key technical contribution of our work is to show that a large number of separate MLP object models can be simultaneously and efficiently optimised on a single GPU during live operation via vectorised training. 

We show that we can achieve much more accurate and complete scene reconstruction by separately modelling objects, compared with using a similar number of weights in a single neural field model of the whole scene. Our real-time system is highly efficient in terms of both computation and memory, and we show that scenes with up to 50 objects can be mapped with 40KB per object of learned parameters across the multiple, independent object networks.

We also demonstrate the flexibility of our disentangled object representation to enable 
recomposition of scenes with new object configurations. Extensive experiments have been conducted on both simulated and real-world datasets,  showing state-of-the-art scene-level and object-level reconstruction performance.

\section{Related Work}
This work follows in long series of efforts to build real-time scene representations which are decomposed into explicit rigid objects, with the promise of flexible and efficient scene representation and even the possibility to represent changing scenes. Different systems assumed varying types of representation and levels of prior knowledge, from CAD models \cite{Salas-Moreno:etal:CVPR2013}, via category-level shape models \cite{Sucar:etal:3DV2020, li2022generative, wang2021dsp, kong2020semantic} to no prior shape knowledge, although in this case only the visible parts of objects could be reconstructed \cite{Runz::Agapito::ICRA2017,McCormac:etal:3DV2018,Xu:etal:ICRA2019}.

\paragraph{Neural Field SLAM} Neural fields have recently been widely used as efficient, accurate and flexible representations of whole scenes
\cite{Mildenhall:etal:ECCV2020, muller2022instant, mescheder2019occupancy, park2019deepsdf}.
To adopt these representations into real-time SLAM systems, iMAP~\cite{Sucar:etal:ICCV2021} demonstrated for the first time that a simple MLP network, incrementally trained with the aid of depth measurements from RGB-D sensors, can represent room-scaled 3D scenes in real-time. 
Some of iMAP's most interesting properties were its tendency to produce watertight reconstructions, even often plausibly completing the unobserved back of objects.
These coherence properties of neural fields were particularly revealed when semantic output channels were added, as in  SemanticNeRF\cite{Zhi:etal:CVPR2019} and iLabel \cite{zhi2022ilabel}, and were found to inherit the coherence. 
To make implicit representation more scalable and efficient, a group of implicit SLAM systems~\cite{zhu2022nice, wang2022go, yang2022voxfusion, zhong2023icra, rosinol2022nerf} fused neural fields with conventional volumetric representations. 

\paragraph{Object Representations with Neural Fields} However, obtaining individual object representations from these neural field methods is difficult, as the correspondences between network parameters and specific scene regions are complicated and difficult to determine. To tackle this, DeRF~\cite{rebain2021derf} decomposed a scene spatially and dedicated smaller networks to each decomposed part. Similarly, KiloNeRF~\cite{reiser2021kilonerf} divided a scene into thousands of volumetric parts, each represented by a tiny MLP, and trained them in parallel with custom CUDA kernels to speed up NeRF. Different from KiloNeRF, \ours{} decomposes the scene into objects which are semantically meaningful. 

To represent multiple objects, ObjectNeRF~\cite{Yang:etal:ICCV2021} and ObjSDF~\cite{wu2022object} took pre-computed instance masks as additional input and conditioned object representation on learnable object activation code. But these methods are still trained offline and tangle object representations with the main scene network, so that they need to optimise the network weights with all object codes during training,  and infer the whole network to get the shape of a desired object. This contrasts with \ours{} which models objects individually, and is able to stop and resume training for any objects without any inter-object interference.

The recent work most similar to ours has used the attractive properties of neural field MLPs to represent single objects.
The analysis in \cite{Davies:etal:ICML2021} explicitly evaluated the use of over-fit neural implicit networks as a 3D shape representation for graphics, considering that they should be taken seriously.
The work in \cite{AbouChakra:etal:ARXIV2022} furthered this analysis, showing how object representation was affected by different observation conditions, though using the hybrid Instant NGP rather than a single MLP representation, so it is not clear whether some object coherence properties would be lost.
Finally, the CodeNeRF system\cite{Jang:etal:ICCV2021} trained a NeRF conditioned on learnable object codes, again proving the attractive properties of neural fields to represent single objects. 

We build on this work in our paper, but for the first time show that many individual neural field models making up a whole scene can be simultaneously trained within a real-time system, resulting in accurate and efficient representation of many-object scenes.

\section{\ours{}: An Efficient Object Mapping System with Vectorised Training}
\begin{figure*}[!h]
   \centering 
   \includegraphics[width=\linewidth]{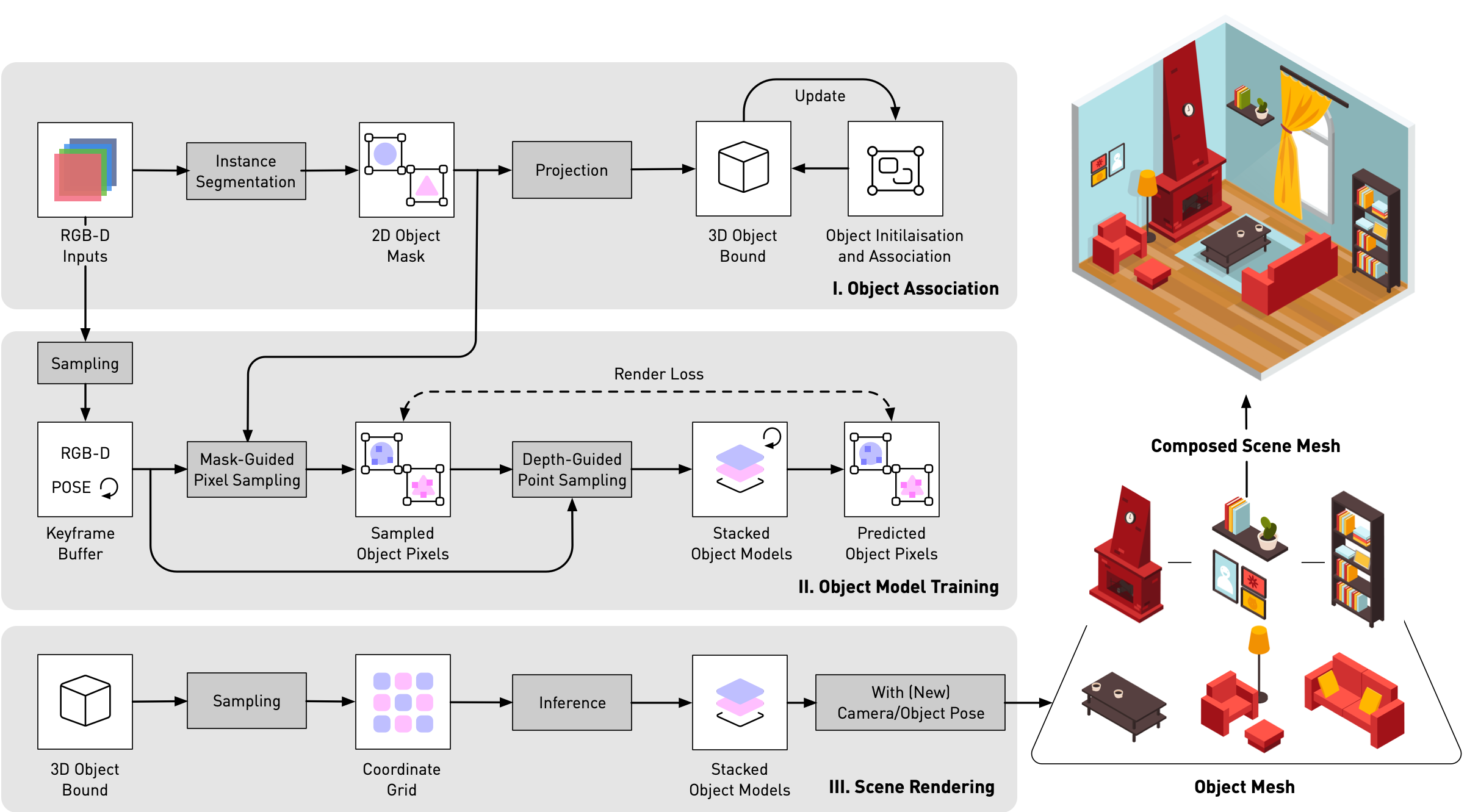}
\caption{An overview of training and rendering pipeline of \ours{}.}
\label{fig:framework}
\vspace{-4mm}
\end{figure*}

\subsection{System Overview}
We first introduce our detailed design for object-level mapping with efficient vectorised training (Section \ref{subsec:objectmapping}), and then explain our improved training strategies of pixel sampling and surface rendering (Section \ref{subsec:representation}). Finally, we show how we may recompose and render a new scene with these learned object models (Section \ref{subsec:compositionalrendering}). An overview of our training and rendering pipeline is shown in Fig.~\ref{fig:framework}.

\subsection{Vectorised Object Level Mapping}
\label{subsec:objectmapping}

\paragraph{Object Initialisation and Association}  To start with, each frame is associated with densely labelled object masks. These object masks are either directly provided in the dataset, or predicted with an off-the-shelf 2D instance segmentation network.   Since those predicted object masks have no temporal consistency across different frames, we perform object association between the previous and the current live frame, based on two criteria: i) {\it Semantic Consistency}: the object in the current frame is predicted as the same semantic class from the previous frame, and ii) {\it Spatial Consistency}: the object in the current frame is spatially close to the object in the previous frames, measured by the mean IoU of their 3D object bounds. When these two criteria are satisfied, we assume they are the same object instance and represent them with the same object model. Otherwise, they are different object instances and we initialise a new object model and append it to the models stack.

For each object in a frame, we estimate its 3D object bound by its 3D point cloud, parameterised by its depth map and the camera pose. 
Camera tracking is externally provided by an off-the-shelf tracking system, which we found to be more accurate and robust 
compared to jointly optimising pose and geometry. If we detect the same object instance in a new frame, we merge its 3D point cloud from the previous frames to the current frame and re-estimate its 3D object bound. Therefore, these object bounds are dynamically updated and refined with more observations.

\paragraph{Object Supervision} We apply object-level supervision only for pixels inside a 2D object bounding box, for maximal training efficiency. For those pixels within an object mask, we encourage the object radiance field to be occupied and supervise them with depth and colour loss. Otherwise we encourage the object radiance field to be empty.

Each object instance samples training pixels from its own independent keyframe buffer. Therefore, we have flexibility to stop or resume the training of any object, with no training interference between objects.

\paragraph{Vectorised Training} Representing a neural field with multiple small networks can lead to efficient training, as shown in prior work~\cite{reiser2021kilonerf}. In \ours{}, all object models are of the same design, except for the background object which we represent with a slightly larger network. Therefore, we are able to stack these small object models together for vectorised training, leveraging the highly optimised vectorised operations in PyTorch~\cite{functorch2021}. Since multiple object models are batched and trained simultaneously as opposed to sequentially, we optimise the use of the available GPU resources. We show that vectorised training is an essential design element to the system, resulting in significantly improved training speed, further discussed in Section \ref{subsec: performance}.

\subsection{Neural Implicit Mapping}
\label{subsec:representation}
\paragraph{Depth Guided Sampling} Neural fields trained on RGB data only have no guarantee to model accurate object geometry, due to the fact that they are optimising for appearance rather than the geometry. To obtain more geometrically accurate object models, we benefit from the depth map available from an RGB-D sensor, providing a strong prior for learning the density field of 3D volumes. Specifically, we sample $N_s$ and $N_c$ points along each ray, for which $N_s$ points are sampled with a Normal distribution centered around the surface $t_s$ (from the depth map), with a small $d_\sigma$ variance, and $N_c$ points are uniformly sampled between the camera $t_n$ (the near bound) and the surface $t_s$, with a stratified sampling approach. When the depth measurement is invalid, the surface $t_s$ is then replaced with the far bound $t_f$.  Mathematically, we have:
\begin{align}
t_i &\sim \mathcal{U}\left(t_n+\frac{i-1}{N_c}\left(t_s-t_n\right), t_n+\frac{i}{N_c}\left(t_s-t_n\right)\right),  \label{equ:sampling}\\ 
t_i &\sim \mathcal{N}(t_s, d_{\sigma}^{2}) \label{equ:surface_sampling}
~.
\end{align}
We choose $d_\sigma=3cm$ which works well in our implementation. We observe that training more points near the surface helps to guide the object models to quickly focus on representing accurate object geometry.

\paragraph{Surface and Volume Rendering} As we are concerned more by 3D surface reconstruction than 2D rendering, we omit the viewing direction from the network input, and model object visibility with a binary indicator (no transparent objects). With similar motivation to UniSURF~\cite{oechsle2021unisurf}, we parameterise the occupancy probability of a 3D point $x_i$ as $o_\theta\left(x_i\right)\to [0, 1]$, where $o_\theta$ is a continuous occupancy field.  Therefore, the termination probability at point $x_i$ along ray $\mathbf{r}$ becomes $T_i=o\left(x_i\right) \prod_{j<i}\left(1-o\left(x_j\right)\right)$, indicating that no occupied samples $x_j$ with $j<i$ exist before $x_i$. The corresponding rendered occupancy, depth and colour are defined as follows:
\begin{equation}\label{equ:depth_color_rendering}
\hat{O}(\mathbf{r})=\sum_{i=1}^N T_i, \, \hat{D}(\mathbf{r})=\sum_{i=1}^N T_i d_i, \, \hat{C}(\mathbf{r})=\sum_{i=1}^N T_i c_i
~.
\end{equation}

\paragraph{Training Objective}
For each object $k$, we only sample training pixels inside that object's 2D bounding box, denoted by $\mathcal{R}^k$, and only optimise depth and colour for pixels inside its 2D object mask, denoted by $M^k$. Note that it is always true that $M^k \subset \mathcal{R}^k$. The depth, colour and occupancy loss for the object $k$ are defined as follows:
\begin{align}\label{equ:depth_color_loss}
L^{k}_{depth}&=M^k\odot \sum_{\mathbf{r}\in{R}^{k}}|\hat{D}(\mathbf{r})-D(\mathbf{r})|, \\ L^{k}_{colour}&=M^k \odot \sum_{\mathbf{r}\in{R}^{k}}|\hat{C}(\mathbf{r})-C(\mathbf{r})|, \\
L^{k}_{occupancy}&=\sum_{\mathbf{r}\in{R}^{k}}|\hat{O}(\mathbf{r})-M^k(\mathbf{r})|
~.
\end{align}
The overall training objective then accumulates losses for all $K$ objects: \begin{align}\label{equ:all_loss}
L = \sum_{k=1}^K L^k_{depth}+\lambda_{1} \cdot L^k_{colour} +\lambda_{2} \cdot L^k_{occupancy}
~.
\end{align}
We choose loss weightings $\lambda_1=5$ and $\lambda_2=10$, which we found to work well in our experiments.

\subsection{Compositional Scene Rendering}
\label{subsec:compositionalrendering}
Since \ours{} represents objects in a purely disentangled representation space, we can obtain each 3D object by querying within its estimated 3D object bounds and easily manipulate it. For 2D novel view synthesis, we use the Ray-Box Intersection algorithm~\cite{majercik2018ray} to calculate near and far bounds for each object, and then rank rendered depths along each ray to achieve occlusion-aware scene-level rendering. This disentangled representation also opens up other types of fine-grained object-level manipulation, such as changing object shape or textures by conditioning on disentangled pre-trained feature fields~\cite{niemeyer2021giraffe, yuan2022nerf}, which we consider as an interesting future direction.

\vspace{-1mm}
\section{Experiments} 
\label{sec:experiments}
We have comprehensively evaluated \ours{} on a range of different datasets, which include both simulated and real-world sequences, with and without ground-truth object masks and poses.  For all datasets, we qualitatively compare our system to prior state-of-the-art SLAM frameworks on 2D and 3D scene-level and object-level rendering. We further quantitatively compare these systems in datasets where ground-truth meshes are available. Please see our attached supplementary material for more results.

\subsection{Experimental Setup}

\paragraph{Datasets}
We evaluated on Replica~\cite{straub2019replica}, ScanNet~\cite{dai2017scannet}, and TUM RGB-D \cite{Endres:etal:ICRA2012}. Each dataset contains sequences with different levels of quality in object masks, depth and pose measurements. Additionally, we also showed \ours{}'s performance in complex real-world with self-captured video sequences recorded by an Azure Kinect RGB-D camera. An overview of these datasets is shown in Tab.~\ref{tab:dataset}. 

\begin{table}[ht!]
    \setlength{\tabcolsep}{0.1em}
    \centering
    \footnotesize
      \begin{tabular}{L{0.24\linewidth}C{0.24\linewidth}C{0.24\linewidth}C{0.24\linewidth}}
      \toprule
         & \makecell[b]{Object Masks} & Depth Quality & Pose Estimation \\
     \midrule
     Replica & Perfect GT & Perfect GT & Perfect GT \\
     ScanNet & Noisy & Noisy  & Perfect GT\\
     TUM RGB-D & Detic & Noisy & ORB-SLAM3 \\
     Our Recording & Detic & Noisy & ORB-SLAM3 \\
      \bottomrule 
      \end{tabular}%
    \caption{An overview of datasets we evaluated.}
    \label{tab:dataset}
    \vspace{-4mm}
\end{table}
Datasets with perfect ground-truth information represent the upper-bound performance of our system. We expect \ours{}'s performance in the real-world setting can be further improved, when coupled with a better instance segmentation and pose estimation framework.

\begin{table*}[ht!]
  \renewcommand{\arraystretch}{0.92} 
    \setlength{\tabcolsep}{0.16em}
    \centering
    \footnotesize
      \begin{tabular}{L{0.23\linewidth}C{0.12\linewidth}C{0.12\linewidth}C{0.12\linewidth}C{0.12\linewidth}C{0.12\linewidth}C{0.12\linewidth}}
      \toprule
         & \makecell[b]{{\bf TSDF-Fusion$^\ast$}} & \makecell[b]{{\bf iMAP}} & \makecell[b]{{\bf iMAP$^\ast$} } & \makecell[b]{{\bf NICE-SLAM}} & \makecell[b]{{\bf NICE-SLAM$^\ast$}} & \makecell[b]{{\bf \ours{}}}  \\
     \midrule
     {\bf Scene Acc.} [cm] $\downarrow$  & {\bf 1.28} & 4.43 & 2.15 & 2.94 & 3.04 & 3.20\\
     {\bf Scene Comp.} [cm] $\downarrow$  & 5.61 & 5.56 & 2.88 & 4.02 & 3.84 & {\bf 2.39} \\
     {\bf Scene Comp. Ratio} [$<$5cm \%] $\uparrow$ & 82.67 & 79.06 & 90.85  & 86.73 & 86.52 & {\bf 92.99}\\
     \midrule
     {\bf Object Acc.} [cm] $\downarrow$  & {\bf 0.45} & - & 3.57 & - & 3.91 & 2.23\\
     {\bf Object Comp.} [cm] $\downarrow$  & 3.69 & - & 2.38 & - & 3.27 & {\bf 1.44} \\
     {\bf Object Comp. Ratio} [$<$5cm \%] $\uparrow$ & 82.98 & - & 90.19 & - & 83.97 & {\bf 94.55}\\
     {\bf Object Comp. Ratio} [$<$1cm \%] $\uparrow$ & 61.70 & - & 47.79 & - & 37.79 & {\bf 69.23}\\
    \bottomrule 
    \end{tabular}
    \vspace{-2mm}
    \caption{Averaged reconstruction results for 8 indoor Replica scenes. $\ast$ represents the baselines we re-trained with ground-truth pose.}
    \label{tab:replica_results}
    \vspace{-2mm}
\end{table*}

\begin{figure*}[ht!]
  \newcommand\makespy[1]{%
      \begin{tikzpicture}[spy using outlines={rectangle, magnification=4, height=1cm, width=2.48cm, every spy on node/.append style={line width=1.5}}]
      \centering
        \node {\includegraphics[trim={5cm 4cm 5cm 4cm}, clip, width=\linewidth]{#1}};
        \spy[color=Melon] on (-0.4, 0.85) in node[line width=1.5] at (1.26,-2.01);
        \spy[color=Lavender] on (-2.0,-0.4) in node[line width=1.5] at (-1.26,-2.01);
       \end{tikzpicture}
   }
 \newcommand\makespyb[1]{%
      \begin{tikzpicture}[spy using outlines={rectangle, magnification=4, height=1cm, width=1.6cm, every spy on node/.append style={line width=1.5}}]
        \centering
        \node {\includegraphics[trim={7cm 3cm 7cm 3cm}, clip, width=\linewidth]{#1}};
        \spy[color=Melon] on (0.9, -0.1) in node[line width=1.5] at (0.82,-2);
        \spy[color=Lavender] on (0.1,-0.4) in node[line width=1.5] at (-0.82,-2);
       \end{tikzpicture}
   }
    \newcommand\makespyc[1]{%
      \begin{tikzpicture}[spy using outlines={rectangle, magnification=4, height=1cm, width=1.6cm, every spy on node/.append style={line width=1.5}}]
        \centering
        \node {\includegraphics[trim={7cm 3cm 7cm 3cm}, clip, width=\linewidth]{#1}};
        \spy[color=Melon] on (0.0, 0.2) in node[line width=1.5] at (0.82,-2);
        \spy[color=Lavender] on (-0.4,-0.8) in node[line width=1.5] at (-0.82,-2);
       \end{tikzpicture}
   }
   \newcommand\makespyd[1]{%
      \begin{tikzpicture}[spy using outlines={rectangle, magnification=4, height=1cm, width=1.35cm, every spy on node/.append style={line width=1.5}}]
        \centering
        \node {\includegraphics[trim={5cm 0 5cm 0}, clip, width=\linewidth]{#1}};
        \spy[color=Melon] on (0.4, -0.0) in node[line width=1.5] at (0.7,-2);
        \spy[color=Lavender] on (-0.7, -0.8) in node[line width=1.5] at (-0.7,-2);
       \end{tikzpicture}
   }
  \centering
  \small
  \setlength{\tabcolsep}{0.5em}
  \renewcommand{\arraystretch}{0.4} 
  \begin{tabular}{>{\centering}m{0.08\textwidth} >{\centering}m{0.29\textwidth} >{\centering}m{0.19\textwidth} >{\centering}m{0.19\textwidth} >{\centering\arraybackslash}m{0.165\textwidth}}
    & {\tt room-1} & {\tt room-2} & {\tt office-0} & {\tt office-4} \\
    \makecell{TSDF \\ Fusion} &
    \makespy{figs/3D_topview/room1_tsdf.png} &
    \makespyb{figs/3D_topview/room2_tsdf.png} &
    \makespyc{figs/3D_topview/office0_tsdf.png} &
    \makespyd{figs/3D_topview/office4_tsdf.png} \\
    iMAP &
    \makespy{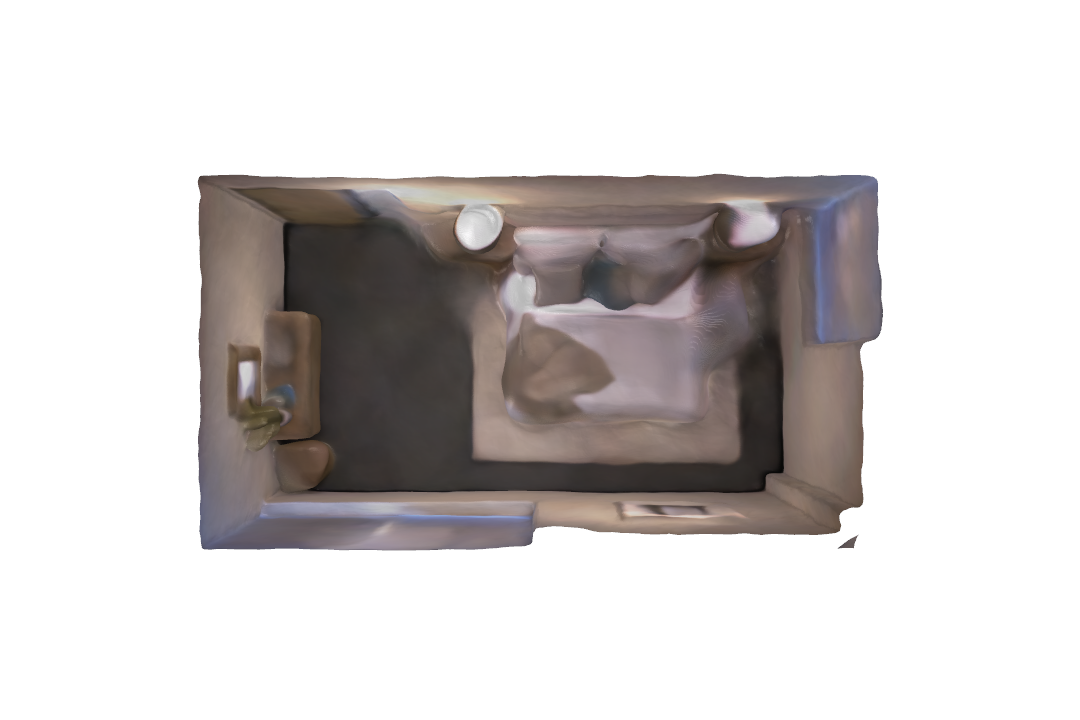} & 
    \makespyb{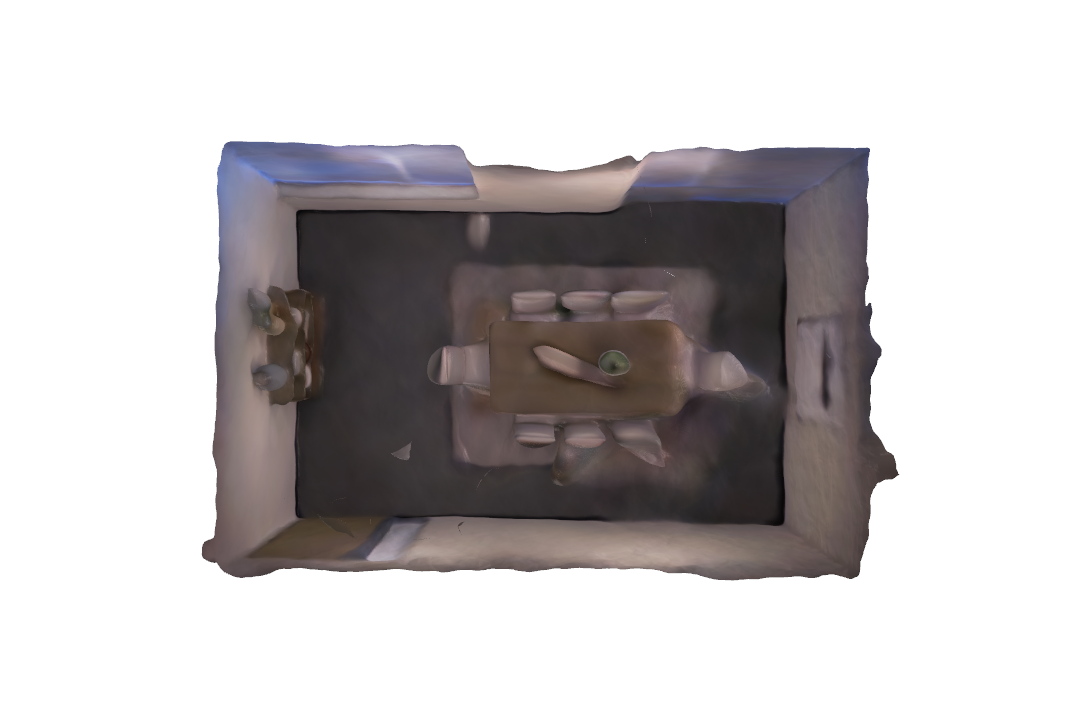} &
    \makespyc{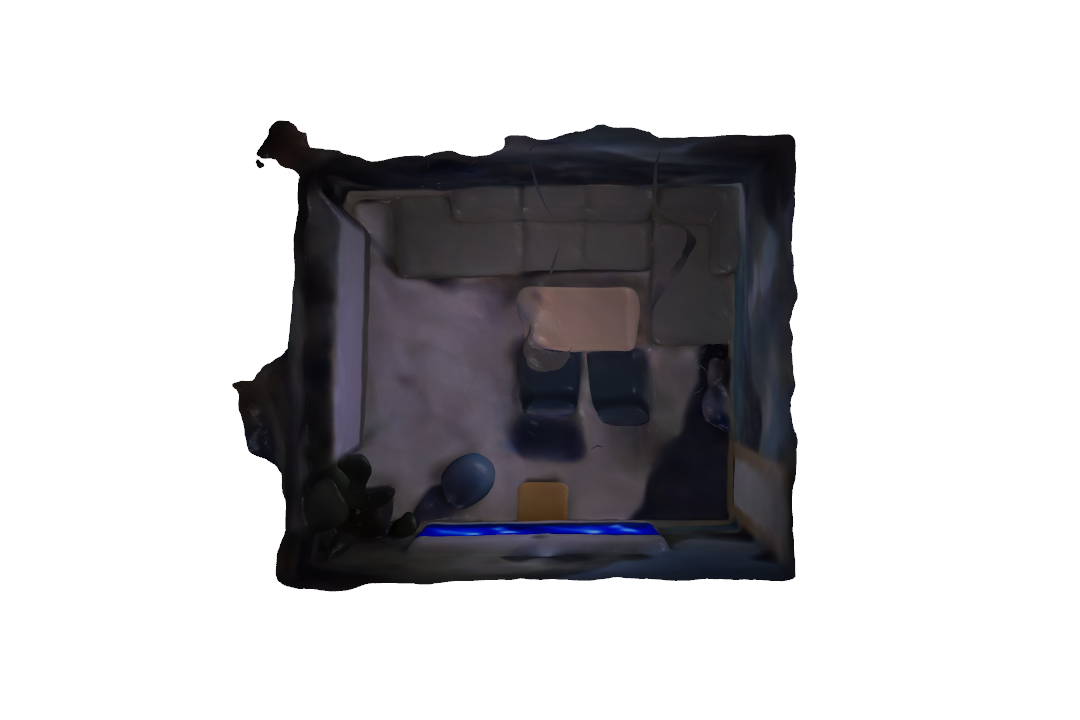} &
    \makespyd{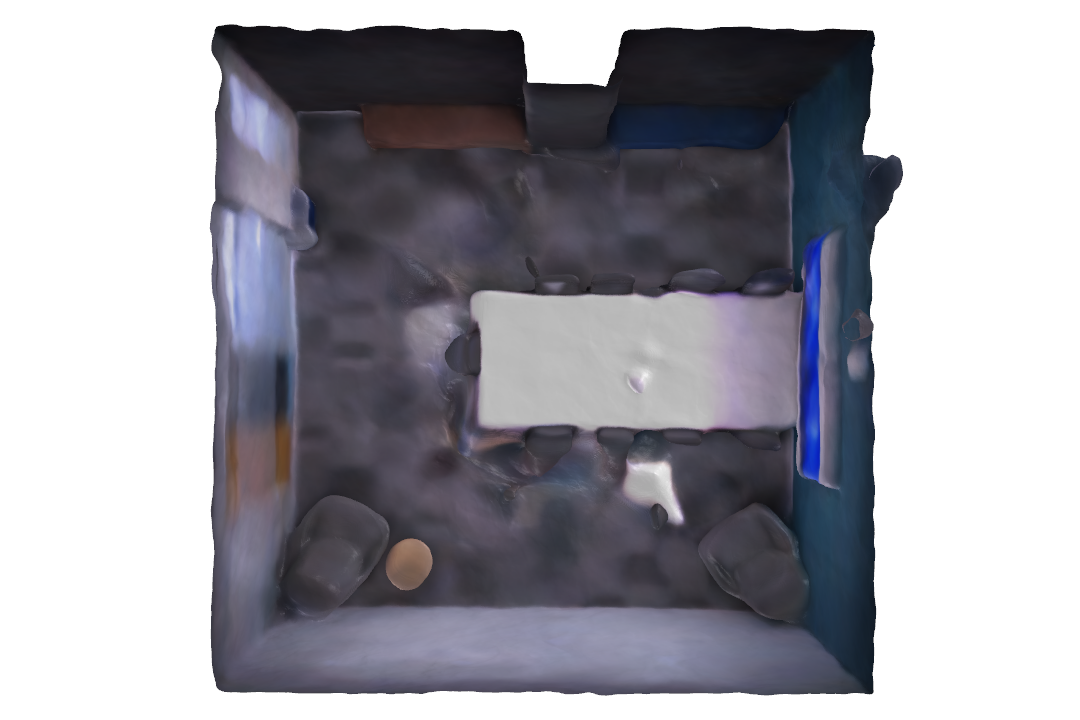} \\
     \makecell{NICE \\ SLAM} &
    \makespy{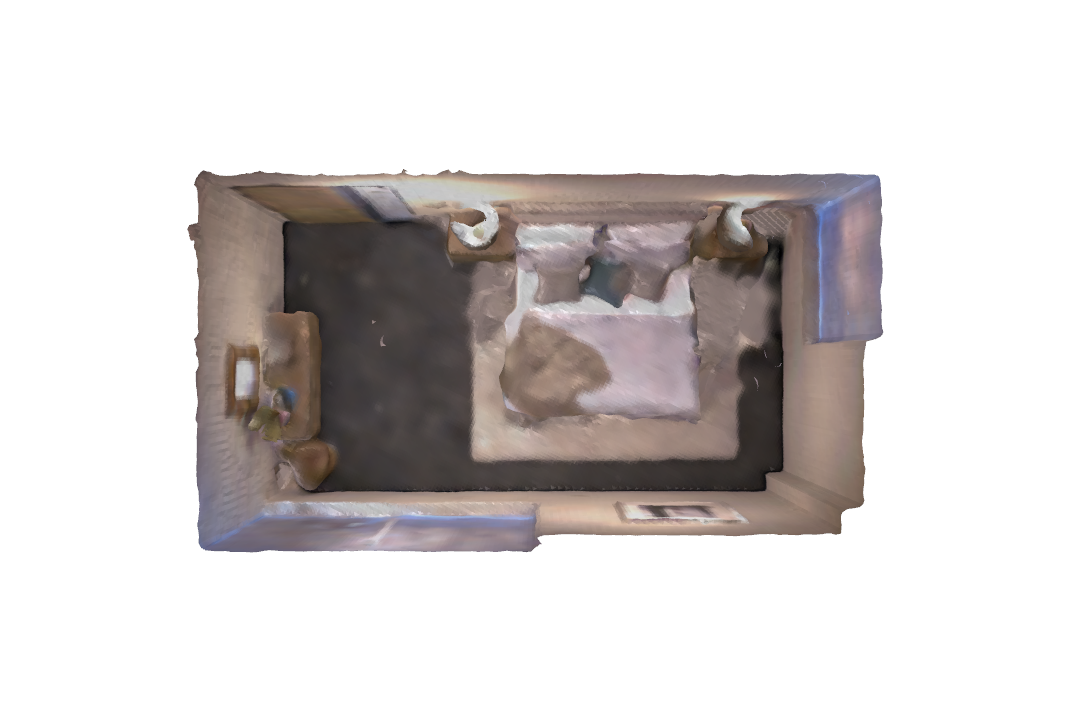} & 
    \makespyb{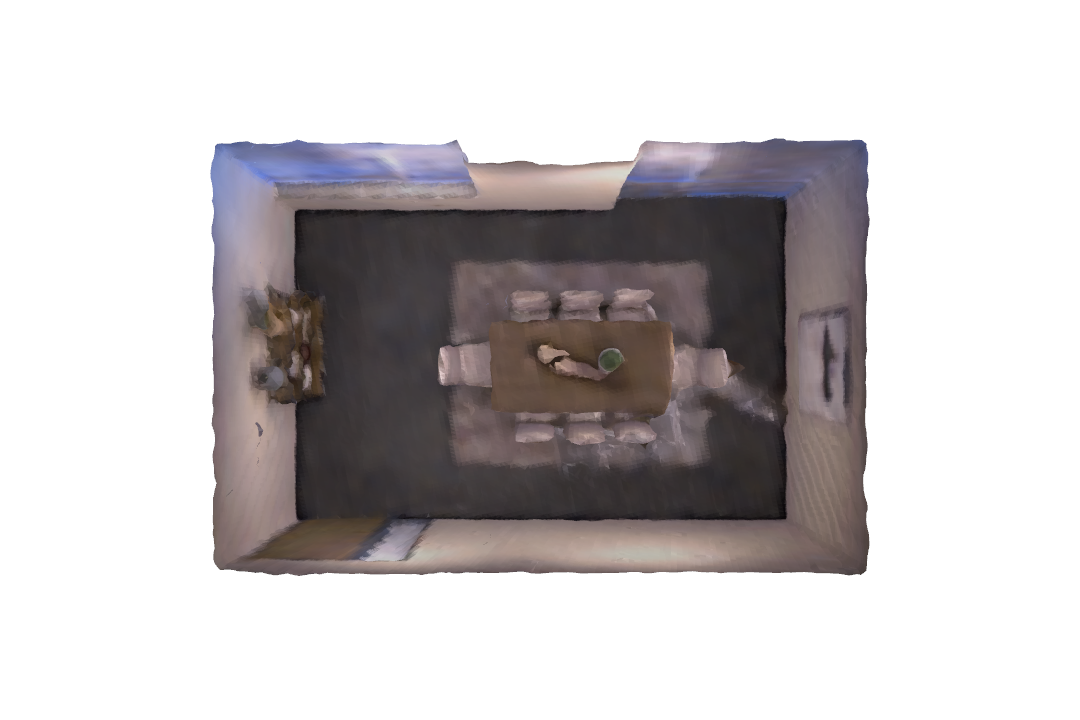} &
    \makespyc{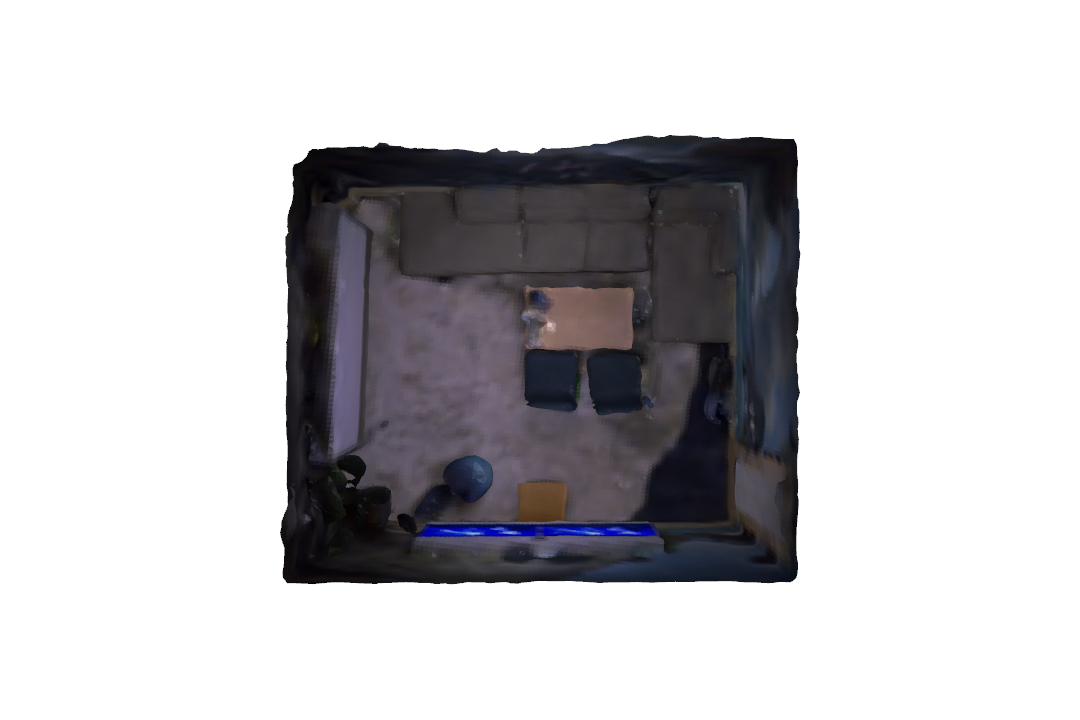} &
    \makespyd{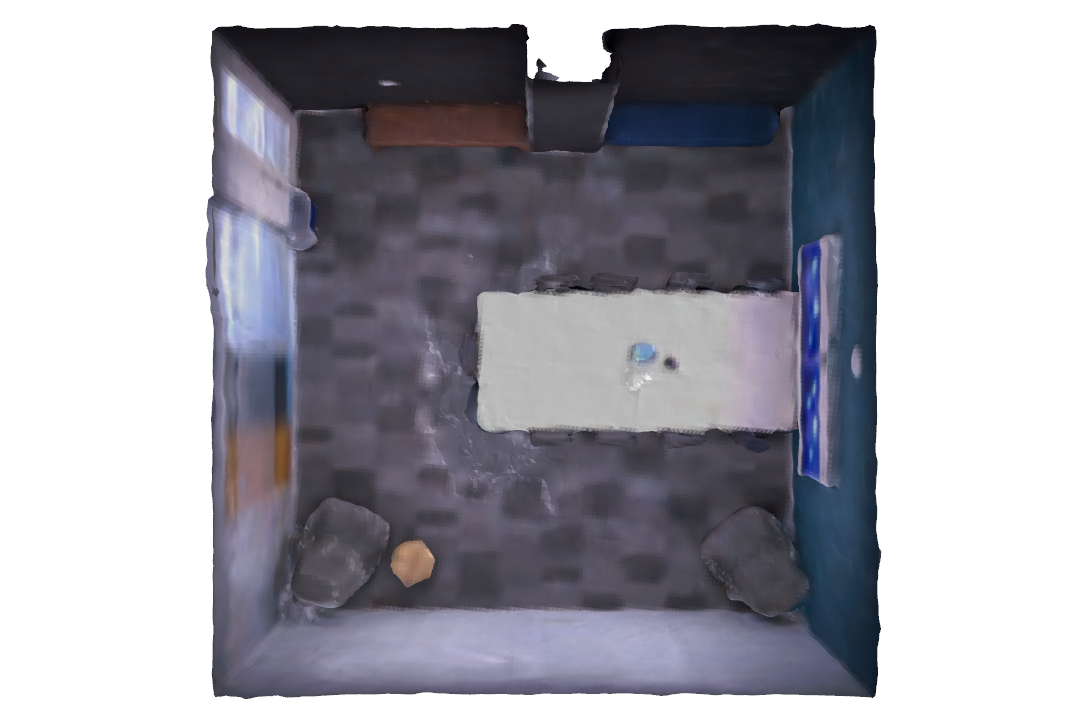} \\
    \ours{} &
    \makespy{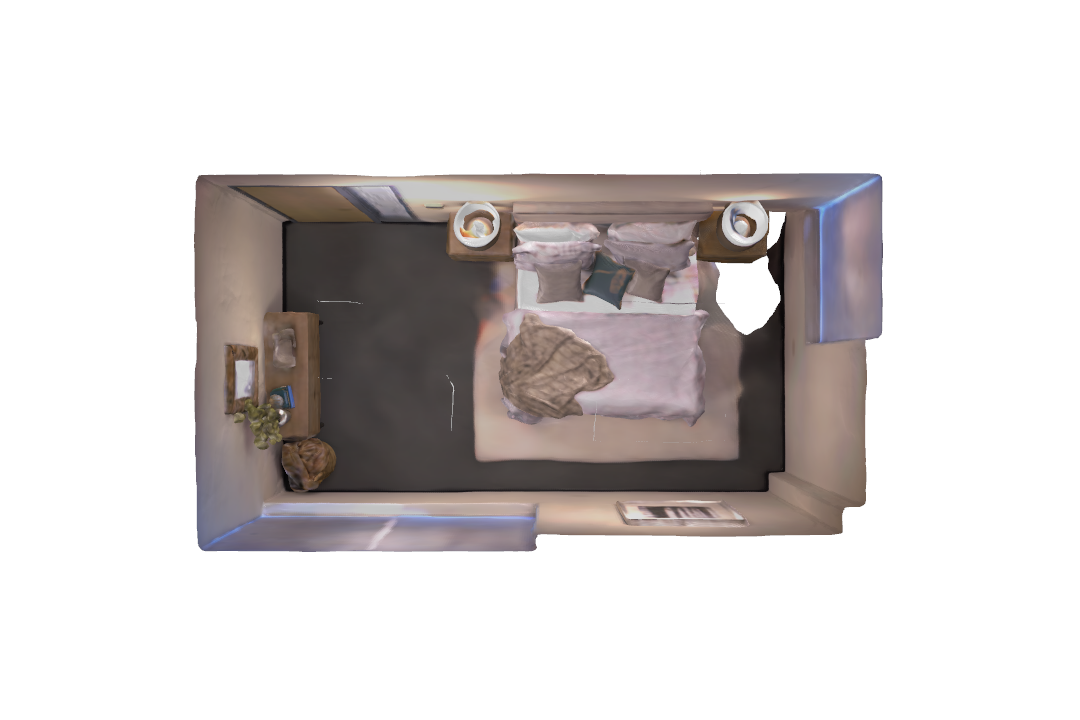} & 
    \makespyb{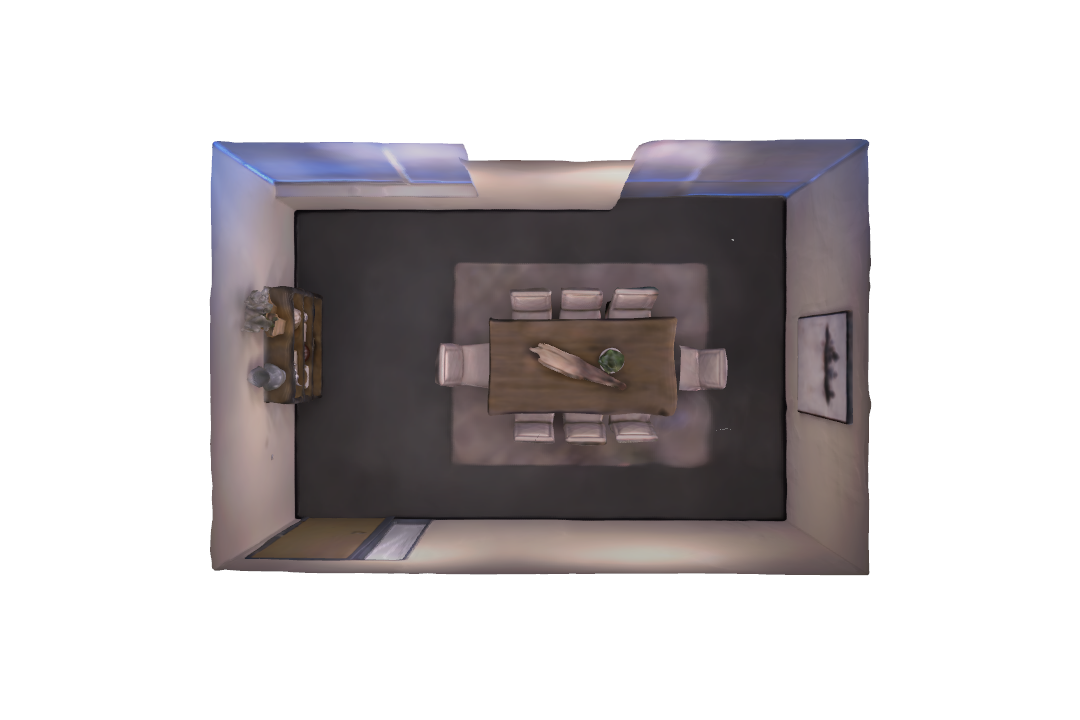} &
    \makespyc{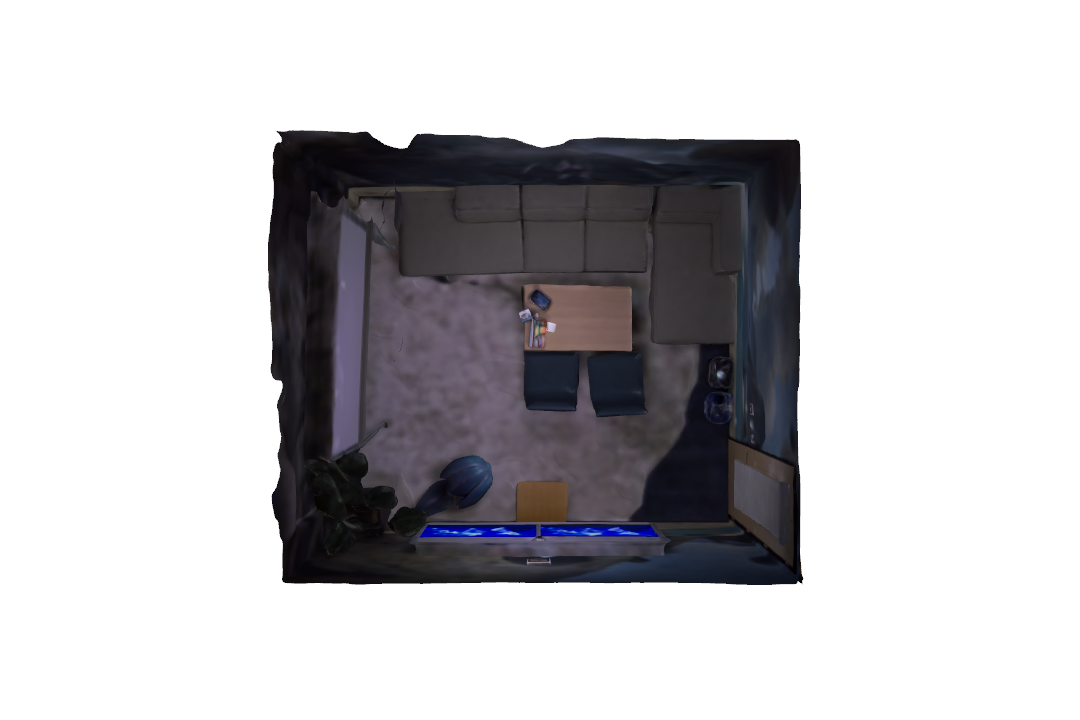} &
    \makespyd{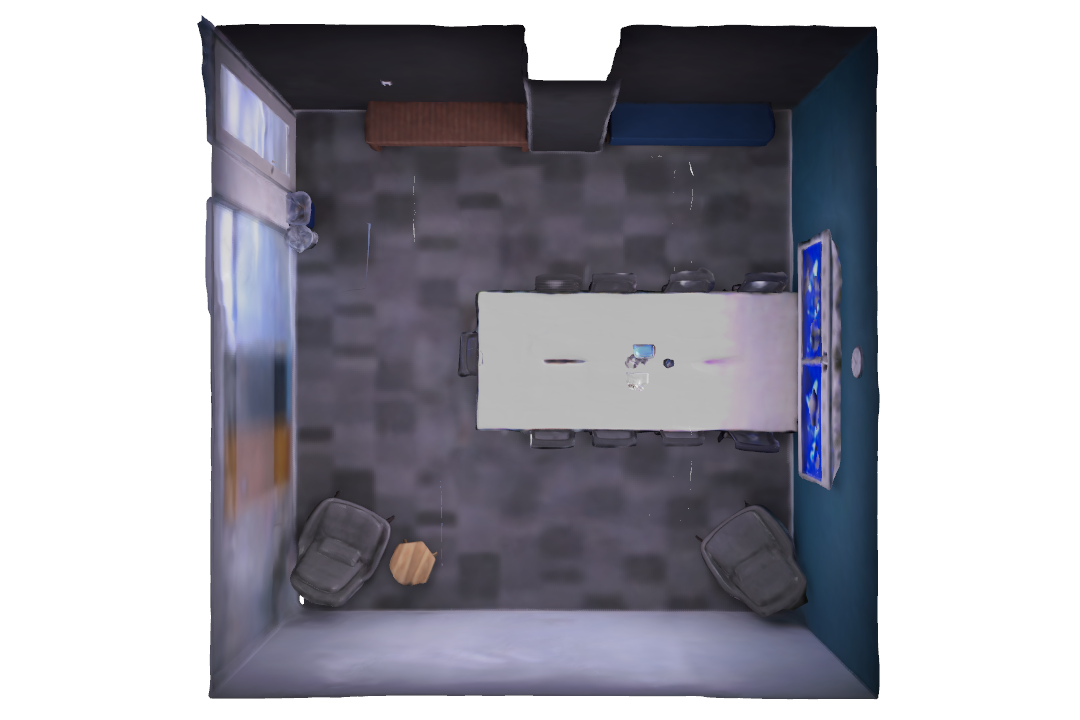} 
  \end{tabular} 
  \vspace{-2mm}
  \caption{Scene reconstruction for 4 selected Replica scenes. Interesting regions are highlighted with coloured boxes, showing \ours{}'s significantly improved reconstruction quality. All scene meshes are provided by the original authors.}
  \label{fig:replica_3D_topview}
  \vspace{-4mm}
\end{figure*}

\begin{figure*}[ht!]
  \centering
  \setlength{\tabcolsep}{0.1em}
  \renewcommand{\arraystretch}{0.4} 
  \begin{tabular}{C{0.165\linewidth}C{0.165\linewidth}C{0.165\linewidth}C{0.165\linewidth}C{0.165\linewidth}C{0.165\linewidth}}
  TSDF-Fusion & ObjSDF & \ours{} & TSDF-Fusion & ObjSDF & \ours{}\\
   \includegraphics[trim={15cm 5cm 15cm 2cm}, clip, width=\linewidth]{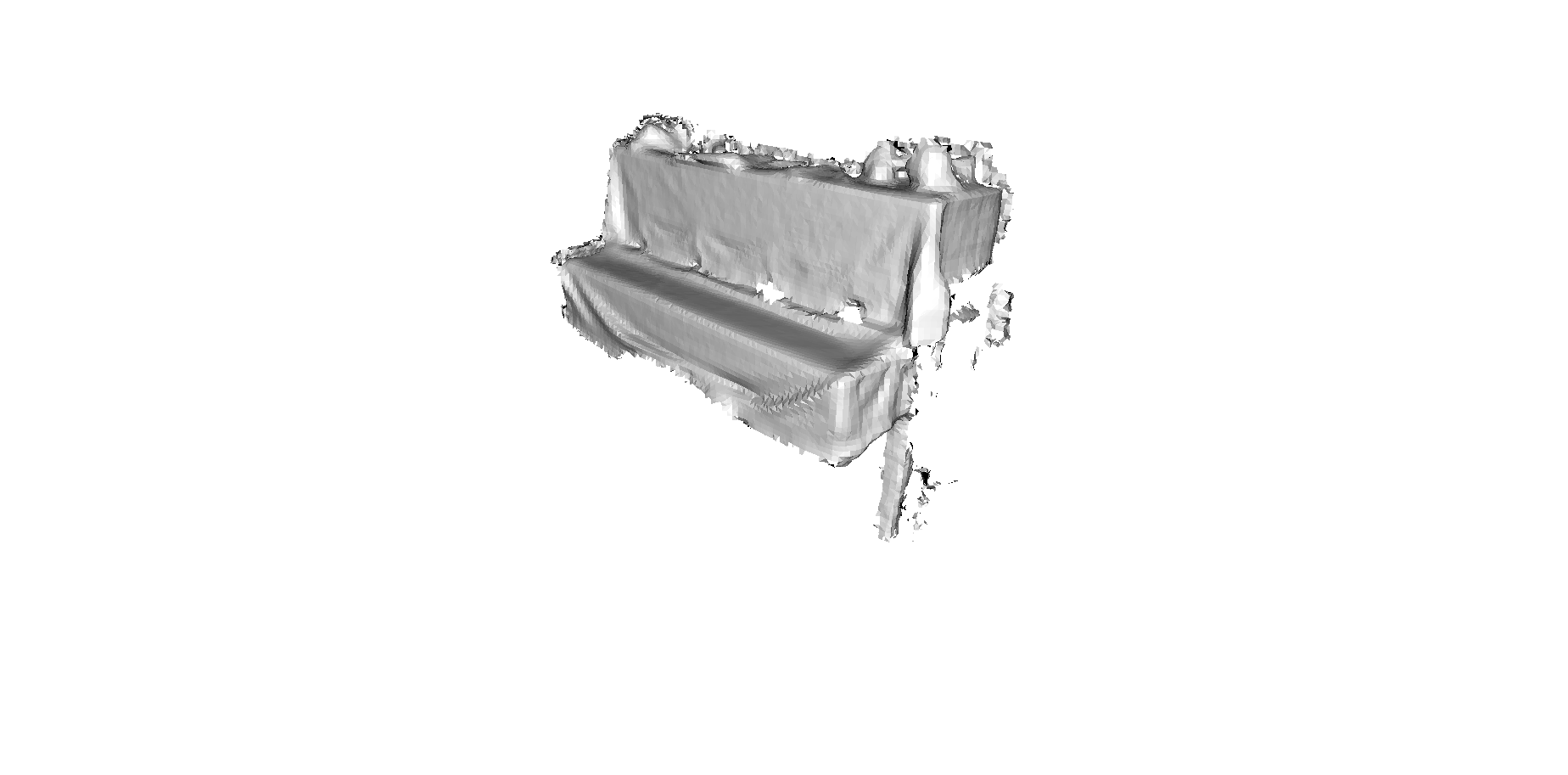} &
        \includegraphics[trim={15cm 5cm 15cm 2cm}, clip, width=\linewidth]{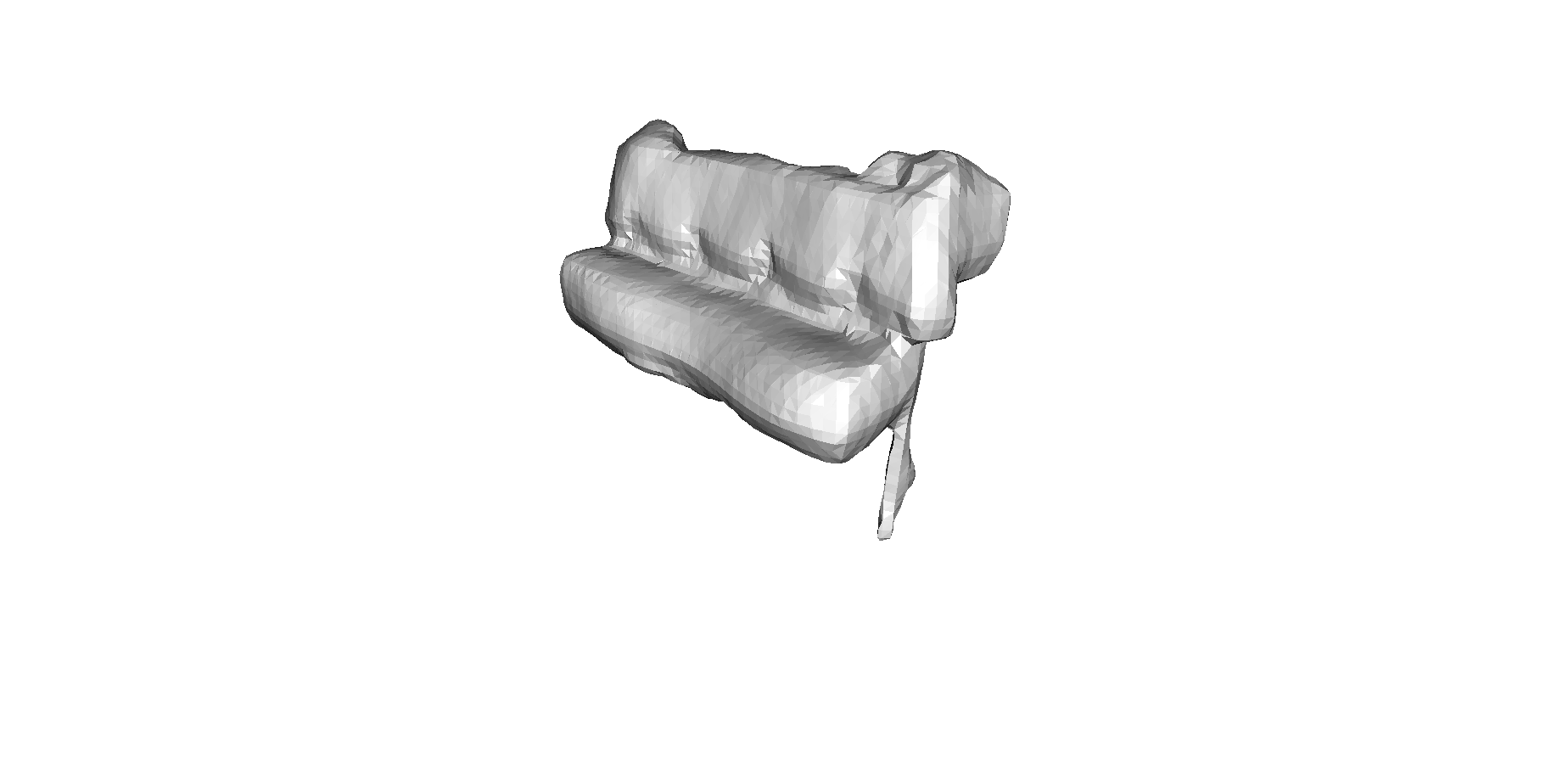} &
     \includegraphics[trim={15cm 5cm 15cm 2cm}, clip, width=\linewidth]{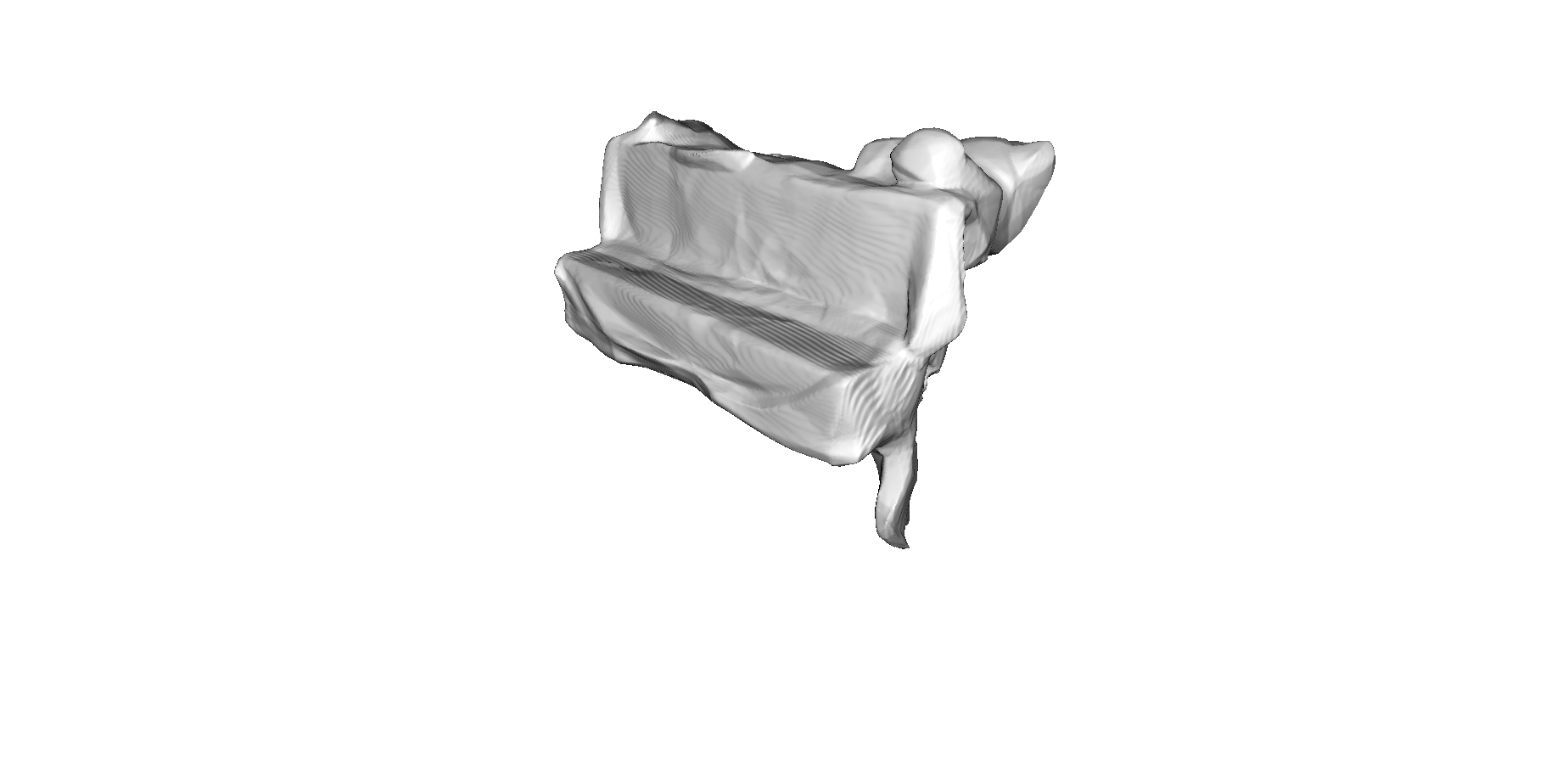} &
     \includegraphics[trim={11cm 1cm 11cm 1cm}, clip, width=\linewidth]{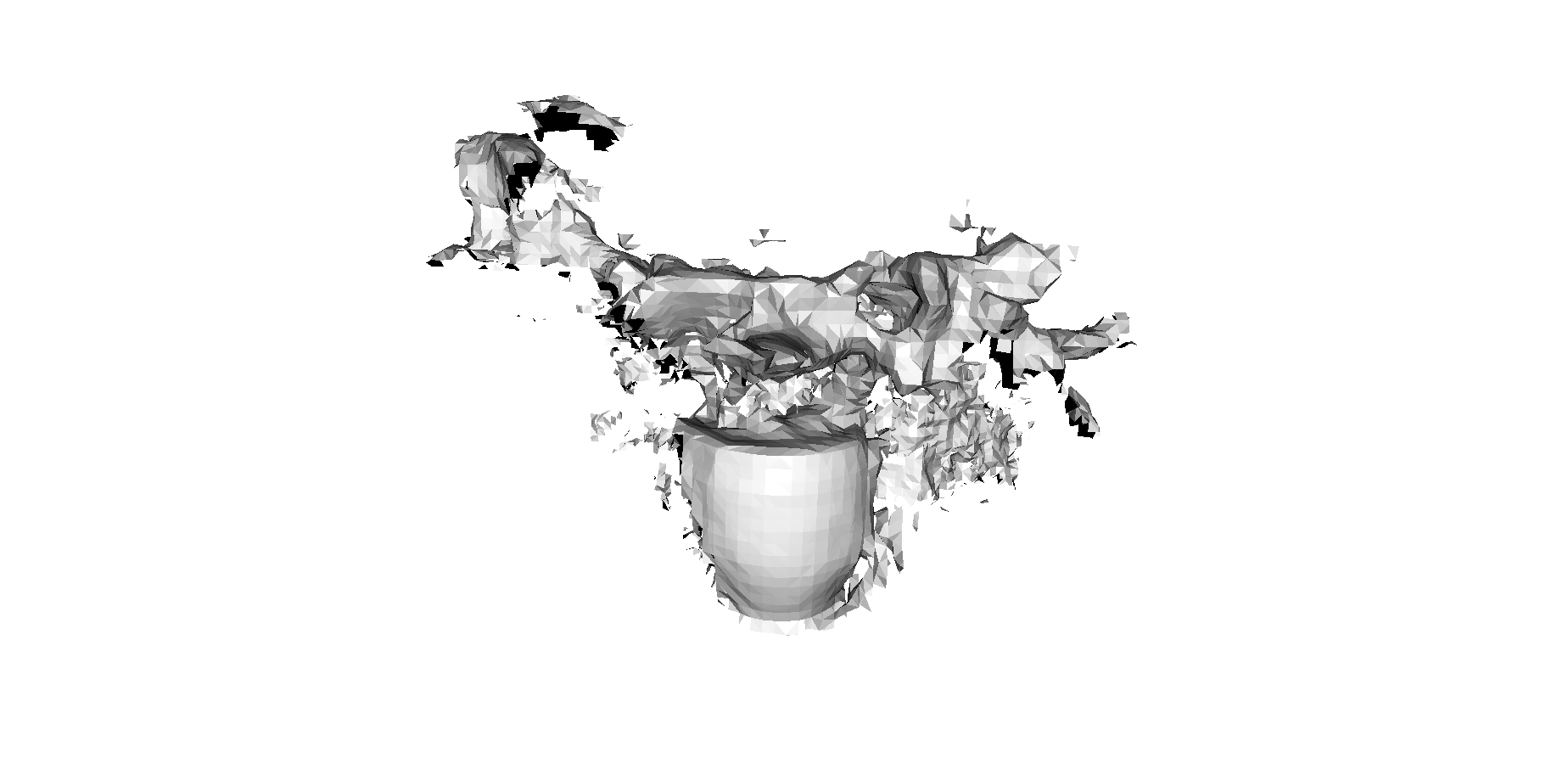} &
      \includegraphics[trim={11cm 1cm 11cm 1cm}, clip, width=\linewidth]{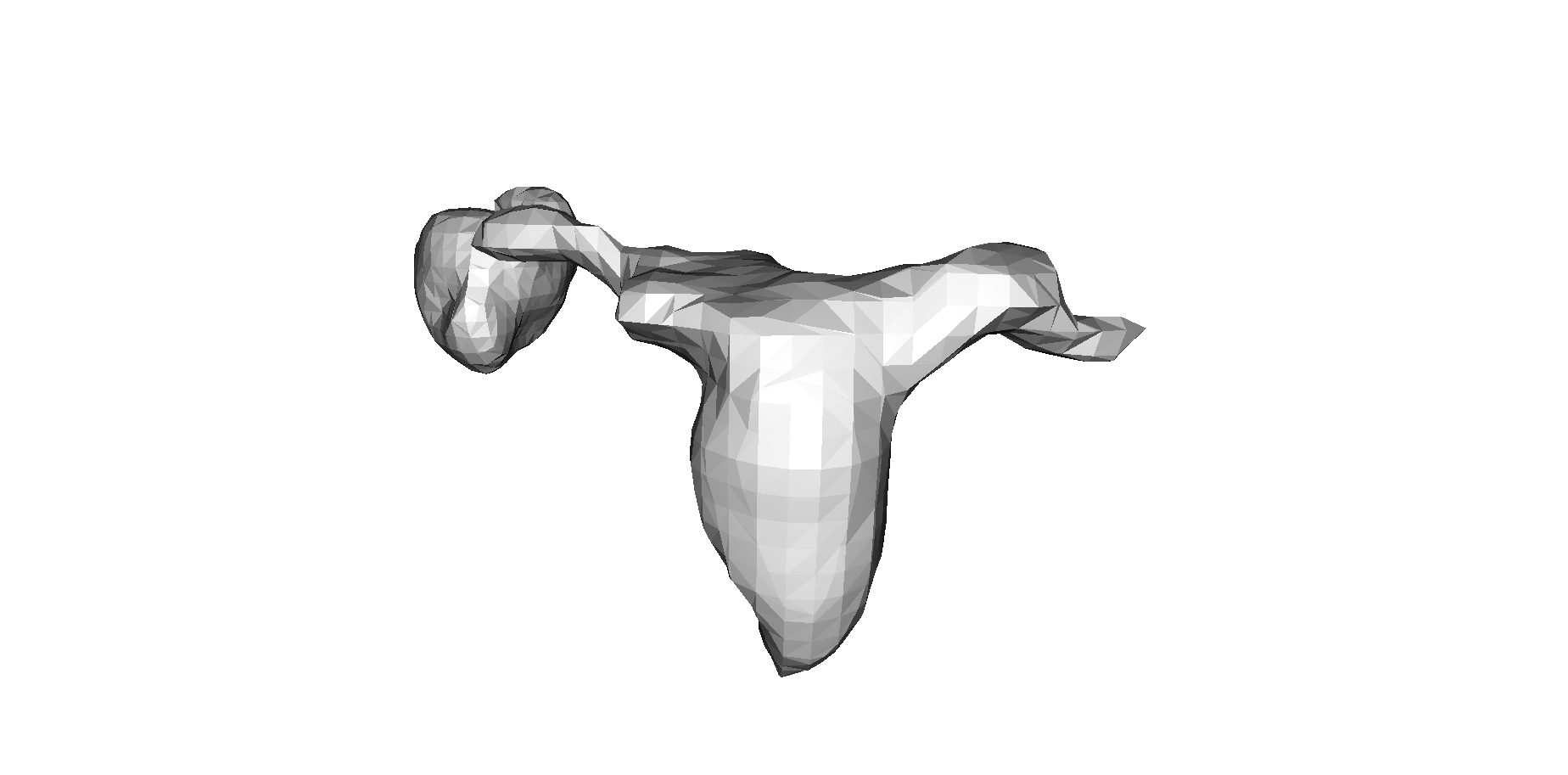} &
     \includegraphics[trim={11cm 1cm 11cm 1cm}, clip, width=\linewidth]{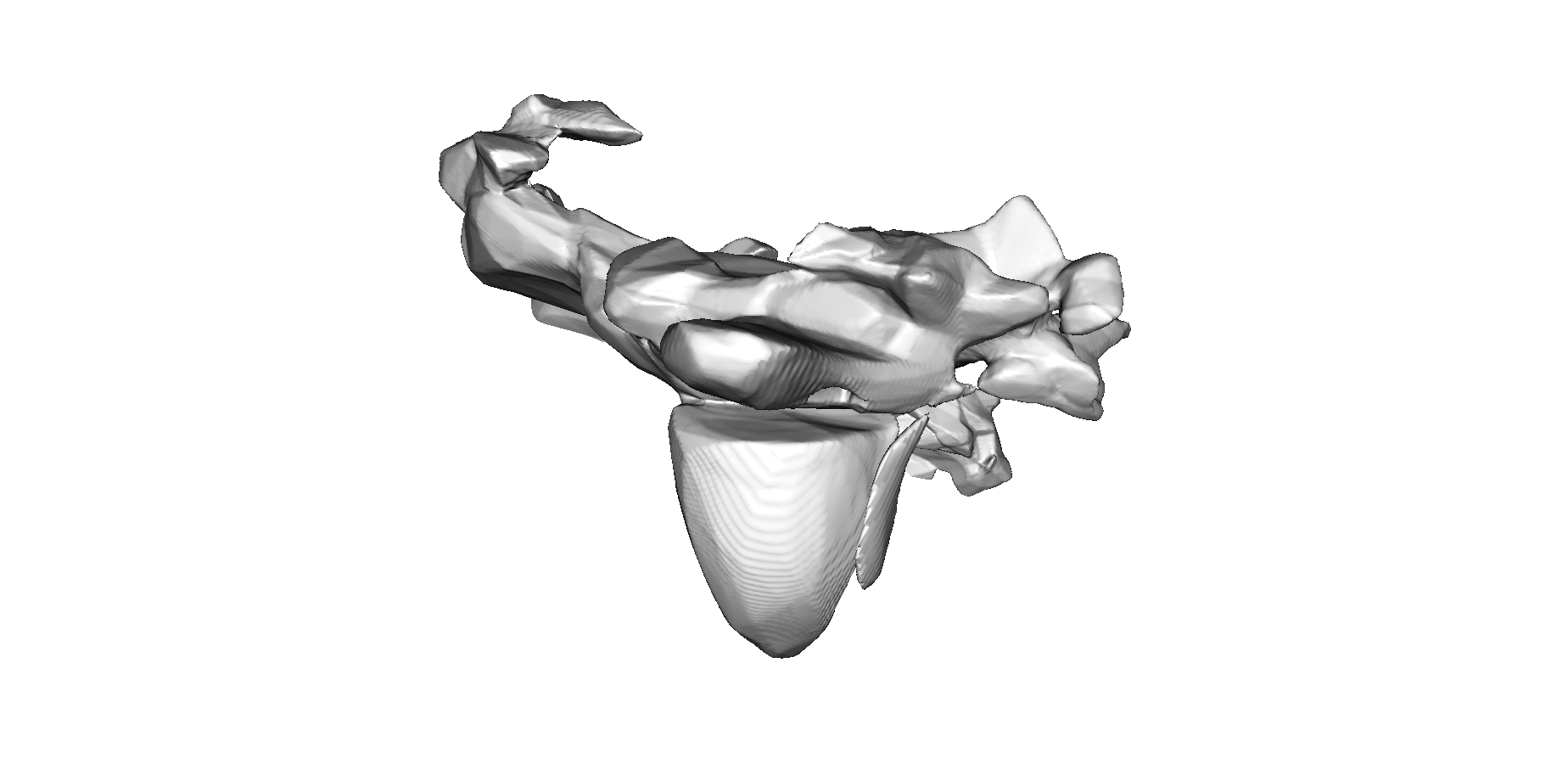} \\
       \includegraphics[trim={9cm 4cm 12cm 0cm}, clip, width=\linewidth]{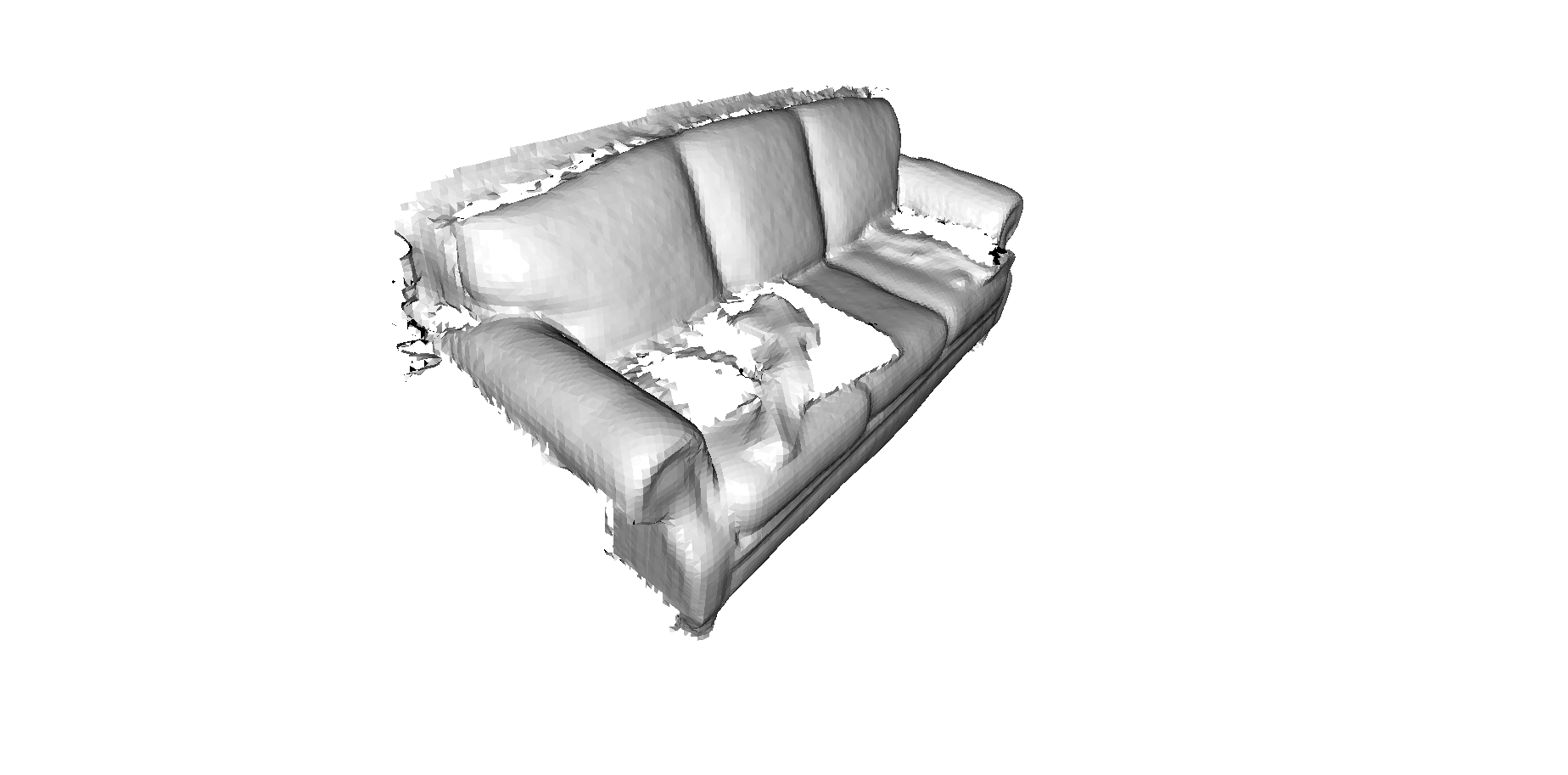} &
    \includegraphics[trim={9cm 4cm 12cm 0cm}, clip, width=\linewidth]{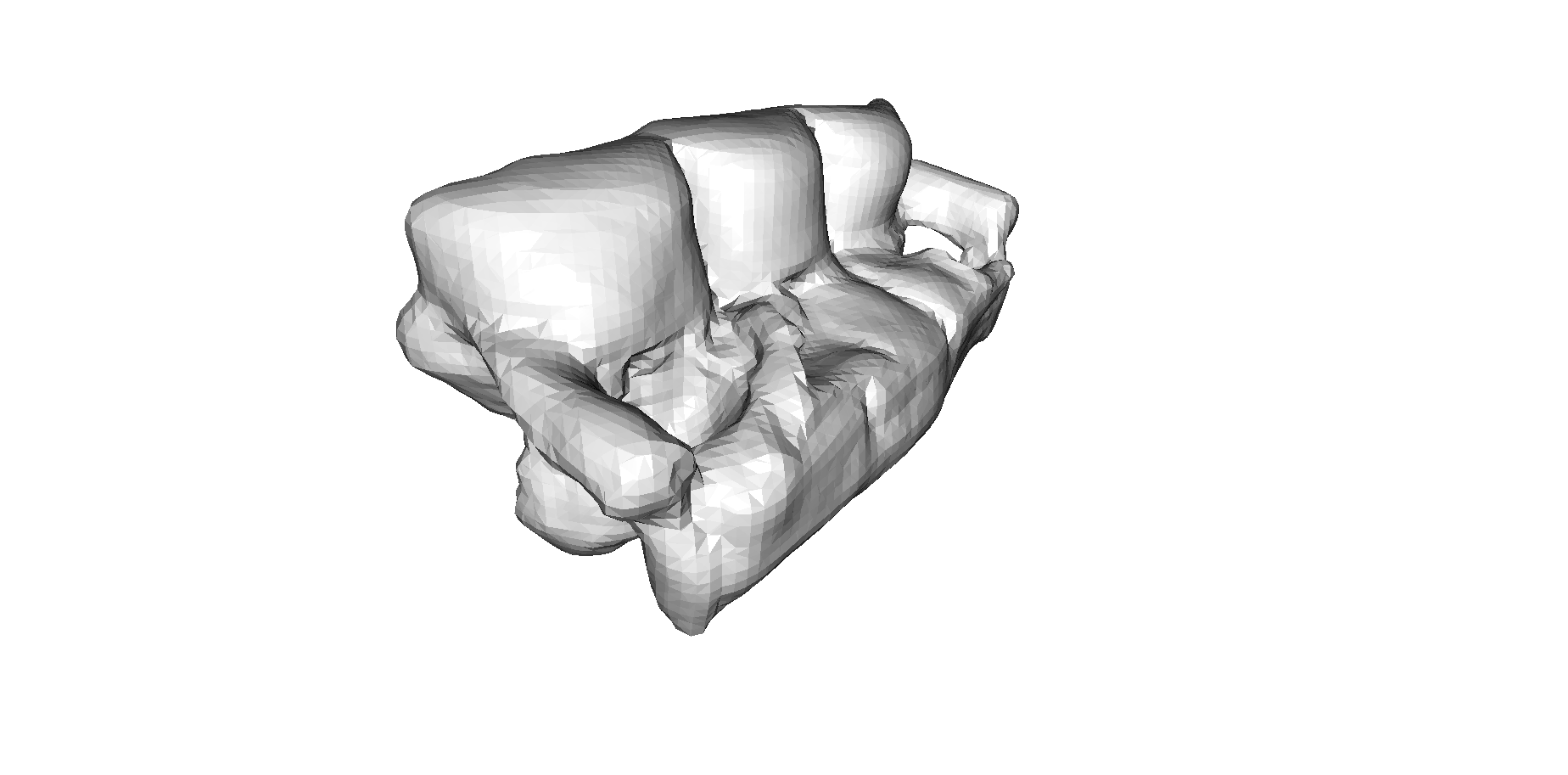} &
      \includegraphics[trim={9cm 4cm 12cm 0cm}, clip, width=\linewidth]{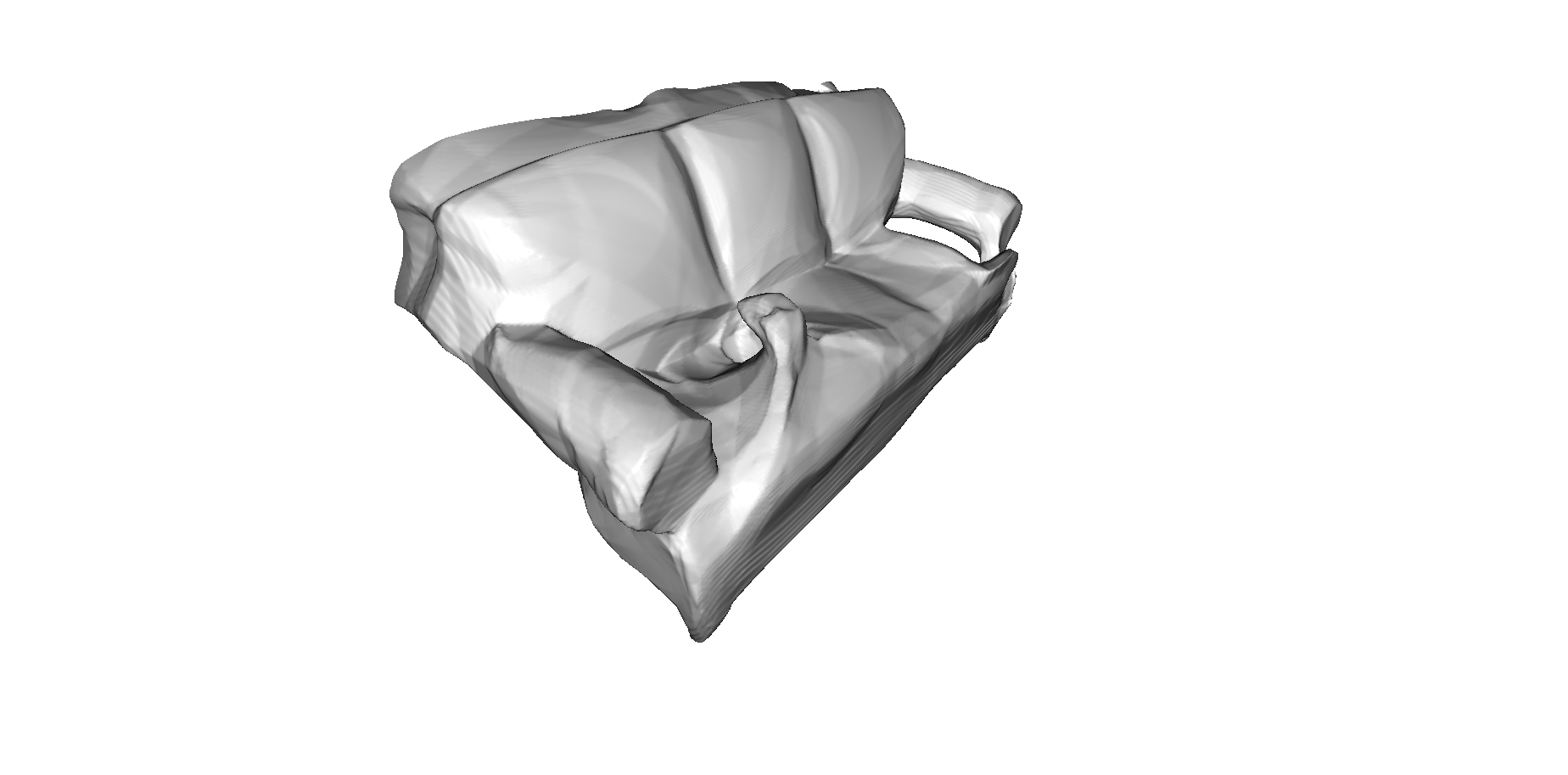} &
       \includegraphics[trim={13cm 5cm 13cm 2cm}, clip, width=\linewidth]{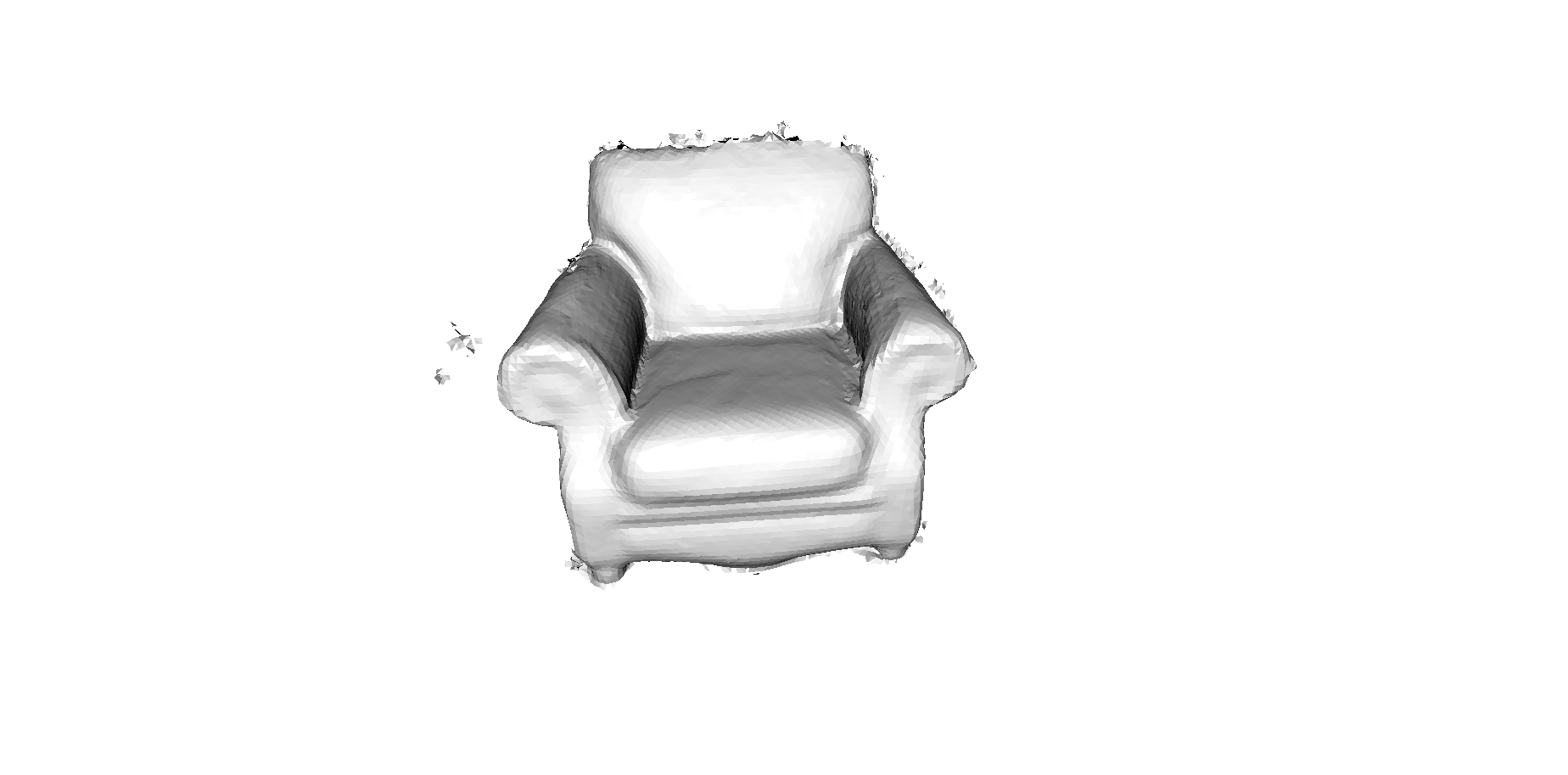} &
     \includegraphics[trim={13cm 5cm 13cm 2cm}, clip, width=\linewidth]{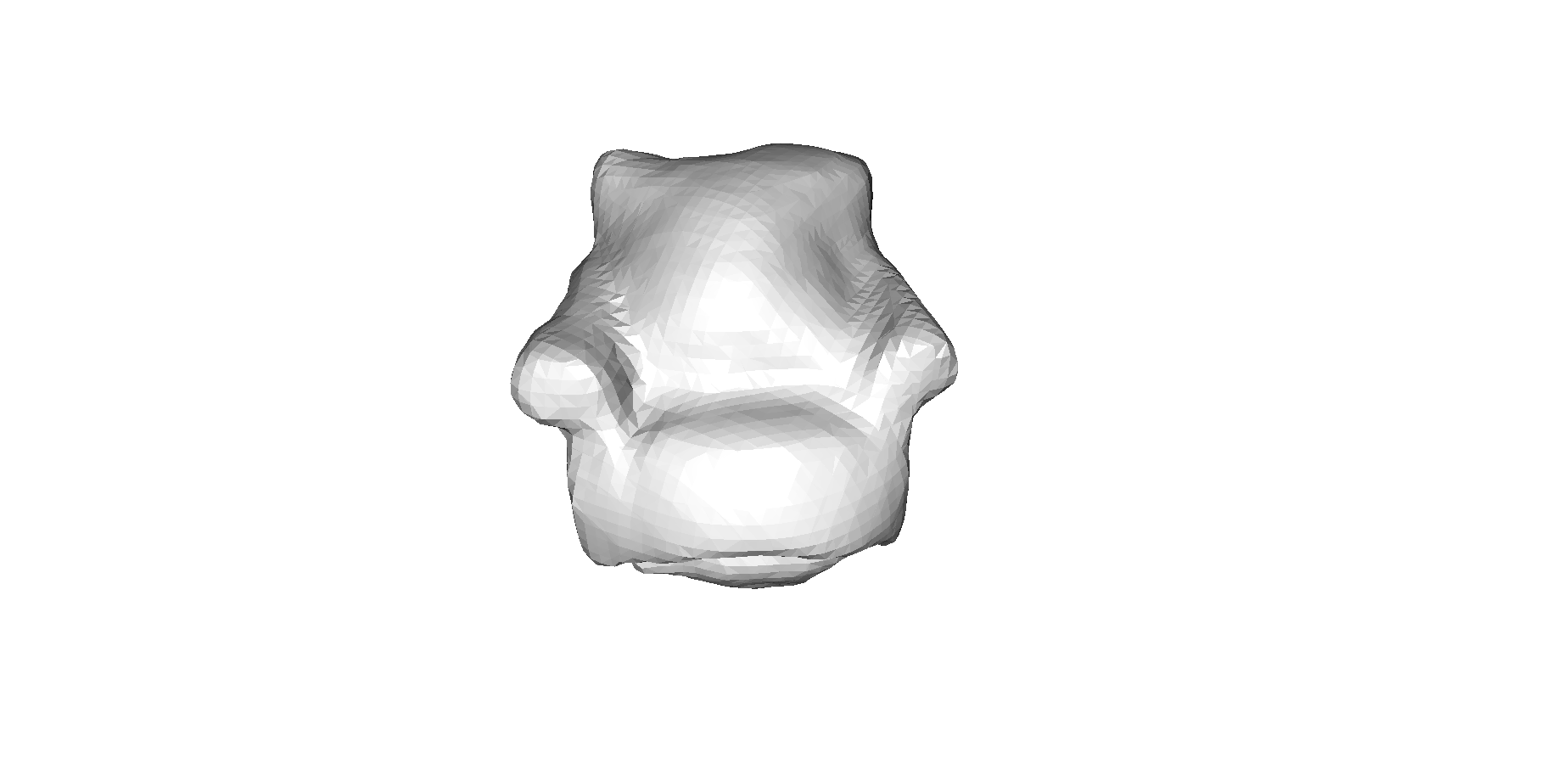} &
     \includegraphics[trim={13cm 5cm 13cm 2cm}, clip, width=\linewidth]{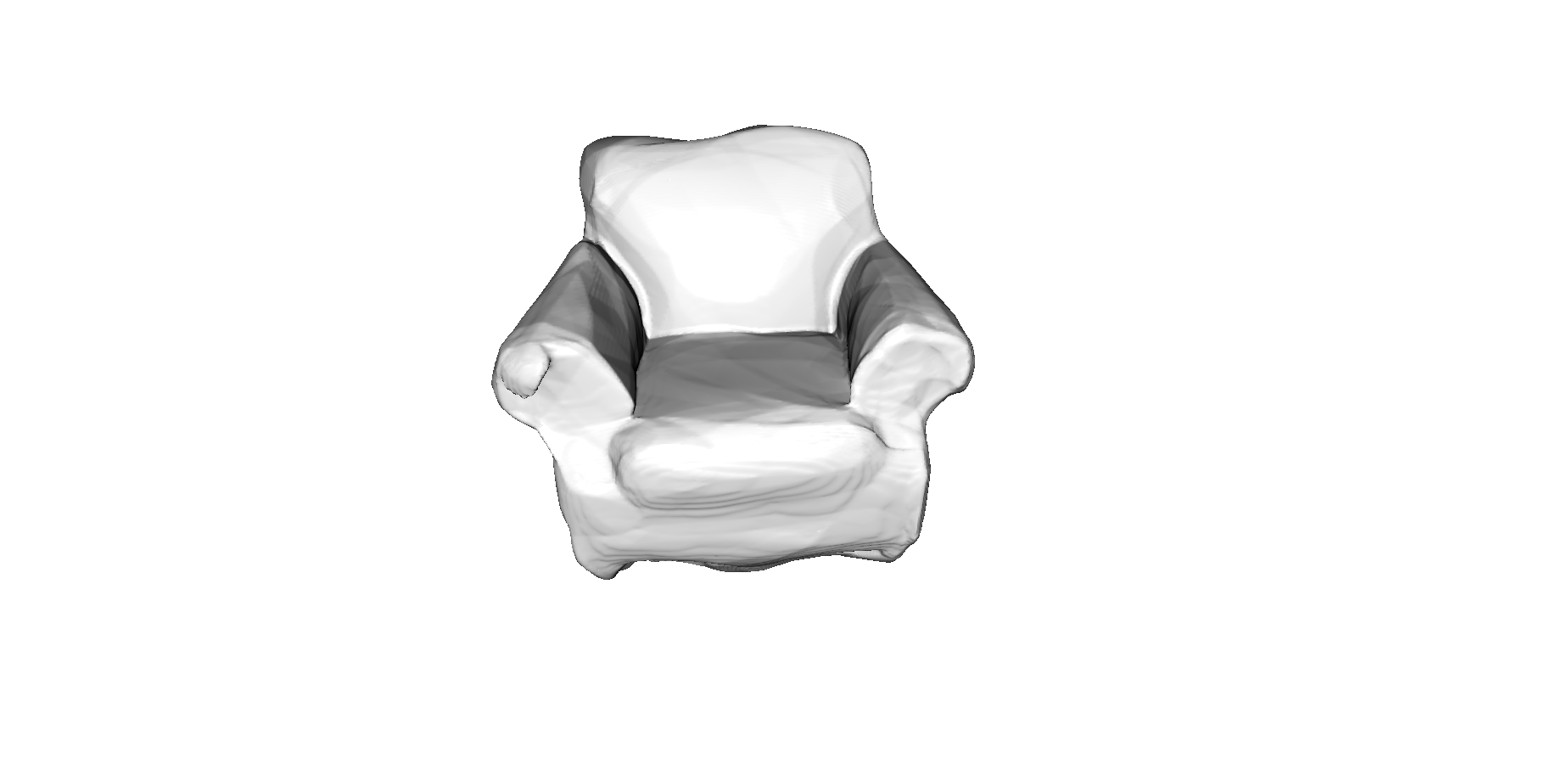} \\
  \end{tabular} 
  \caption{Visualisation of object reconstructions with \ours{} compared to TSDF-Fusion and ObjSDF. Note that all object reconstructions from ObjSDF require much longer off-line training. All object meshes from ObjSDF are provided by the original authors. }  
  \label{fig:scannet_obj}
  \vspace{-1mm}
\end{figure*}

\begin{figure*}[ht!]
    \centering
     \newcommand\makespy[1]{%
      \begin{tikzpicture}[spy using outlines={circle, magnification=2.5, height=1cm, width=1.4cm, every spy on node/.append style={line width=1.5}}]
      \centering
        \node {\includegraphics[width=\linewidth]{#1}};
        \spy[color=Bittersweet] on (-1.2, 1.1) in node[line width=1.5] at (-1.2, 1.1);
        \spy[color=Melon] on (1.0, 1.7) in node[line width=1.5] at (1.0, 1.7);
        \spy[color=Lavender] on (-2.0,-0.4) in node[line width=1.5] at (-2.0,-0.4);
       \end{tikzpicture}
   }
  \small
  \setlength{\tabcolsep}{0.5em}
  \renewcommand{\arraystretch}{0.2} 
    \begin{tabular}{C{0.495\linewidth}C{0.495\linewidth}}
    NICE-SLAM$^\ast$ & \ours{} \\
    \makespy{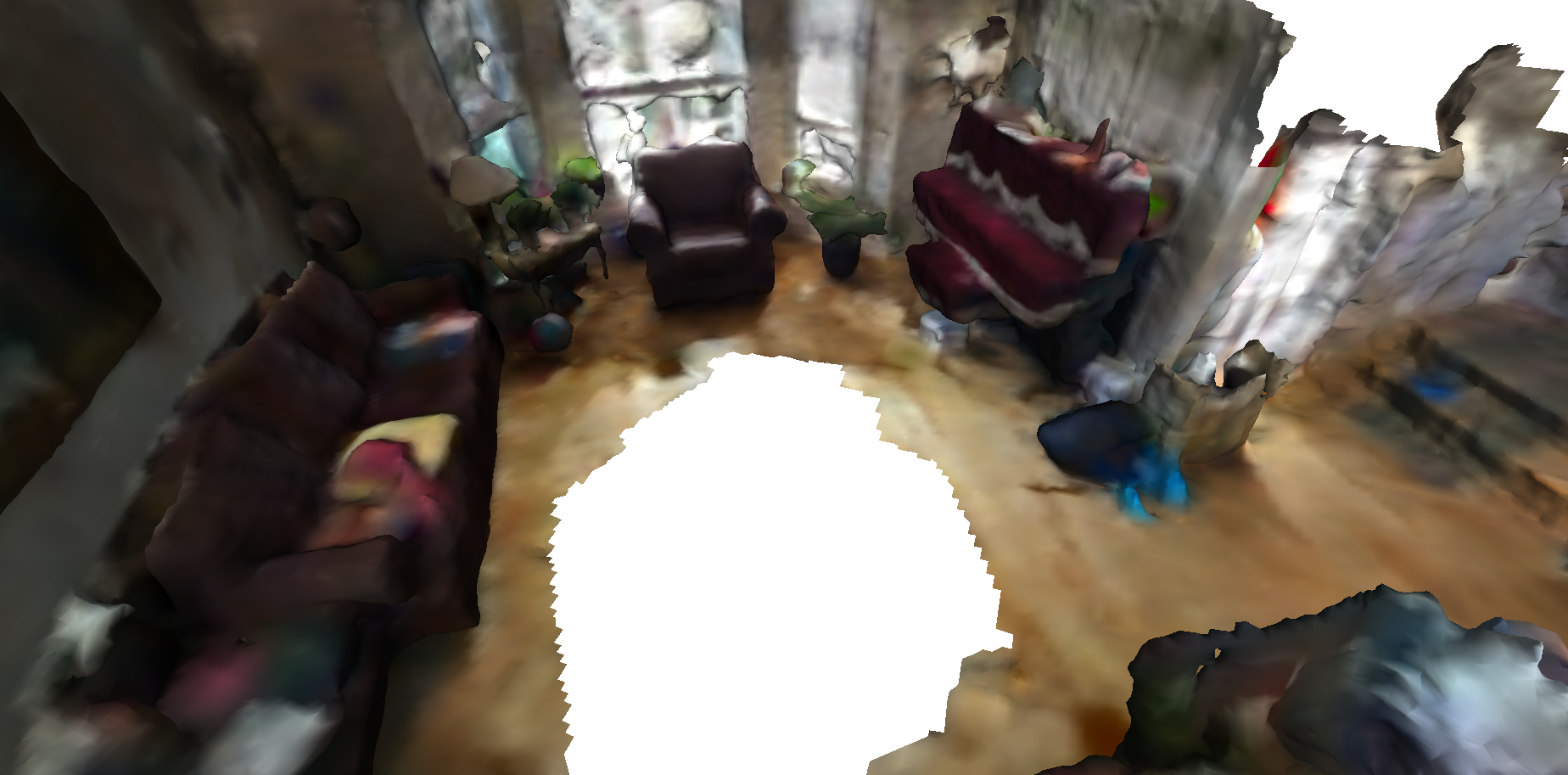} &
    \makespy{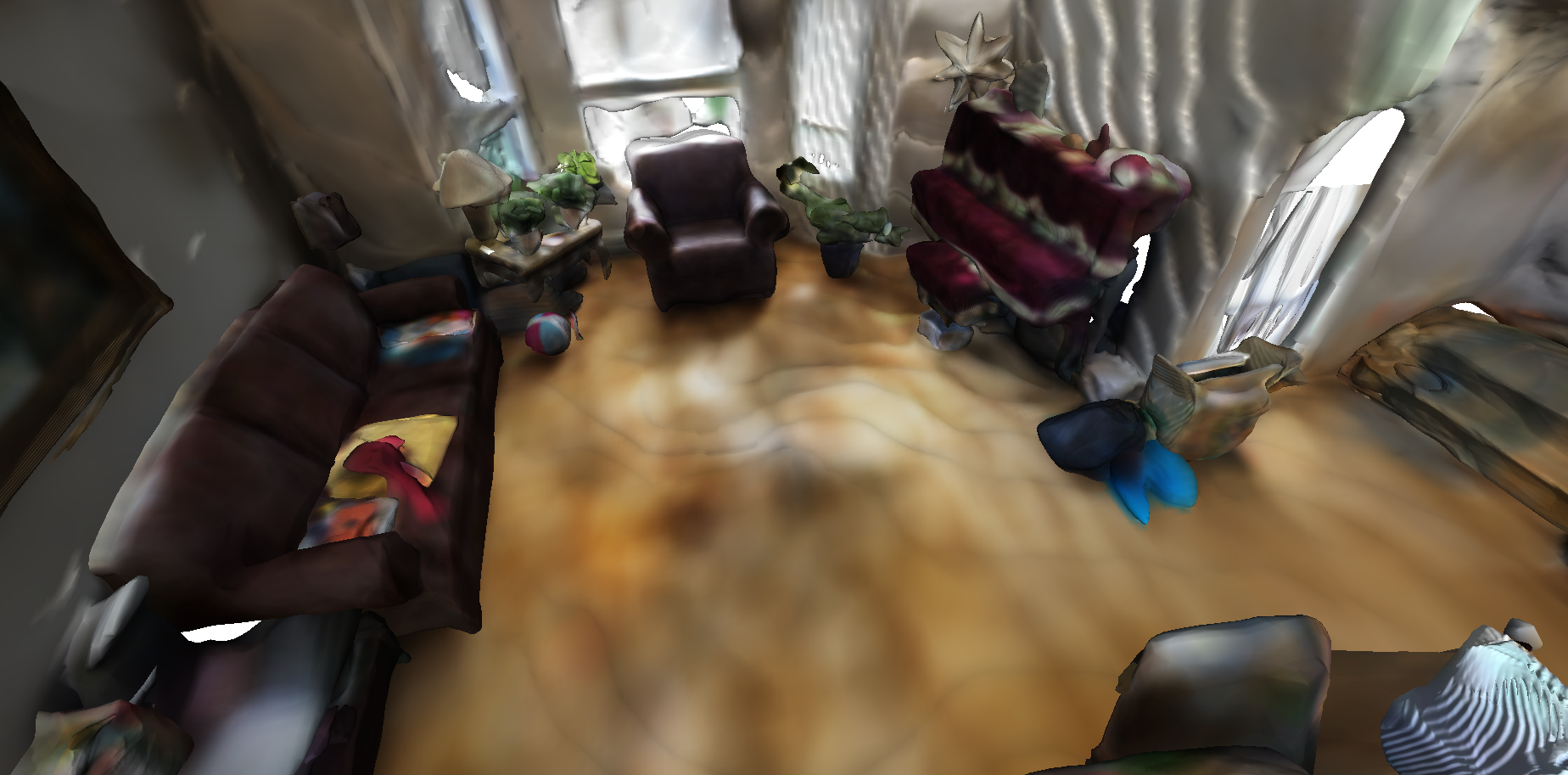}
    \end{tabular}
   \vspace{-2mm}
   \caption{Visualisation of scene reconstruction from NICE-SLAM$^\ast$ (left) and \ours{} (right) in a selected ScanNet sequence. Interesting regions are zoomed in. NICE-SLAM$^\ast$ was re-trained with ground-truth poses.}
   \label{fig:scannet_scene}  
   \vspace{-3mm}
\end{figure*}

\paragraph{Implementation Details}
We conduct all experiments on a desktop PC with a 3.60~GHz i7-11700K CPU and a single Nvidia RTX 3090 GPU. We choose our instance segmentation detector to be Detic \cite{zhou2022detic}, pre-trained on an open-vocabulary LVIS dataset \cite{gupta2019lvis} which contains more than 1000 object classes. We choose our pose estimation framework to be ORB-SLAM3 \cite{campos2021orb}, for its fast and accurate tracking performance. We continuously update the keyframe poses using the latest estimates from ORB-SLAM3.

We applied the same set of hyper-parameters for all datasets. Both our object and background model use 4-layer MLPs, with each layer having hidden size 32 (object) and 128 (background). For object / background, we selected keyframes every 25 / 50 frames, 120 / 1200 rays each training step with 10 points per ray. The number of objects in a scene typically varies between 20 and 70, among which the largest number of objects are in Replica and ScanNet scenes with an average of 50 objects per scene.

\paragraph{Metrics}
Following the convention of prior work \cite{Sucar:etal:ICCV2021, zhu2022nice}, we adopt {\it Accuracy}, {\it Completion}, and {\it Completion Ratio} for 3D scene-level reconstruction metrics. Besides, we note that such scene-level metrics are heavily biased towards the reconstruction of large objects like walls and floors. Therefore, we additionally provide these metrics at  the object-level, by averaging metrics for all objects in each scene.

\subsection{Evaluation on Scene and Object Reconstruction}

\paragraph{Results on Replica}
We experimented on 8 Replica scenes, using the rendered trajectories provided in \cite{Sucar:etal:ICCV2021}, with 2000 RGB-D frames in each scene. Tab.~\ref{tab:replica_results} shows the averaged quantitative reconstruction results in these Replica indoor sequences. For scene-level reconstruction, we compared with TSDF-Fusion~ \cite{zhou2013dense}, iMAP~\cite{Sucar:etal:ICCV2021} and NICE-SLAM~ \cite{zhu2022nice}. To isolate reconstruction, we also provided results for these baselines re-trained with ground-truth pose (marked with $\ast$), with their open-sourced code for the fair comparison. Specifically, iMAP$^\ast$ was implemented as a special case of \ours{}, when considering the entire scene as one object instance. For object-level reconstruction, we compared baselines trained with ground-truth pose.

\ours{}'s significant advantage thanks to object-level representation is to reconstruct tiny objects and objects with fine-grained details. Noticeably, \ours{} achieved more than 50 -- 70\% improvement over iMAP and NICE-SLAM for object-level completion. The scene reconstructions of 4 selected Replica sequences are shown in Fig.~\ref{fig:replica_3D_topview}, with interesting regions highlighted in coloured boxes. The quantitative results for 2D novel view rendering are further provided in the supplementary material.

\paragraph{Results on ScanNet}
To evaluate on a more challenging setting, we experimented on ScanNet \cite{dai2017scannet}, a dataset composed of real scenes, with much noisier ground-truth depth maps and object masks. We choose a ScanNet sequence selected by ObjSDF \cite{wu2022object}, and we compared with TSDF-Fusion and ObjSDF for object-level reconstruction, and we compared with NICE-SLAM (re-trained with ground-truth pose) for scene-level reconstruction. Unlike ObjSDF, which was optimised from pre-selected posed images without depth for much longer off-line training, we ran both \ours{} and TSDF-Fusion in an online setting with depth.  As shown in Fig.~\ref{fig:scannet_obj}, we see that \ours{} generates objects with more coherent geometry than TSDF-Fusion; and with much finer details than ObjSDF, though with a much shorter training time. And consistently, we can see that \ours{} generates much sharper object boundaries and textures compared to NICE-SLAM, as shown in Fig.~\ref{fig:scannet_scene}.

\begin{figure}[t!]
    \centering
    \includegraphics[trim={5cm 0cm 4cm 0cm}, clip, width=0.495\linewidth]{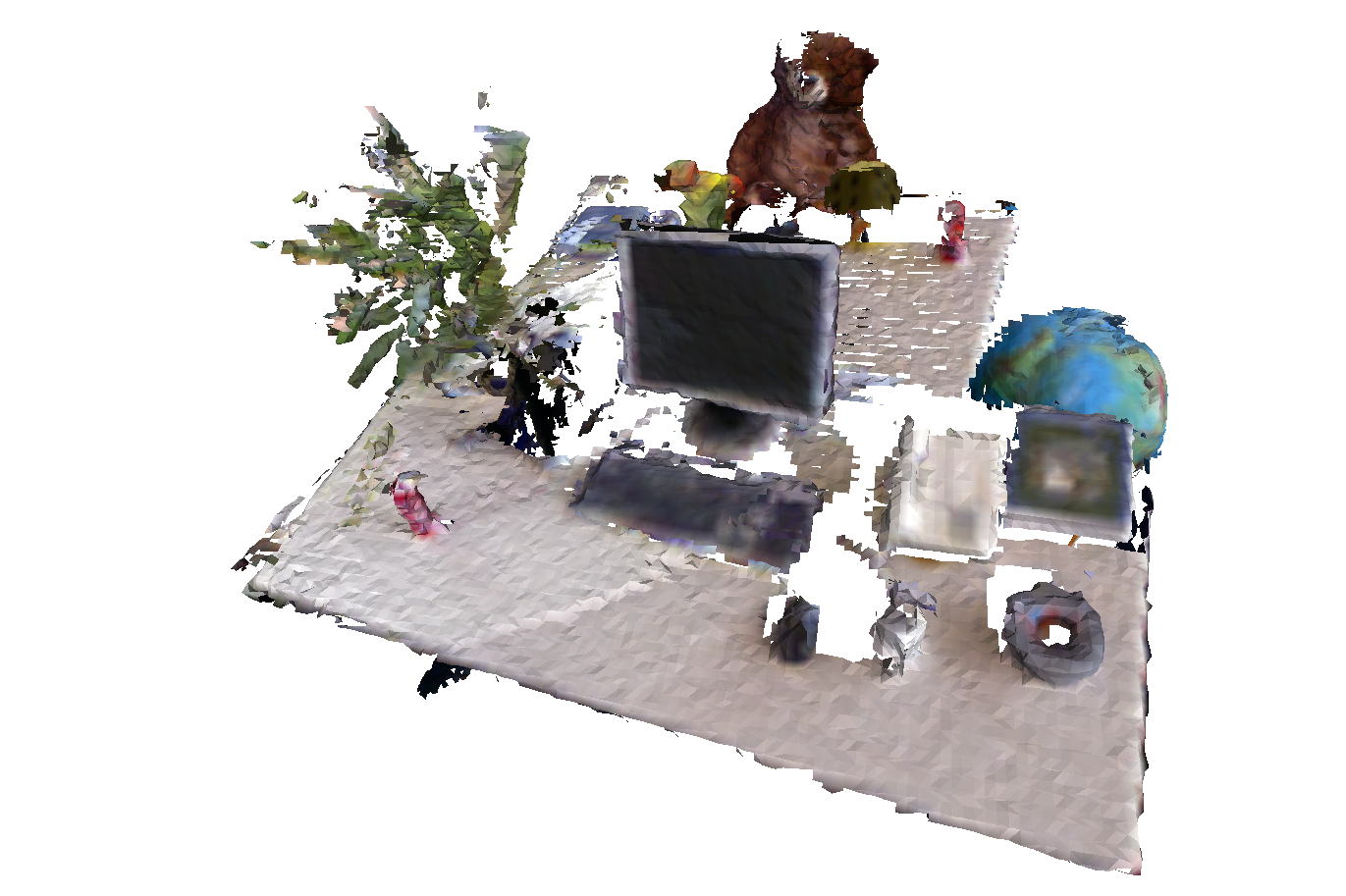} \hfill
    \includegraphics[trim={5cm 0cm 4cm 0cm}, clip, width=0.495\linewidth]{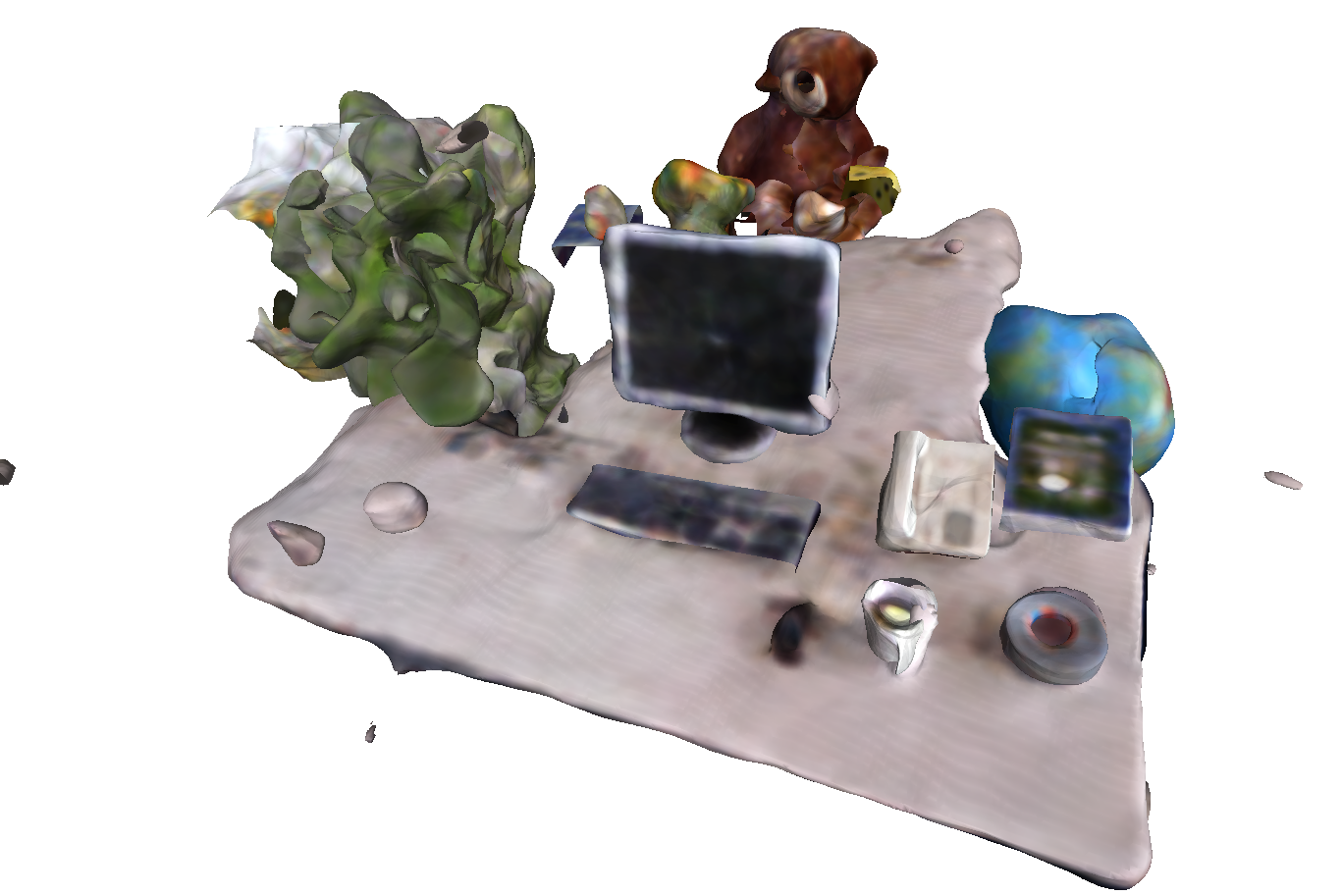}
   \caption{Visualisation of scene reconstruction from TSDF-Fusion (left) and \ours{} (right) in a selected TUM RGB-D sequence, trained in real time for 99 seconds.}
   \label{fig:tum}  
\end{figure}

\paragraph{Results on TUM RGB-D}
We evaluated on a TUM RGB-D sequence captured in the real-world, with object masks predicted by an off-the-shelf pre-trained instance segmentation network \cite{zhou2022detic}, and poses estimated by ORB-SLAM3 \cite{campos2021orb}. Since our object detector has no spatio-temporal consistency, we found that the same object can be occasionally detected as two different instances, which leads to some reconstruction artifacts. For example, the object `globe'  shown in Fig.~\ref{fig:tum} was also detected as `balloon' in some frames, resulting the `splitting` artifacts in the final object reconstruction. Overall, \ours{} still predicts more coherent reconstruction for most objects in a scene, with realistic hole-filling capabilities compared to TSDF-Fusion. However, we acknowledge that the completion of complete out-of-view regions (e.g., the back of a chair) is beyond the reach of our system due to the lack of general 3D prior.

\begin{table}[t!]
  \centering
  \footnotesize
  \setlength{\tabcolsep}{0.5em}
    \begin{tabular}{lcccc}
      \toprule
        ATE RMSE [cm]$\downarrow$ & iMAP & NICE-SLAM & vMAP & ORB-SLAM2 \\
         \midrule
        fr1/desk & 4.9 & 2.7   & 2.6 &\textbf{1.6}  \\
        fr2/xyz & 2.0 & 1.8    & 1.6 &\textbf{0.4} \\
        fr3/office & 5.8 & 3.0 & 3.0 &\textbf{1.0}  \\
       \bottomrule
    \end{tabular}
    \vspace{-1mm}
    \caption{Camera tracking results on TUM RGB-D.}
    \label{tab:track}
    \vspace{-4mm} 
\end{table}

Though our work focuses more on mapping performance than pose estimation, we also report {\it ATE RMSE}~\cite{Sturm:etal:IROS2012} in Tab.~\ref{tab:track} following \cite{Sucar:etal:ICCV2021, zhu2022nice}, by jointly optimising camera pose with map. We can observe that vMAP achieves superior performance, due to the fact that reconstruction and tracking quality are typically highly interdependent. However, there is a noticeable performance gap compared to ORB-SLAM. As such, we directly choose ORB-SLAM as
our external tracking system, which leads to faster training
speed, cleaner implementation, and higher tracking quality.

\begin{figure}[t!]
    \centering
    \includegraphics[width=0.75\linewidth]{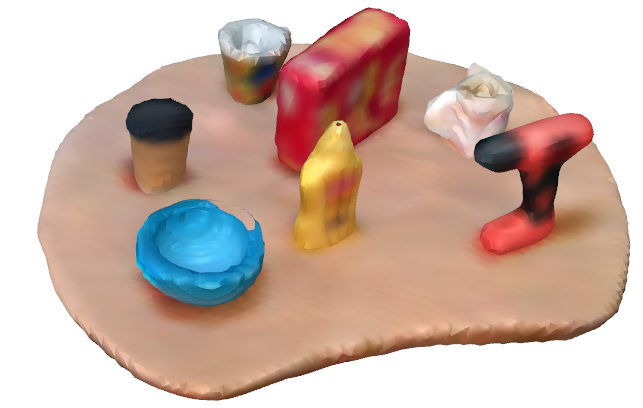} \\
    \includegraphics[width=\linewidth]{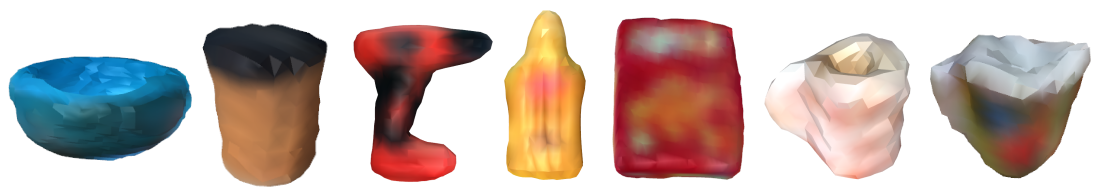}
   \caption{Visualisation of table-top reconstruction (top) and individual object reconstructions (bottom), from \ours{} running in real time using an Azure Kinect RGB-D camera for 170 seconds. }
   \label{fig:real_world}  
\end{figure}

\paragraph{Results on Live Kinect Data}
Finally, we show the reconstruction result of \ours{} on a table-top scene, from running in real-time with an Azure Kinect RGB-D camera. As shown in Fig.~\ref{fig:real_world}, \ours{} is able to generate a range of realistic, watertight object meshes from different categories.

\subsection{Performance Analysis}
\label{subsec: performance}
In this section, we compare different training strategies and architectural design choices for our \ours{} system. For simplicity, all experiments were done on the Replica Room-0 sequence, with our default training hyper-parameters. 

\begin{figure}[t!]
    \centering
    \begin{minipage}[t]{.50\linewidth}
    \centering
    \large
    \scalebox{0.55}{\begin{tikzpicture}

\definecolor{color0}{RGB}{125, 111, 190}
\definecolor{color1}{RGB}{211, 82, 66}
\definecolor{color2}{RGB}{63, 159, 204}
\definecolor{color3}{RGB}{38, 192, 159}

\begin{axis}[
every axis y label/.style={at={(current axis.north  west)},above=0mm,xshift=0.8cm},
every axis x label/.style={at={(current axis.south)},yshift=-1cm},
width=8cm, height=5.0cm,
legend cell align={left},
legend style={fill opacity=0.0, draw opacity=1, text opacity=1, draw=none, column sep=2pt, very thick, anchor=south east, at={(0.98,0.02)}, font=\small},
tick align=outside,
tick pos=left,
xlabel={Param. Size (MB)},
xmin=-0.1, xmax=2.0,
ylabel={\makecell{Obj. Acc. (cm) $\downarrow$}},
ymin=0, ymax=4.5,
ytick={0,1,2,3,4},
]

\addplot [ultra thick, color1, mark=*, mark size=1.5, mark options={solid}]
table {%
0.16 2.38
0.33 1.98 
0.66 1.97 
1.32 1.97 
};
\node[above] at (axis cs:0.16, 2.38) {\small 8};
\node[above] at (axis cs:0.33, 1.98) {\small 16};
\node[above] at (axis cs:0.66, 1.97) {\small 32};
\node[above] at (axis cs:1.32, 1.97) {\small 64};
\addlegendentry{vMAP}

\addplot [ultra thick, color0, mark=*, mark size=1.5, mark options={solid}]
table {
0.32 1.99
};
\node[below] at (axis cs:0.32, 1.99) {\small vMAP-S};

\addplot [ultra thick, color2, mark=*, mark size=1.5, mark options={solid}]
table {%
0.09 3.65
0.32 3.57
0.68 3.55
1.16 3.51
};
\node[above] at (axis cs:0.09, 3.65) {\small 128};
\node[above] at (axis cs:0.32, 3.57) {\small 256};
\node[above] at (axis cs:0.68, 3.55) {\small 384};
\node[above] at (axis cs:1.16, 3.51) {\small 512};
\addlegendentry{iMAP}

\end{axis}

\end{tikzpicture}}
    \end{minipage}\hfill
    \begin{minipage}[t]{.50\linewidth}
    \centering
    \large
    \scalebox{0.55}{\begin{tikzpicture}

\definecolor{color0}{RGB}{125, 111, 190}
\definecolor{color1}{RGB}{211, 82, 66}
\definecolor{color2}{RGB}{63, 159, 204}
\definecolor{color3}{RGB}{38, 192, 159}

\begin{axis}[
every axis y label/.style={at={(current axis.north west)},above=0mm,xshift=1.8cm},
every axis x label/.style={at={(current axis.south)},yshift=-1cm},
width=8cm, height=5cm,
legend cell align={left},
legend style={fill opacity=0.0, draw opacity=1, text opacity=1, draw=none, column sep=2pt, very thick, anchor=south east, at={(0.98,0.02)}, font=\small},
tick align=outside,
tick pos=left,
xlabel={Param. Size (MB)},
xmin=-0.1, xmax=2.0,
ylabel={Comp. Ratio ($<1$cm \%) $\uparrow$},
ylabel={\makecell{Obj. Comp. Ratio ($<1$cm \%) $\uparrow$}},
ymin=25, ymax=75,
ytick={30, 40, 50, 60, 70},
]

\addplot [ultra thick, color1, mark=*, mark size=1.5, mark options={solid}]
table {%
0.16 62.66
0.33 66.17
0.66 66.49
1.32 66.46
};
\node[above] at (axis cs:0.16, 62.66) {\small 8};
\node[above] at (axis cs:0.33, 66.17) {\small 16};
\node[above] at (axis cs:0.66, 66.49) {\small 32};
\node[above] at (axis cs:1.32, 66.46) {\small 64};
\addlegendentry{vMAP}

\addplot [ultra thick, color0, mark=*, mark size=1.5, mark options={solid}]
table {
0.32 62.97
};
\node[below] at (axis cs:0.32, 62.97) {\small vMAP-S};

\addplot [ultra thick, color2, mark=*, mark size=1.5, mark options={solid}]
table {%
0.09 46.25
0.32 47.79
0.68 52.82
1.16 49.07
};
\node[above] at (axis cs:0.09, 46.25) {\small 128};
\node[above] at (axis cs:0.32, 47.79) {\small 256};
\node[above] at (axis cs:0.68, 52.82) {\small 384};
\node[above] at (axis cs:1.16, 49.07) {\small 512};
\addlegendentry{iMAP}

\end{axis}

\end{tikzpicture}}
    \end{minipage}
    \caption{Object-level Reconstruction v.s. Model Param. (denoted by network hidden size). \ours{} is more compact than iMAP, with the performance starting to saturate from hidden size 16.}
    \label{fig:ablation_hiddensize}
    \vspace{-4mm} 
\end{figure}
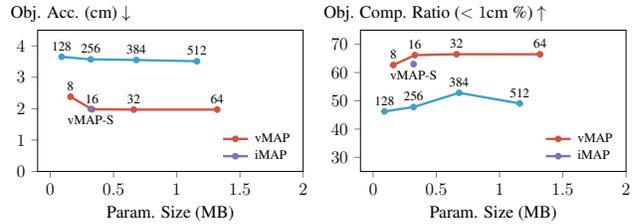

\paragraph{Memory and Runtime}
We compared memory usage and runtime with iMAP and NICE-SLAM in Tab.~\ref{tab:time} and Fig.~\ref{fig:ablation_hiddensize}, all trained with ground-truth pose, and with the default training hyper-parameters listed in each method, for fair comparison. Specifically, we reported the 
{\it Runtime} for training the entire sequence, and {\it Mapping Time} for training each single frame, given the exact same hardware.  We can observe that \ours{} is highly memory efficient with less than 1M parameters. We want to highlight that \ours{} achieves better reconstruction quality, and runs significantly faster ($\sim$5Hz) than iMAP and NICE-SLAM with 1.5x and 4x training speed improvement respectively.

\begin{table}[t!]
  \centering
  \footnotesize
 \setlength{\tabcolsep}{0.2em}
    \begin{tabular}{lcccc}
      \toprule
         & NICE-SLAM$^\ast$ & iMAP &  \ours{} & \ours{} (w/o BG)  \\
         \midrule
        Model Param. $\downarrow$ & 12.12M & 0.32M  & 0.66M  & 0.56M \\
        Runtime $\downarrow$   & 34min34s & 12min29s  & 8min16s & 6min01s \\
        Mapping Time $\downarrow$ & 845ms& 360ms & 226ms & 120ms \\
      \bottomrule
    \end{tabular}%
    \caption{\ours{} is extremely memory-efficient and runs 1.5x and 4x faster than iMAP and NICE-SLAM respectively, with even higher performance gains without the background (BG) model. }
    \label{tab:time}
    \vspace{-4mm}
\end{table}

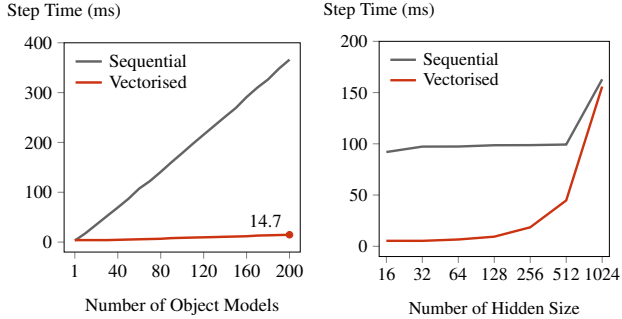
\begin{figure}[t!]
  \centering
  \resizebox{0.495\linewidth}{!}{{\begin{tikzpicture}

\definecolor{color0}{RGB}{100, 100, 100}
\definecolor{color1}{RGB}{204, 51, 17}

\begin{axis}[
width=0.4\textwidth, height=0.365\textwidth,
every axis y label/.style={at={(current axis.north west)},above=4mm},
tick align=outside,
tick pos=left,
xmin=-1, xmax=21,
ymin=-20, ymax=400,
legend cell align={left},
legend style={fill opacity=0.8, draw opacity=1, text opacity=1, draw=none, font=\large, column sep=2pt, very thick, anchor=north west, at={(0.02,0.98)}},
ylabel={Step Time (ms)},
xlabel={Number of Object Models},
xtick={0, 4, 8, 12, 16, 20},
xticklabels={1, 40, 80, 120, 160, 200},
ytick={0, 100, 200, 300, 400},
yticklabels={0, 100, 200, 300, 400},
label style={font=\large},
tick label style={font=\large},
xlabel style={yshift=-0.4cm},
]
\addplot [ultra thick, color0]
table {%
0 3.069
1 17.8644573499696
2 35.1226655499886
3 52.3330609999903
4 69.3959250000262
5 87.0228850499188
6 107.472953600063
7 122.548537400053
8 140.891750800074
9 160.154428850001
10 178.565934299968
11 197.439839649996
12 215.628281600038
13 233.648759150037
14 251.652081300017
15 269.1933393
16 291.04271289998
17 309.945945449999
18 326.426208999965
19 347.83625834998
20 366.167546750057
};
\addlegendentry{Sequential}
\addplot [ultra thick, color1]
table {%
0 3.98
1 3.98844760002248
2 3.98
3 3.98367099996904
4 4.53484544996172
5 5.10845845001313
6 5.62858385001164
7 6.13066300002174
8 6.66054040002564
9 7.9678575500111
10 8.55245929997182
11 9.08759895000912
12 9.58261550003954
13 10.0759106499936
14 10.6708340000296
15 11.1718187000406
16 11.7364633500074
17 13.0638557499879
18 13.5948568999993
19 14.1322661999766
20 14.697650749985
};
\node[fill=color1, draw=color1, scale=1.0, circle, inner sep=1.5pt, label=above left:\large 14.7] at (axis cs:20, 14.69){};
\addlegendentry{Vectorised}

\end{axis}

\end{tikzpicture}}} \hfill
  \resizebox{0.495\linewidth}{!}{{\begin{tikzpicture}

\definecolor{color0}{RGB}{100, 100, 100}
\definecolor{color1}{RGB}{204, 51, 17}

\begin{axis}[
width=0.4\textwidth, height=0.38\textwidth,
every axis y label/.style={at={(current axis.north west)},above=4mm},
tick align=outside,
tick pos=left,
xmin=-0.2, xmax=6.2,
ymin=-10, ymax=200,
legend cell align={left},
legend style={fill opacity=0.8, draw opacity=1, text opacity=1, draw=none, font=\large, column sep=2pt, very thick, anchor=north west, at={(0.02,0.98)}},
ylabel={Step Time (ms)},
xlabel={Number of Hidden Size},
xtick={0, 1, 2, 3, 4, 5, 6},
xticklabels={16, 32, 64, 128, 256, 512, 1024},
ytick={0, 50, 100, 150, 200},
yticklabels={0, 50, 100, 150, 200},
label style={font=\large},
tick label style={font=\large},
xlabel style={yshift=-0.42cm},
]
\addplot [ultra thick, color0]
table {%
0 91.9727738000802
1 97.2776642498502
2 97.348181799971
3 98.6101566499201
4 98.7197530499543
5 99.2922018000172
6 162.942676449893
};
\addlegendentry{Sequential}
\addplot [ultra thick, color1]
table {%
0 5.33503509996081
1 5.31104279998544
2 6.62859000003664
3 9.37032884994551
4 18.5283687500487
5 44.7338832000241
6 155.883042849928
};
\addlegendentry{Vectorised}
\end{axis}

\end{tikzpicture}}}
\caption{Vectorised operation allows extremely fast training speed compared to standard sequential operations using for loops. }
\vspace{-4mm}
\label{fig:ablative_vectorised}
\end{figure}

\paragraph{Vectorised v.s. Sequential Training}
We ablated training speed with vectorised and sequential operations (for loops), conditioned on different numbers of objects and different sizes of object model. In Fig.~\ref{fig:ablative_vectorised}, we can see that vectorised training enables tremendous improvements in optimisation speed, especially when we have a large number of objects.  And with vectorised training, each optimisation step takes no more than 15ms even when we train as many as 200 objects. Additionally, vectorised training is also stable across a wide range of model sizes, suggesting that we can train our object models with an even larger size if required, with minimal  additional training time. As expected, vectorised training and for loops will eventually have similar training speed, when we reach the hardware's memory limit.

To train multiple models in parallel, an initial approach we tried was  spawning a process per object. However, we were only able to spawn a very limited number of processes, due to the per process CUDA memory overhead, which significantly limited the number of objects.

\paragraph{Object Model Capacity} As vectorised training has minimal effect on training speed in terms of object model design, we also investigated how the object-level reconstruction quality is affected by different object model sizes. We experimented with different object model sizes by varying the hidden size of each MLP layer. In Fig.~\ref{fig:ablation_hiddensize}, we can see that the object-level performance starts to saturate starting from hidden size 16, with minimal or no improvement by further increasing model sizes. This indicates that object-level representation is highly compressible, and can be efficiently and accurately parameterised by very few parameters.

\paragraph{Stacked MLPs v.s. Shared MLP}
Apart from representing each object by a single individual MLP, we also explored a shared MLP design by considering multi-object mapping as a multi-task learning problem \cite{vandenhende2021mtl_survey,ruder2017mtl_survey}. Here, each object is additionally associated with a learnable latent code, and this latent code is considered as an conditional input to the network, jointly optimised with the network weights. Though we have tried multiple multi-task learning architectures \cite{misra2016cross_stitch,shikun2019mtan}, early experiments (denoted as vMAP-S in Fig.~\ref{fig:ablation_hiddensize}) showed that this shared MLP design achieved slightly degraded reconstruction quality and had no distinct training speed improvement compared to stacked MLPs, particularly when powered by vectorised training. Furthermore, we found that shared MLP design can lead to undesired training properties: i) The shared MLP needs to be optimised along with the latent codes from all the objects, since the network weights and all object codes are {\it entangled} in a shared representation space. ii) The shared MLP capacity is {\it fixed} during training, and therefore the representation space might not be sufficient with an increasing number of objects.  This accentuates the advantages of disentangled object representation space, which is a crucial design element of \ours{} system.

\section{Conclusion}
We have presented \ours{}, a real-time object-level mapping system with simple and compact neural implicit representation. By decomposing the 3D scene into meaningful instances, represented by a batch of tiny separate MLPs, the system models the 3D scene in an efficient and flexible way, enabling scene re-composition, independent tracking and continually updating of objects of interest. In addition to more accurate and compact object-centric 3D reconstruction, our system is able to predict plausible watertight surfaces for each object, even under partial occlusion.

\paragraph{Limitations and Future Work} Our current system relies on an off-the-shelf detector for instance masks, which are not necessarily spatio-temporally consistent. Though the ambiguity is partially alleviated by data association and multi-view supervision, a reasonable global constraints will be better. As objects are modelled independently, dynamic objects can be continually tracked and reconstructed to enable downstream tasks, e.g., robotic manipulation\cite{wada2020morefusion}. To extend our system to a monocular dense mapping system, depth estimation networks\cite{Yu2022MonoSDF, lyu2021hr} or more efficient neural rendering approaches\cite{muller2022instant} could be further integrated. 

\vspace{-2mm}
\section*{Acknowledgements}
Research presented in this paper has been supported by
Dyson Technology Ltd. Xin Kong holds a China Scholarship Council-Imperial Scholarship. We are very grateful
to Edgar Sucar, Binbin Xu, Hidenobu Matsuki and
Anagh Malik for fruitful discussions.

{\small
\bibliographystyle{ieee_fullname}
\bibliography{robotvision}

\begin{thebibliography}{10}\itemsep=-1pt

\bibitem{AbouChakra:etal:ARXIV2022}
Jad Abou-Chakra, Feras Dayoub, and Niko S{\"u}nderhauf.
\newblock Implicit object mapping with noisy data.
\newblock {\em arXiv preprint arXiv:2204.10516}, 2022.

\bibitem{barron2022mip}
Jonathan~T Barron, Ben Mildenhall, Dor Verbin, Pratul~P Srinivasan, and Peter
  Hedman.
\newblock Mip-nerf 360: Unbounded anti-aliased neural radiance fields.
\newblock In {\em {Proceedings of the {IEEE} Conference on Computer Vision and
  Pattern Recognition ({CVPR})}}, 2022.

\bibitem{campos2021orb}
Carlos Campos, Richard Elvira, Juan J~G{\'o}mez Rodr{\'\i}guez, Jos{\'e}~MM
  Montiel, and Juan~D Tard{\'o}s.
\newblock Orb-slam3: An accurate open-source library for visual,
  visual--inertial, and multimap slam.
\newblock {\em {{IEEE} Transactions on Robotics ({T-RO})}}, 2021.

\bibitem{dai2017scannet}
Angela Dai, Angel~X Chang, Manolis Savva, Maciej Halber, Thomas Funkhouser, and
  Matthias Nie{\ss}ner.
\newblock Scannet: Richly-annotated 3d reconstructions of indoor scenes.
\newblock In {\em {Proceedings of the {IEEE} Conference on Computer Vision and
  Pattern Recognition ({CVPR})}}, 2017.

\bibitem{dai2017bundlefusion}
Angela Dai, Matthias Nie{\ss}ner, Michael Zollh{\"o}fer, Shahram Izadi, and
  Christian Theobalt.
\newblock Bundlefusion: Real-time globally consistent 3d reconstruction using
  on-the-fly surface reintegration.
\newblock {\em ACM Transactions on Graphics (ToG)}, 2017.

\bibitem{Davies:etal:ICML2021}
Thomas Davies, Derek Nowrouzezahrai, and Alec Jacobson.
\newblock On the effectiveness of weight-encoded neural implicit 3d shapes.
\newblock In {\em {Proceedings of the International Conference on Machine
  Learning ({ICML})}}, 2021.

\bibitem{Endres:etal:ICRA2012}
Felix Endres, J{\"u}rgen Hess, Nikolas Engelhard, J{\"u}rgen Sturm, Daniel
  Cremers, and Wolfram Burgard.
\newblock {An Evaluation of the {RGB-D SLAM} System}.
\newblock In {\em {Proceedings of the {IEEE} International Conference on
  Robotics and Automation ({ICRA})}}, 2012.

\bibitem{gupta2019lvis}
Agrim Gupta, Piotr Dollar, and Ross Girshick.
\newblock Lvis: A dataset for large vocabulary instance segmentation.
\newblock In {\em {Proceedings of the {IEEE} Conference on Computer Vision and
  Pattern Recognition ({CVPR})}}, 2019.

\bibitem{functorch2021}
He Horace and Zou Richard.
\newblock functorch: Jax-like composable function transforms for pytorch.
\newblock \url{https://github.com/pytorch/functorch}, 2021.

\bibitem{Jang:etal:ICCV2021}
Wonbong Jang and Lourdes Agapito.
\newblock Codenerf: Disentangled neural radiance fields for object categories.
\newblock In {\em {Proceedings of the International Conference on Computer
  Vision ({ICCV})}}, 2021.

\bibitem{kong2020semantic}
Xin Kong, Xuemeng Yang, Guangyao Zhai, Xiangrui Zhao, Xianfang Zeng, Mengmeng
  Wang, Yong Liu, Wanlong Li, and Feng Wen.
\newblock Semantic graph based place recognition for 3d point clouds.
\newblock In {\em {Proceedings of the {IEEE/RSJ} Conference on Intelligent
  Robots and Systems ({IROS})}}, 2020.

\bibitem{li2022generative}
Guanglin Li, Yifeng Li, Zhichao Ye, Qihang Zhang, Tao Kong, Zhaopeng Cui, and
  Guofeng Zhang.
\newblock Generative category-level shape and pose estimation with semantic
  primitives.
\newblock In {\em {Conference on Robot Learning ({CoRL})}}, 2022.

\bibitem{shikun2019mtan}
Shikun Liu, Edward Johns, and Andrew~J Davison.
\newblock End-to-end multi-task learning with attention.
\newblock In {\em {Proceedings of the {IEEE} Conference on Computer Vision and
  Pattern Recognition ({CVPR})}}, 2019.

\bibitem{lyu2021hr}
Xiaoyang Lyu, Liang Liu, Mengmeng Wang, Xin Kong, Lina Liu, Yong Liu, Xinxin
  Chen, and Yi Yuan.
\newblock Hr-depth: High resolution self-supervised monocular depth estimation.
\newblock In {\em {Proceedings of the National Conference on Artificial
  Intelligence ({AAAI})}}, 2021.

\bibitem{majercik2018ray}
Alexander Majercik, Cyril Crassin, Peter Shirley, and Morgan McGuire.
\newblock A ray-box intersection algorithm and efficient dynamic voxel
  rendering.
\newblock {\em Journal of Computer Graphics Techniques (JCGT)}, 2018.

\bibitem{McCormac:etal:3DV2018}
John McCormac, Ronald Clark, Michael Bloesch, Andrew Davison, and Stefan
  Leutenegger.
\newblock Fusion++: Volumetric object-level slam.
\newblock In {\em {Proceedings of the International Conference on 3D Vision
  ({3DV})}}, 2018.

\bibitem{mescheder2019occupancy}
Lars Mescheder, Michael Oechsle, Michael Niemeyer, Sebastian Nowozin, and
  Andreas Geiger.
\newblock Occupancy networks: Learning 3d reconstruction in function space.
\newblock In {\em {Proceedings of the {IEEE} Conference on Computer Vision and
  Pattern Recognition ({CVPR})}}, 2019.

\bibitem{Mildenhall:etal:ECCV2020}
Ben Mildenhall, Pratul~P. Srinivasan, Matthew Tancik, Jonathan~T. Barron, Ravi
  Ramamoorthi, and Ren Ng.
\newblock {NeRF}: Representing scenes as neural radiance fields for view
  synthesis.
\newblock In {\em {Proceedings of the European Conference on Computer Vision
  ({ECCV})}}, 2020.

\bibitem{misra2016cross_stitch}
Ishan Misra, Abhinav Shrivastava, Abhinav Gupta, and Martial Hebert.
\newblock Cross-stitch networks for multi-task learning.
\newblock In {\em {Proceedings of the {IEEE} Conference on Computer Vision and
  Pattern Recognition ({CVPR})}}, 2016.

\bibitem{muller2022instant}
Thomas M{\"u}ller, Alex Evans, Christoph Schied, and Alexander Keller.
\newblock Instant neural graphics primitives with a multiresolution hash
  encoding.
\newblock {\em ACM Transactions on Graphics (ToG)}, 2022.

\bibitem{niemeyer2021giraffe}
Michael Niemeyer and Andreas Geiger.
\newblock Giraffe: Representing scenes as compositional generative neural
  feature fields.
\newblock In {\em {Proceedings of the {IEEE} Conference on Computer Vision and
  Pattern Recognition ({CVPR})}}, 2021.

\bibitem{oechsle2021unisurf}
Michael Oechsle, Songyou Peng, and Andreas Geiger.
\newblock Unisurf: Unifying neural implicit surfaces and radiance fields for
  multi-view reconstruction.
\newblock In {\em {Proceedings of the International Conference on Computer
  Vision ({ICCV})}}, 2021.

\bibitem{park2019deepsdf}
Jeong~Joon Park, Peter Florence, Julian Straub, Richard Newcombe, and Steven
  Lovegrove.
\newblock {DeepSDF}: Learning continuous signed distance functions for shape
  representation.
\newblock In {\em {Proceedings of the {IEEE} Conference on Computer Vision and
  Pattern Recognition ({CVPR})}}, 2019.

\bibitem{rebain2021derf}
Daniel Rebain, Wei Jiang, Soroosh Yazdani, Ke Li, Kwang~Moo Yi, and Andrea
  Tagliasacchi.
\newblock Derf: Decomposed radiance fields.
\newblock In {\em {Proceedings of the {IEEE} Conference on Computer Vision and
  Pattern Recognition ({CVPR})}}, 2021.

\bibitem{reiser2021kilonerf}
Christian Reiser, Songyou Peng, Yiyi Liao, and Andreas Geiger.
\newblock Kilonerf: Speeding up neural radiance fields with thousands of tiny
  mlps.
\newblock In {\em {Proceedings of the International Conference on Computer
  Vision ({ICCV})}}, 2021.

\bibitem{rosinol2022nerf}
Antoni Rosinol, John~J Leonard, and Luca Carlone.
\newblock Nerf-slam: Real-time dense monocular slam with neural radiance
  fields.
\newblock {\em arXiv preprint arXiv:2210.13641}, 2022.

\bibitem{ruder2017mtl_survey}
Sebastian Ruder.
\newblock An overview of multi-task learning in deep neural networks.
\newblock {\em arXiv preprint arXiv:1706.05098}, 2017.

\bibitem{Runz::Agapito::ICRA2017}
Martin R{\"u}nz and Lourdes Agapito.
\newblock Co-fusion: Real-time segmentation, tracking and fusion of multiple
  objects.
\newblock In {\em {Proceedings of the {IEEE} International Conference on
  Robotics and Automation ({ICRA})}}, 2017.

\bibitem{Salas-Moreno:etal:CVPR2013}
Renato~F Salas-Moreno, Richard~A Newcombe, Hauke Strasdat, Paul~HJ Kelly, and
  Andrew~J Davison.
\newblock {{SLAM++}: Simultaneous Localisation and Mapping at the Level of
  Objects}.
\newblock In {\em {Proceedings of the {IEEE} Conference on Computer Vision and
  Pattern Recognition ({CVPR})}}, 2013.

\bibitem{straub2019replica}
Julian Straub, Thomas Whelan, Lingni Ma, Yufan Chen, Erik Wijmans, Simon Green,
  Jakob~J Engel, Raul Mur-Artal, Carl Ren, Shobhit Verma, et~al.
\newblock The replica dataset: A digital replica of indoor spaces.
\newblock {\em arXiv preprint arXiv:1906.05797}, 2019.

\bibitem{Sturm:etal:IROS2012}
J. Sturm, N. Engelhard, F. Endres, W. Burgard, and D. Cremers.
\newblock {A Benchmark for the Evaluation of {RGB-D} {SLAM} Systems}.
\newblock In {\em {Proceedings of the {IEEE/RSJ} Conference on Intelligent
  Robots and Systems ({IROS})}}, 2012.

\bibitem{Sucar:etal:ICCV2021}
Edgar Sucar, Shikun Liu, Joseph Ortiz, and Andrew~J. Davison.
\newblock imap: Implicit mapping and positioning in real-time.
\newblock In {\em {Proceedings of the International Conference on Computer
  Vision ({ICCV})}}, 2021.

\bibitem{Sucar:etal:3DV2020}
Edgar Sucar, Kentaro Wada, and Andrew Davison.
\newblock {NodeSLAM}: Neural object descriptors for multi-view shape
  reconstruction.
\newblock In {\em {Proceedings of the International Conference on 3D Vision
  ({3DV})}}, 2020.

\bibitem{vandenhende2021mtl_survey}
Simon Vandenhende, Stamatios Georgoulis, Wouter Van~Gansbeke, Marc Proesmans,
  Dengxin Dai, and Luc Van~Gool.
\newblock Multi-task learning for dense prediction tasks: A survey.
\newblock {\em {{IEEE} Transactions on Pattern Analysis and Machine
  Intelligence ({PAMI})}}, 2021.

\bibitem{wada2020morefusion}
Kentaro Wada, Edgar Sucar, Stephen James, Daniel Lenton, and Andrew~J Davison.
\newblock Morefusion: Multi-object reasoning for 6d pose estimation from
  volumetric fusion.
\newblock In {\em {Proceedings of the {IEEE} Conference on Computer Vision and
  Pattern Recognition ({CVPR})}}, 2020.

\bibitem{wang2022go}
Jingwen Wang, Tymoteusz Bleja, and Lourdes Agapito.
\newblock Go-surf: Neural feature grid optimization for fast, high-fidelity
  rgb-d surface reconstruction.
\newblock In {\em {Proceedings of the International Conference on 3D Vision
  ({3DV})}}, 2022.

\bibitem{wang2021dsp}
Jingwen Wang, Martin R{\"u}nz, and Lourdes Agapito.
\newblock Dsp-slam: object oriented slam with deep shape priors.
\newblock In {\em 2021 International Conference on 3D Vision (3DV)}, 2021.

\bibitem{wu2022object}
Qianyi Wu, Xian Liu, Yuedong Chen, Kejie Li, Chuanxia Zheng, Jianfei Cai, and
  Jianmin Zheng.
\newblock Object-compositional neural implicit surfaces.
\newblock In {\em {Proceedings of the European Conference on Computer Vision
  ({ECCV})}}, 2022.

\bibitem{Xu:etal:ICRA2019}
Binbin Xu, Wenbin Li, Dimos Tzoumanikas, Michael Bloesch, Andrew Davison, and
  Stefan Leutenegger.
\newblock {MID-Fusion}: Octree-based object-level multi-instance dynamic slam.
\newblock In {\em {Proceedings of the {IEEE} International Conference on
  Robotics and Automation ({ICRA})}}, 2019.

\bibitem{Yang:etal:ICCV2021}
Bangbang Yang, Yinda Zhang, Yinghao Xu, Yijin Li, Han Zhou, Hujun Bao, Guofeng
  Zhang, and Zhaopeng Cui.
\newblock Learning object-compositional neural radiance field for editable
  scene rendering.
\newblock In {\em {Proceedings of the International Conference on Computer
  Vision ({ICCV})}}, 2021.

\bibitem{yang2022voxfusion}
Xingrui Yang, Hai Li, Hongjia Zhai, Yuhang Ming, Yuqian Liu, and Guofeng Zhang.
\newblock {Vox-Fusion}: Dense tracking and mapping with voxel-based neural
  implicit representation.
\newblock In {\em {Proceedings of the International Symposium on Mixed and
  Augmented Reality ({ISMAR})}}, 2022.

\bibitem{Yu2022MonoSDF}
Zehao Yu, Songyou Peng, Michael Niemeyer, Torsten Sattler, and Andreas Geiger.
\newblock Monosdf: Exploring monocular geometric cues for neural implicit
  surface reconstruction.
\newblock {\em Advances in Neural Information Processing Systems (NeurIPS)},
  2022.

\bibitem{yuan2022nerf}
Yu-Jie Yuan, Yang-Tian Sun, Yu-Kun Lai, Yuewen Ma, Rongfei Jia, and Lin Gao.
\newblock Nerf-editing: geometry editing of neural radiance fields.
\newblock In {\em {Proceedings of the {IEEE} Conference on Computer Vision and
  Pattern Recognition ({CVPR})}}, 2022.

\bibitem{Zhi:etal:CVPR2019}
Shuaifeng Zhi, Michael Bloesch, Stefan Leutenegger, and Andrew~J Davison.
\newblock {SceneCode}: Monocular dense semantic reconstruction using learned
  encoded scene representations.
\newblock In {\em {Proceedings of the {IEEE} Conference on Computer Vision and
  Pattern Recognition ({CVPR})}}, 2019.

\bibitem{zhi2022ilabel}
Shuaifeng Zhi, Edgar Sucar, Andre Mouton, Iain Haughton, Tristan Laidlow, and
  Andrew~J Davison.
\newblock ilabel: Revealing objects in neural fields.
\newblock {\em {{IEEE} Robotics and Automation Letters ({RA-L})}}, 2022.

\bibitem{zhong2023icra}
Xingguang Zhong, Yue Pan, Jens Behley, and Cyrill Stachniss.
\newblock Shine-mapping: Large-scale 3d mapping using sparse hierarchical
  implicit neural representations.
\newblock In {\em Proceedings of the IEEE International Conference on Robotics
  and Automation (ICRA)}, 2023.

\bibitem{zhou2013dense}
Qian-Yi Zhou and Vladlen Koltun.
\newblock Dense scene reconstruction with points of interest.
\newblock {\em ACM Transactions on Graphics (ToG)}, 2013.

\bibitem{zhou2022detic}
Xingyi Zhou, Rohit Girdhar, Armand Joulin, Philipp Kr{\"a}henb{\"u}hl, and
  Ishan Misra.
\newblock Detecting twenty-thousand classes using image-level supervision.
\newblock In {\em {Proceedings of the European Conference on Computer Vision
  ({ECCV})}}, 2022.

\bibitem{zhu2022nice}
Zihan Zhu, Songyou Peng, Viktor Larsson, Weiwei Xu, Hujun Bao, Zhaopeng Cui,
  Martin~R Oswald, and Marc Pollefeys.
\newblock Nice-slam: Neural implicit scalable encoding for slam.
\newblock In {\em {Proceedings of the {IEEE} Conference on Computer Vision and
  Pattern Recognition ({CVPR})}}, 2022.

\end{thebibliography}
}

\clearpage
\onecolumn

\setcounter{section}{0}
\setcounter{figure}{0}
\setcounter{table}{0}
\renewcommand\thesection{\Alph{section}}
\renewcommand\thetable{\Alph{table}}
\renewcommand\thefigure{\Alph{figure}}

\section{Interactive Visualisation}
We recommend readers to check out our project website \url{https://kxhit.github.io/vMAP}, showing the real-time scene-level and object-level reconstructions of some selected sequences.

\section{Implement Details and Discussions}\label{sec:discussions}

\begin{figure}[!h]
    \centering
    \includegraphics[trim={4cm 10cm 12cm 6cm}, clip, width=0.8\textwidth]{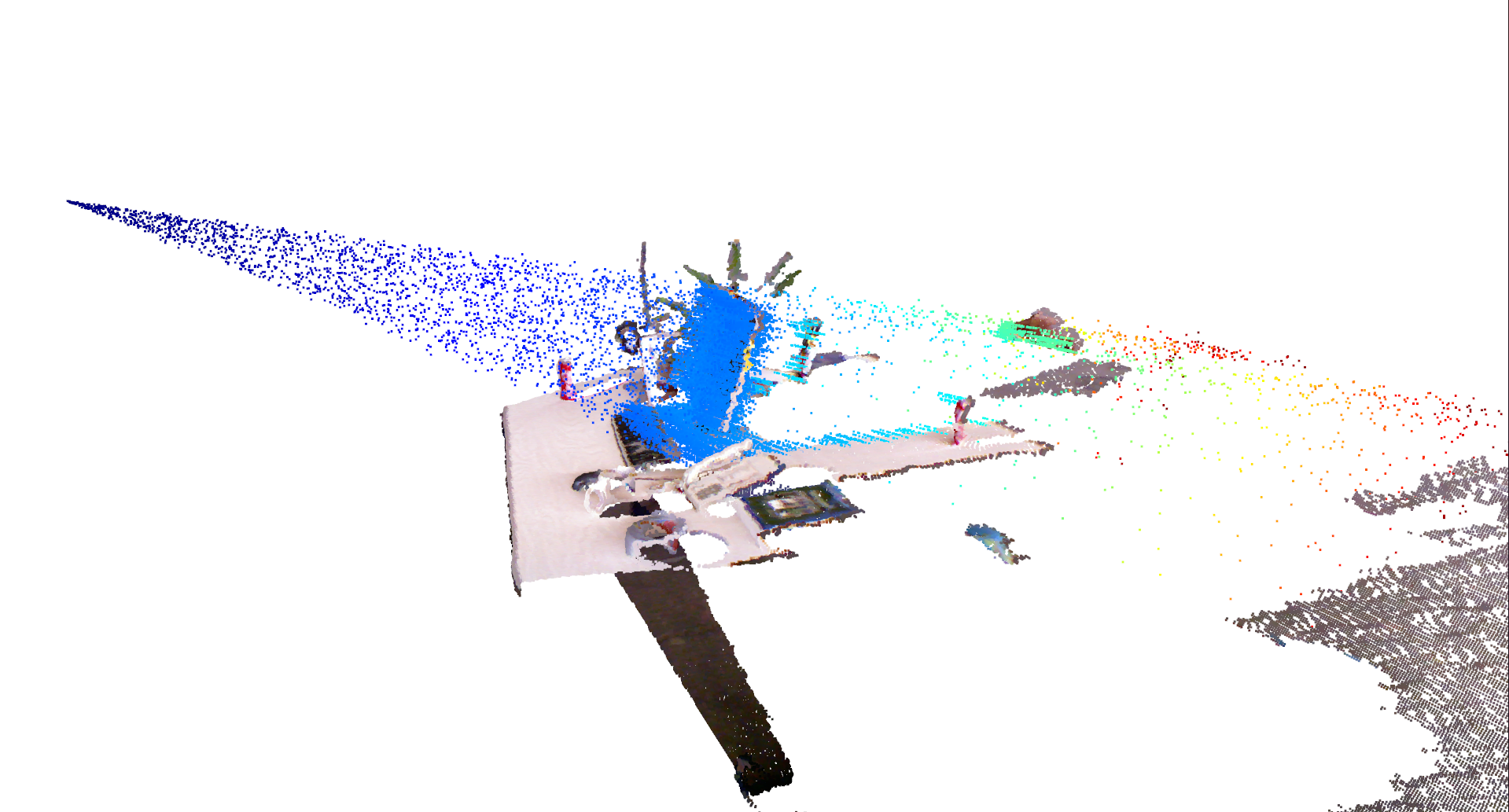}
    \caption{Visualisation of depth guided sampling.}
    \label{fig:sampling}
\end{figure}

\paragraph{Depth-Guided Sampling}
As described in the main paper, we sampled more points near the object surface guided by the depth measurements. For rays that go through the 3D object bounding box but do not belong to the current instance, we then terminate these rays when they hit the object surface, to minimise the impact on the occluded objects, similar to ObjectNeRF~\cite{Yang:etal:ICCV2021}. A visualisation of depth guided sampling is shown in Fig.~\ref{fig:sampling}, and the sampled points are coloured by the measured depth.

\paragraph{Object-Level Positional Encoding}

Since object instances are different in size, the reconstruction quality can be maximised when trained with a suitable positional encoding frequency. Otherwise, the network training would be biased towards reconstructing large objects and overlook small objects or vice versa. To mitigate this scaling issue, we applied integrated positional encoding \cite{barron2022mip} and introduced an additional hyper-parameter, the scaling factor $s$, which is applied to all objects, such that they are bounded in a unit box within the range of $[-1, 1]$.  We separately set this scaling factor slightly larger in the background model. 

\begin{figure*}[ht!]
  \centering
  \small
  \setlength{\tabcolsep}{0.1em}
  \begin{tabular}{C{0.06\linewidth}C{0.32\linewidth}C{0.32\linewidth}C{0.32\linewidth}}
  & $s=5$  & $s=10$  & $s=15$ \\
    Global View &
    \includegraphics[width=\linewidth]{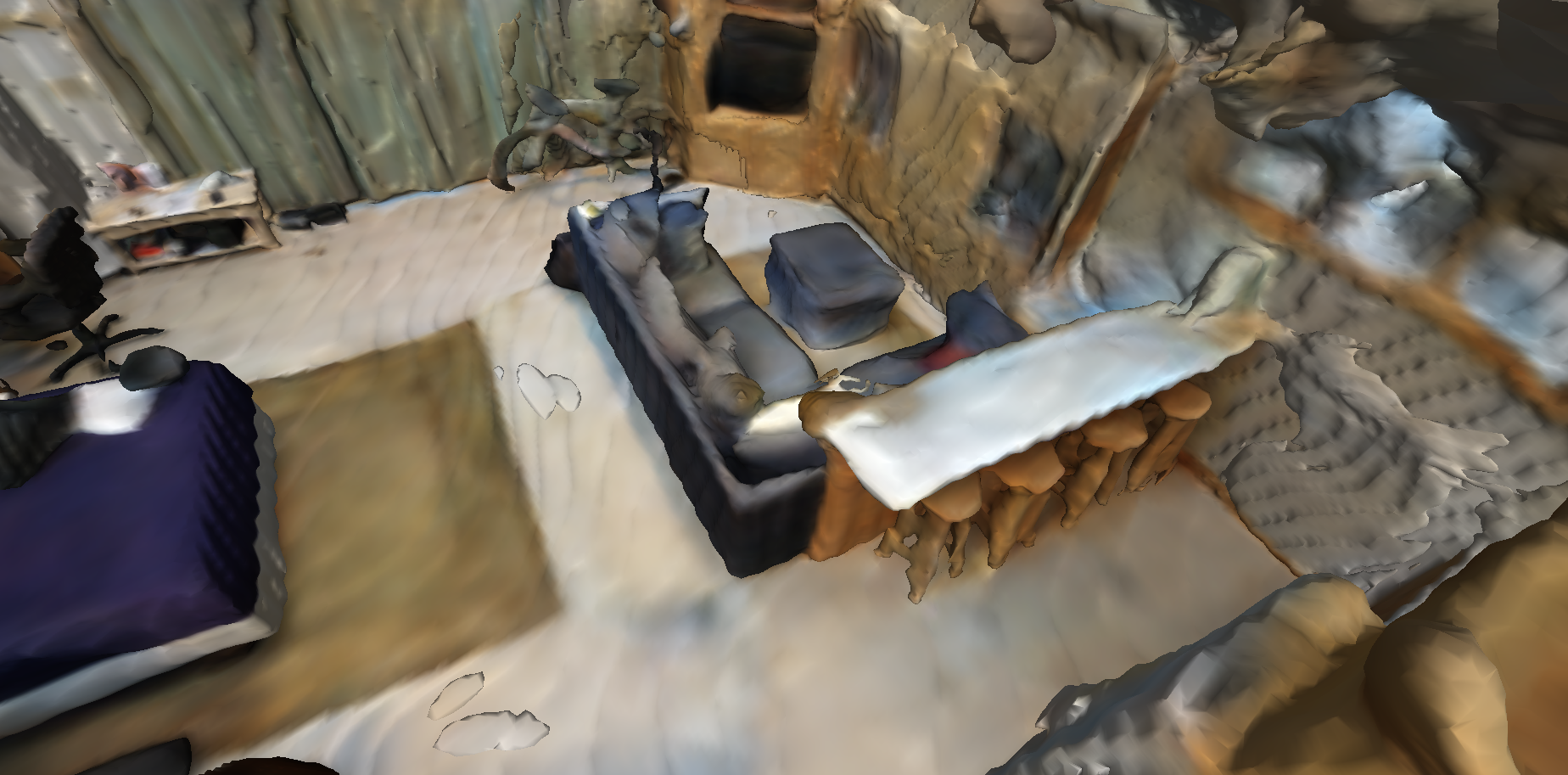}&
    \includegraphics[width=\linewidth]{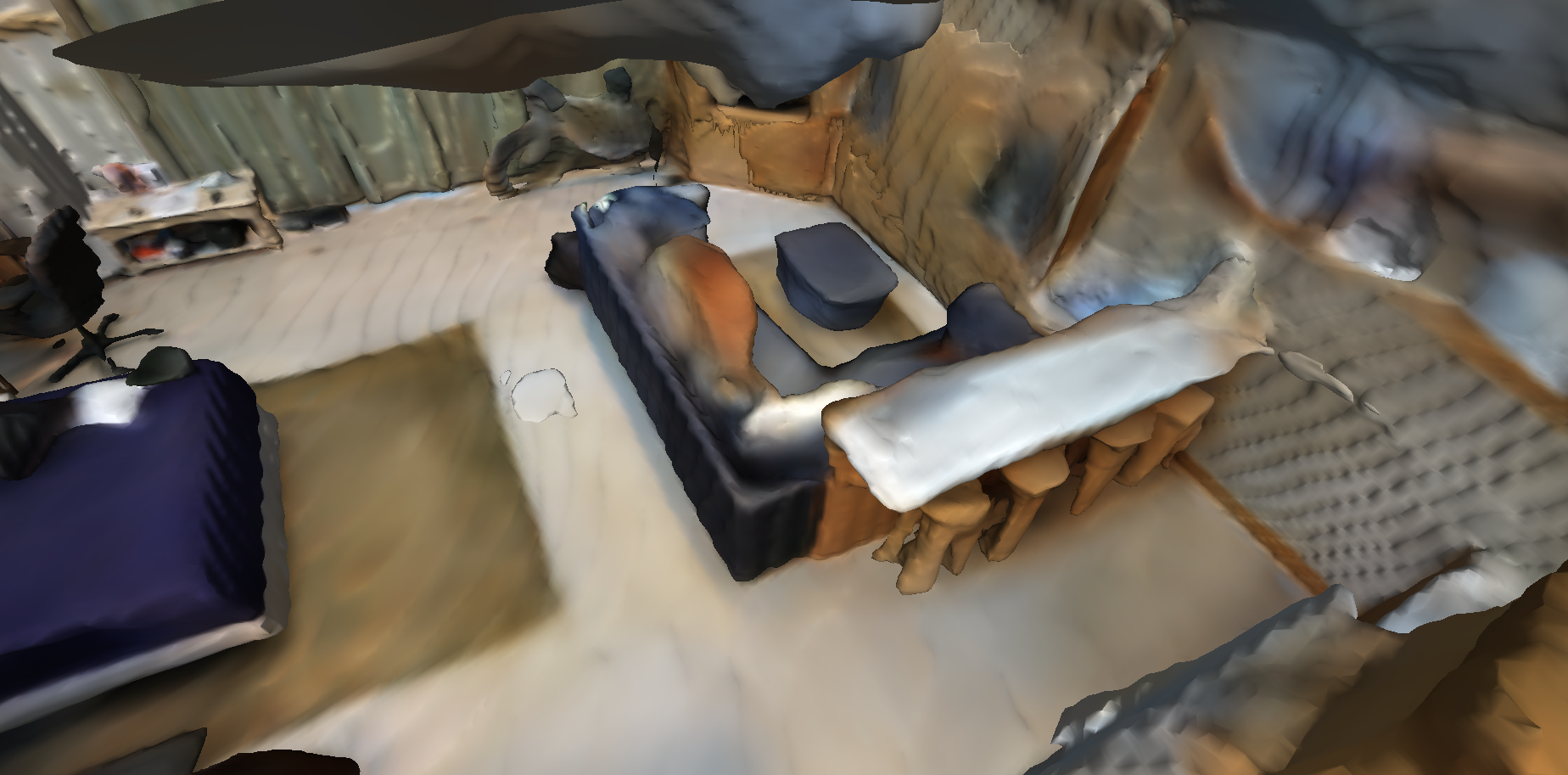}&
    \includegraphics[width=\linewidth]{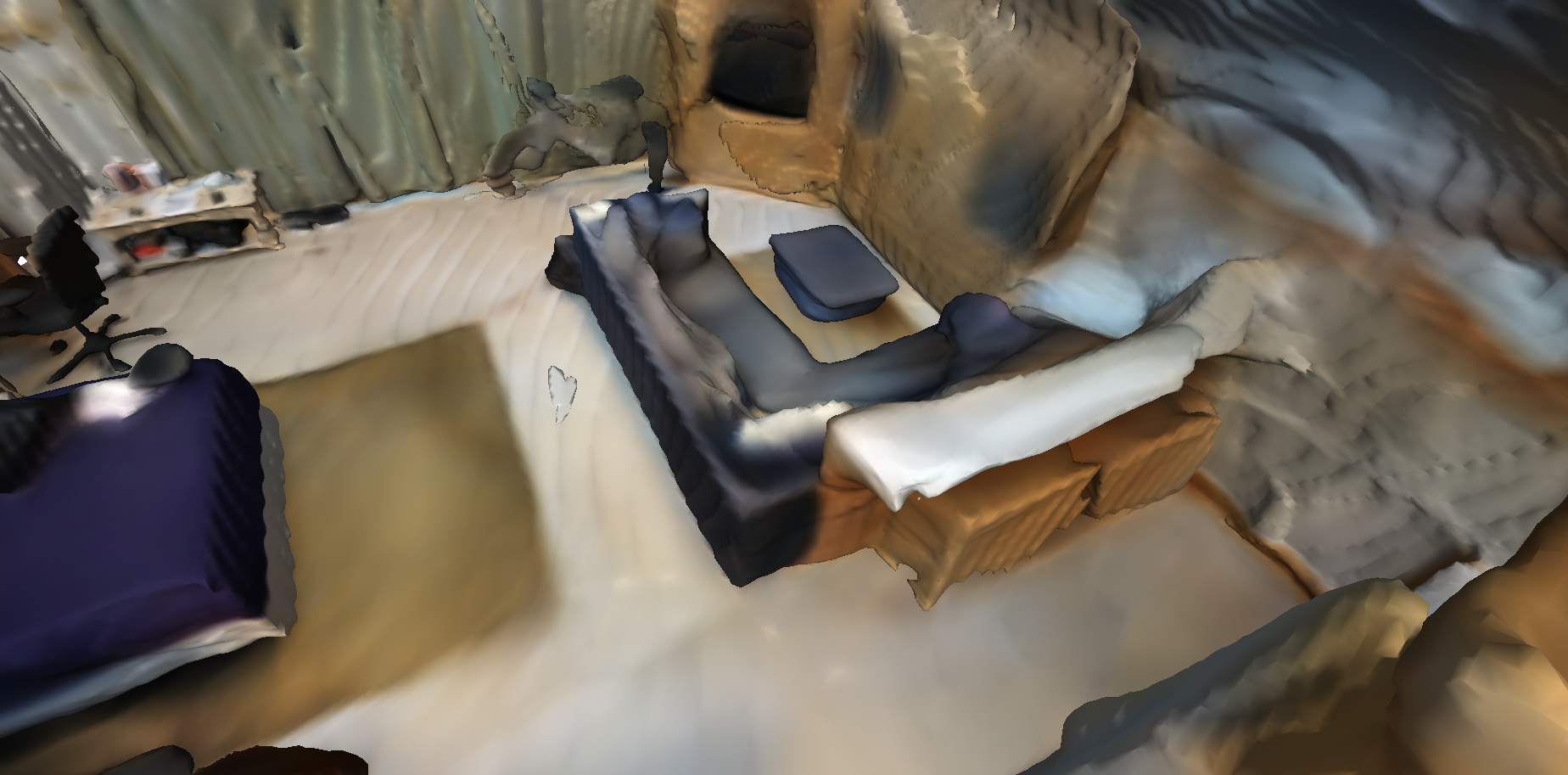} \\
    Local View &
    \includegraphics[width=\linewidth]{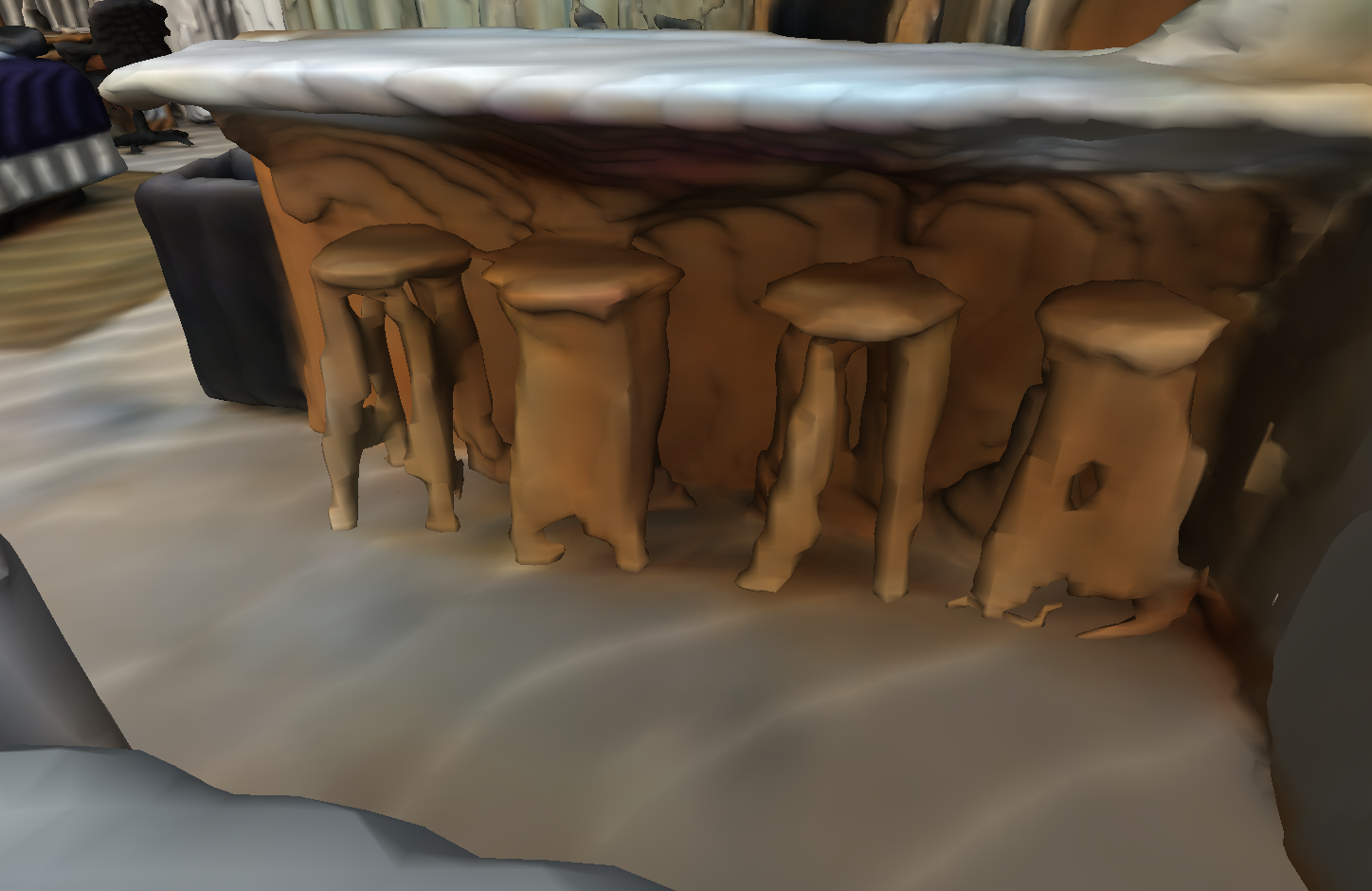}&
    \includegraphics[width=\linewidth]{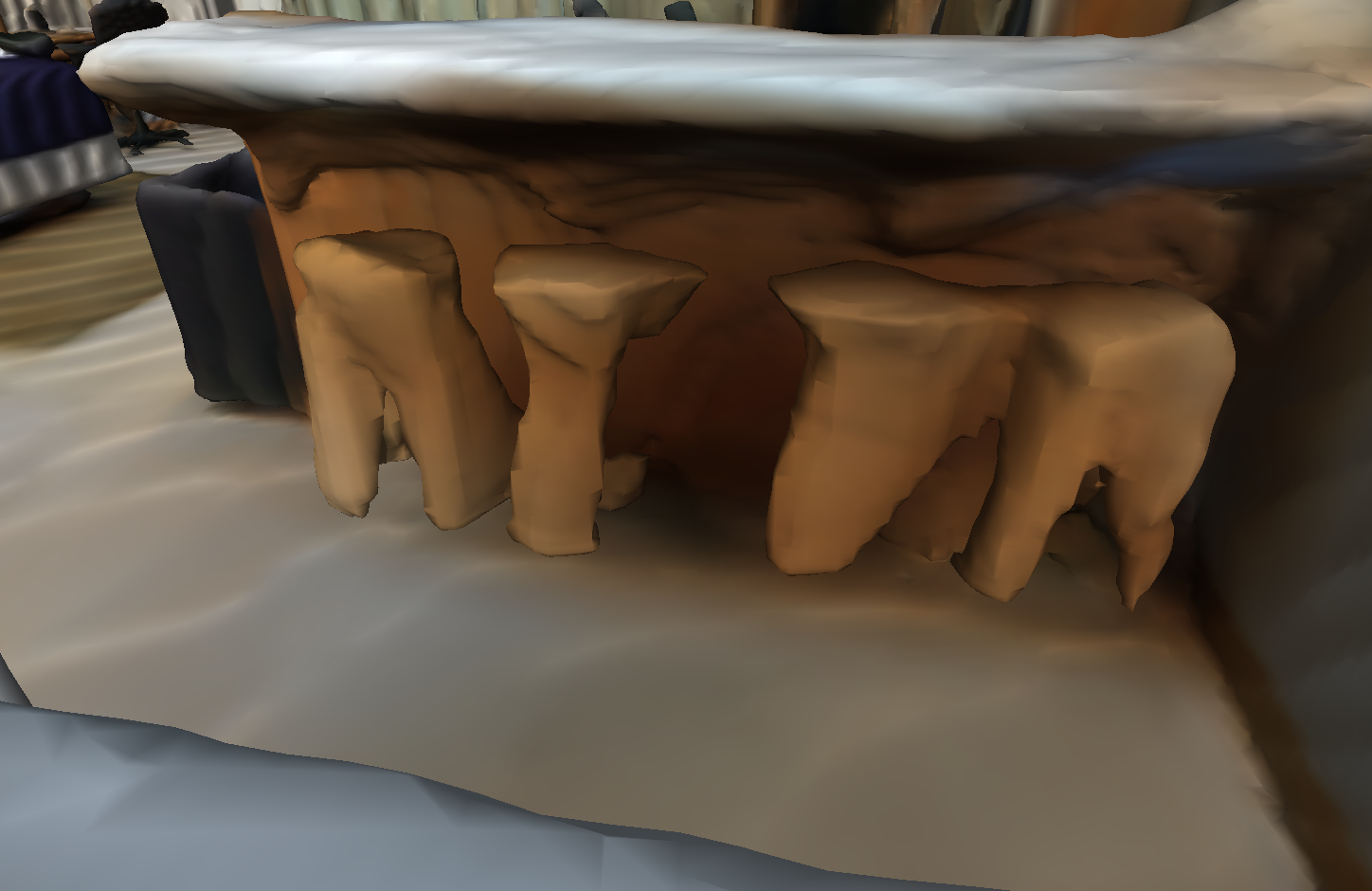}&
    \includegraphics[width=\linewidth]{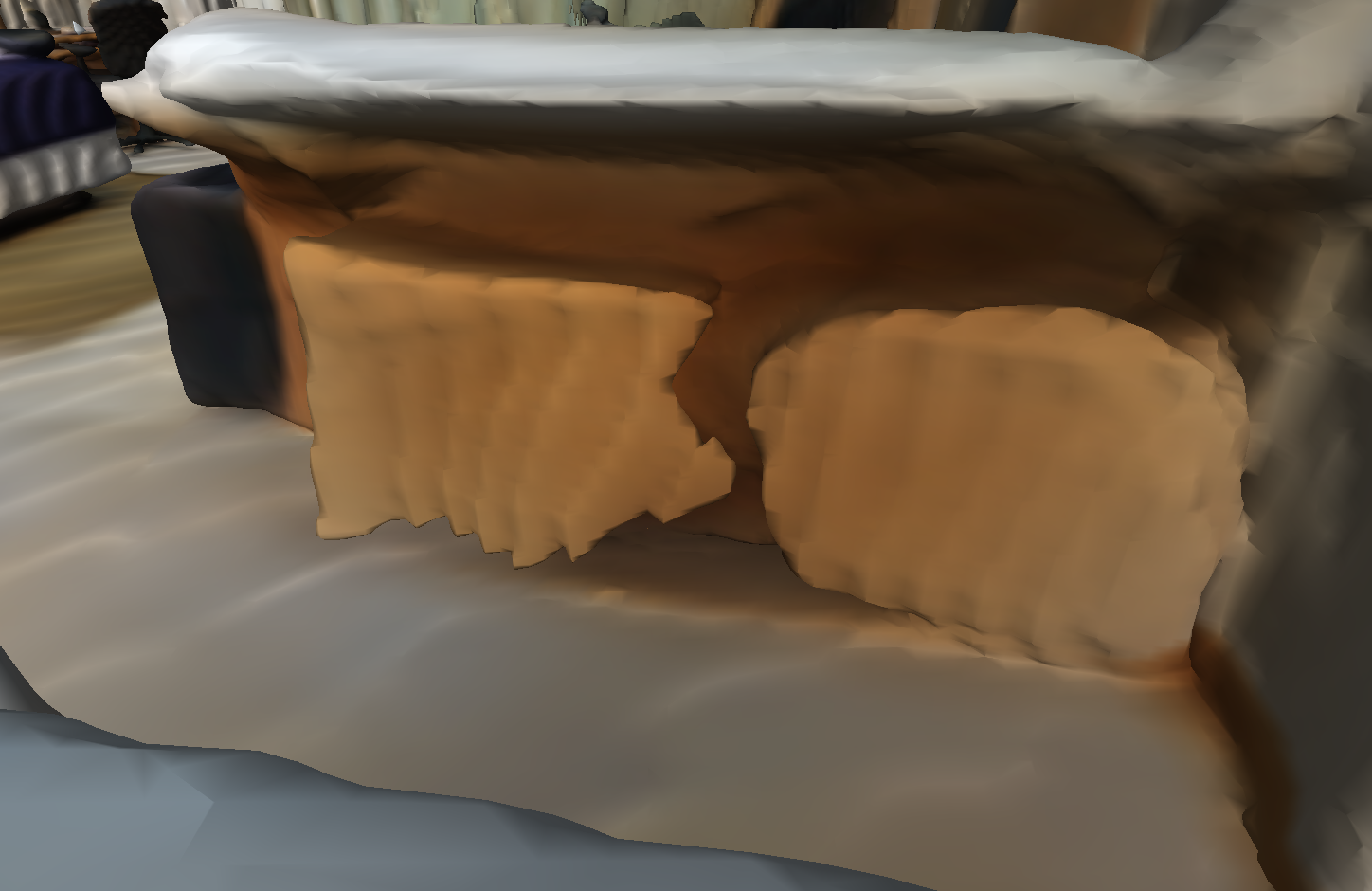}
  \end{tabular}
  \caption{Visualisation of 3D object reconstructions trained with different scales.}
  \label{fig:scale_freq}
\end{figure*}

This scaling factor can be set as object specific if such object-specific prior is known, i.e. we can set a large $s$ when training the object `sofa‘, and a small $s$ when training the object `cup’, because a sofa is typically larger than a cup. A visualisation of the object reconstruction with different choices of $s$ is shown in Fig.~\ref{fig:scale_freq}. We can see a large scale $s$ results a smoother geometry which is more suitable for reconstructing large objects like `walls' and `blankets', and a small $s$ is more suitable for objects with complex geometries like `chairs'.

\section{Additional Experimental Results on Replica Scenes}
\label{sec:results}

In Tab.~\ref{tab:replica_scene_3D} and Tab.~\ref{tab:replica_obj_3D}, we listed the detailed scene-level and object-level 3D reconstruction results for each sequence on Replica dataset.

\begin{table}[!ht]
  \centering
  \footnotesize
  \setlength{\tabcolsep}{0.2em}    
    \begin{tabular}{clccccccccc}
      \toprule
         & & \tt{room-0} & \tt{room-1} & \tt{room-2}  & \tt{office-0} &  \tt{office-1} & \tt{office-2} & \tt{office-3} & \tt{office-4} & Avg. \\
        \midrule
      \multirow{3}{*}{\makecell{\textbf{TSDF-Fusion}$^\ast$ \\ }}  & {\bf Acc.} [cm] $\downarrow$  
      & 1.46 & 1.13 & 1.22 & 1.13 & 0.91 & 1.33 & 1.56 & 1.48 & 1.28 \\
      & {\bf Comp.} [cm] 
      & 3.73 & 3.51 & 4.41 & 10.26 & 9.57 & 5.50 & 3.87 & 4.04 & 5.61 \\
      & {\bf Comp. Ratio} [$<$ 5cm \%] 
      & 86.54 & 87.12 & 84.87 & 78.86 & 75.85 & 80.48 & 83.19 & 84.41 & 82.67 \\
      \midrule
      \multirow{3}{*}{\makecell{\textbf{iMAP}}}    
          & {\bf Acc.} [cm] $\downarrow$
          & 3.58 & 3.69 & 4.68 & 5.87 & 3.71 & 4.81 & 4.27 & 4.83 & 4.43 \\
          & {\bf Comp.} [cm] $\downarrow$
          & 5.06 & 4.87 & 5.51 & 6.11 & 5.26  & 5.65 & 5.45 & 6.59 &  5.56\\
          & {\bf Comp. Ratio} [$<$ 5cm \%] $\uparrow$
          & 83.91 & 83.45 & 75.53 & 77.71& 79.64 & 77.22& 77.34 & 77.63 & 79.06\\
      \midrule
            \multirow{3}{*}{\makecell{\textbf{iMAP}$^\ast$}} & {\bf Acc.} [cm] $\downarrow$ 
 
          & 2.06 & 1.65 & 1.92 & 2.36 & 1.94 & 2.61 & 2.41 & 2.23 & 2.15 \\
          & {\bf Comp. } [cm] $\downarrow$ 
          & 2.21 & 1.94 & 2.77 & 4.81 & 3.19 & 2.81 & 2.78 & 2.56 & 2.88 \\
          & {\bf Comp. Ratio} [$<$ 5cm \%] $\uparrow$ 
          & 94.93&94.87 &90.78 & 86.85 &87.79&89.61 & 90.54 & 91.46 & 90.85 \\
    \midrule
      \multirow{3}{*}{\makecell{\textbf{NICE-SLAM}}} & {\bf Acc.} [cm] $\downarrow$ 
 
          & 2.69 & 2.49 & 2.55 & 3.03 & 3.31 & 3.56     & 3.26 & 2.63 & 2.94 \\
          & {\bf Comp. } [cm] $\downarrow$ 
          & 2.92 & 2.33 & 2.96 & 8.34 & 5.18  & 3.35     & 3.37 & 3.68 & 4.02 \\
          & {\bf Comp. Ratio} [$<$ 5cm \%] $\uparrow$ 
          & 90.77 & 93.07 & 87.83 & 81.99 & 82.24 & 85.82 & 85.44 & 86.64 & 86.73 \\
      \midrule  
      \multirow{3}{*}{\makecell{\textbf{NICE-SLAM$^\ast$}}} & {\bf Acc.} [cm] $\downarrow$ 
 
          & 2.71 & 2.28 & 2.69 & 2.93 & 4.23        & 3.45 & 3.26 & 2.74 & 3.04 \\
          & {\bf Comp. } [cm] $\downarrow$ 
          & 2.84 & 2.23 & 3.02 & 7.54 & 4.52       & 3.31 & 3.58 & 3.64 & 3.84 \\
          & {\bf Comp. Ratio} [$<$ 5cm \%] $\uparrow$ 
          &  91.00 & 93.37 & 87.23 & 82.70 & 82.09  & 85.42 & 84.28 & 86.10 & 86.52\\
      \midrule
      \multirow{3}{*}{\makecell{\textbf{\ours{}}}} & {\bf Acc.} [cm] $\downarrow$ 
 
          & 2.77 & 3.87 & 1.83 & 4.82 & 3.51 & 3.35 & 3.19 & 2.26 & 3.20 \\
          & {\bf Comp. } [cm] $\downarrow$ 
          & 1.99 & 1.81 & 2.00 & 3.65 & 2.14 & 2.45 & 2.49 & 2.56 & \textbf{2.39} \\
          & {\bf Comp. Ratio} [$<$ 5cm \%] $\uparrow$ 
          & 97.10& 96.59& 95.72& 87.53& 85.08& 94.70&93.65 & 93.56 & \textbf{92.99} \\
      \bottomrule
    \end{tabular}%
    \caption{Scene-level reconstruction results on 8 indoor Replica scenes.  $\ast$ represents the baselines we re-trained with ground-truth pose.}
    \label{tab:replica_scene_3D}
\end{table}

\newpage
\begin{table}[!th]
  \centering
  \footnotesize
  \setlength{\tabcolsep}{0.2em}    %
    \begin{tabular}{clccccccccc}
      \toprule
         & & \tt{room-0} & \tt{room-1} & \tt{room-2}  & \tt{office-0} &  \tt{office-1} & \tt{office-2} & \tt{office-3} & \tt{office-4} & Avg. \\
        \midrule
      \multirow{4}{*}{\makecell{\textbf{TSDF-Fusion}$^\ast$}}  & {\bf Acc.} [cm]  $\downarrow$ 
          & 0.43 & 0.45 & 0.45 & 0.49   & 0.43 & 0.41 & 0.45 & 0.52 & \textbf{0.45} \\
          & {\bf Comp. } [cm] $\downarrow$ 
          & 3.03 & 4.42 & 4.16 & 2.23   & 5.60 & 3.32 & 3.31 & 3.48 & 3.69 \\
          & {\bf Comp. Ratio} [$<$ 1cm \%] $\uparrow$ 
          & 62.27& 59.44& 53.57 &68.16  &55.73 & 67.34& 63.35& 63.75& 61.70 \\
          & {\bf Comp. Ratio} [$<$ 5cm \%] $\uparrow$ 
          & 84.34& 79.12& 76.89&86.74   &80.36 & 87.30& 85.74& 83.38 & 82.98 \\
      \midrule
      \multirow{4}{*}{\makecell{\textbf{NICE-SLAM$^\ast$}}}  & {\bf Acc.} [cm] $\downarrow$  
      & 3.48 & 3.77 & 4.61 & 4.08   & 3.42 & 3.45 & 3.96 & 4.53 & 3.91 \\
      & {\bf Comp. } [cm] $\downarrow$ 
      & 2.51 & 2.82 & 3.19 & 3.05   & 3.29 & 3.47 & 3.61 & 4.23 & 3.27 \\
      & {\bf Comp. Ratio} [$<$ 1cm \%] $\uparrow$ 
      & 41.19& 37.06& 33.03 &38.86  &44.55 &41.84 &31.21 &34.54 & 37.79 \\
      & {\bf Comp. Ratio} [$<$ 5cm \%] $\uparrow$ 
      & 86.88& 86.43& 83.96 &84.54  &89.08 &83.77 &79.40 &77.64 & 83.96 \\
      \midrule
      \multirow{4}{*}{\makecell{\textbf{iMAP$^\ast$}}} & {\bf Acc.} [cm] $\downarrow$ 
          & 3.02 & 3.35 & 4.50 & 3.84 & 2.62 & 3.22 & 3.58 & 4.43  & 3.57 \\
          & {\bf Comp. } [cm] $\downarrow$ 
          & 1.71 & 1.93 & 3.45 & 1.66 & 2.58 & 2.28 & 2.32 & 3.14  & 2.38 \\
          & {\bf Comp. Ratio} [$<$ 1cm \%] $\uparrow$ 
          & 52.57& 43.56& 45.06& 48.16& 48.93& 53.59& 51.07 & 39.36& 47.79 \\
          & {\bf Comp. Ratio} [$<$ 5cm \%] $\uparrow$ 
          & 93.72& 92.95& 85.30& 94.56& 91.09& 89.99& 89.32 & 84.60& 90.19 \\
    \midrule
      \multirow{4}{*}{\makecell{\textbf{\ours{}}}} & {\bf Acc.} [cm] $\downarrow$ 
          & 2.18 & 3.46 & 2.01 & 2.37 & 2.27 & 1.75 & 1.90 & 1.93 & 2.23 \\
          & {\bf Comp. } [cm] $\downarrow$ 
          & 1.13 & 1.54 & 1.58 & 1.15 & 1.77 & 1.03 & 1.42 & 1.94 & \textbf{1.44} \\
          & {\bf Comp. Ratio} [$<$ 1cm \%] $\uparrow$ 
          & 74.09& 68.51& 66.81& 67.00& 65.24& 77.98& 68.62 & 65.56& \textbf{69.23} \\
          & {\bf Comp. Ratio} [$<$ 5cm \%] $\uparrow$ 
          & 96.68& 95.02& 92.98& 96.53& 92.94& 96.97& 94.21 & 91.03 & \textbf{94.55} \\
      \bottomrule
    \end{tabular}%
    \caption{Object-level reconstruction results on 8 indoor Replica scenes.  $\ast$ represents the baselines we re-trained with ground-truth pose.}
    \label{tab:replica_obj_3D}
\end{table}

We generated a new / different sequence for each scene in Replica dataset. We performed 2D novel view synthesis and compared it to the ground-truth views from the generated sequence. We compared baselines in depth L1 error, PSNR, SSIM, and LPIPS in Tab.~\ref{tab:replica_eval_2D}, and the 2D renderings for 3 selected scenes are shown in Fig.~\ref{fig:2D_novel_view}. 

\begin{table}[!th]
  \centering
  \footnotesize
  \setlength{\tabcolsep}{0.36em}
    \begin{tabular}{clccccccccc}
      \toprule
         & & \tt{room-0} & \tt{room-1} & \tt{room-2}  & \tt{office-0} &  \tt{office-1} & \tt{office-2} & \tt{office-3} & \tt{office-4} & Avg. \\
        \midrule
      \multirow{4}{*}{\makecell{\textbf{NICE-SLAM}}}    
          & {\bf Depth L1.} [cm] $\downarrow$
          & 1.99 & 1.57 & 2.72 & 12.50 & 7.37 & 3.03 & 2.39 & 2.18 & 4.22 \\
          & {\bf PSNR.} $\uparrow$
          & 24.11 & 23.43 & 23.48 & 23.91 & 22.69  & 23.78 & 23.78 & 26.00 & 23.90 \\
          & {\bf SSIM} $\uparrow$
          & 0.73 & 0.74 & 0.82 & 0.83 & 0.82 & 0.83 & 0.84 & 0.85 & 0.81 \\
          & {\bf LPIPS} $\downarrow$
          & 0.11 & 0.09 & 0.09 & 0.15 & 0.28 & 0.11 & 0.10 & 0.09 & 0.13 \\
    \midrule
      \multirow{4}{*}{\makecell{\textbf{NICE-SLAM$^\ast$}}}    
          & {\bf Depth L1.} [cm] $\downarrow$
          & 1.87 & 1.63 & 2.94 & 13.43 & 7.63 & 2.83 & 2.62 & 1.97 & 4.36 \\
          & {\bf PSNR.} $\uparrow$
          & 24.03 & 23.61 & 23.54 & 23.59 & 23.19 & 22.22 & 23.32 & 26.20 & 23.71 \\
          & {\bf SSIM} $\uparrow$
          & 0.73 & 0.75 & 0.82 & 0.83 & 0.84 & 0.85 & 0.84 & 0.86 & 0.82 \\
          & {\bf LPIPS} $\downarrow$
          & 0.11 & 0.09 & 0.09 & 0.16 & 0.26 & 0.10 & 0.10 & 0.09 & 0.13 \\
      \midrule
      \multirow{4}{*}{\makecell{\textbf{iMAP$^\ast$}}}
          & {\bf Depth L1.} [cm] $\downarrow$
          & 1.23 & 2.16     & 2.53 & 13.29 & 5.14 & 2.31 & 1.77 & 1.44 & 3.73\\
          & {\bf PSNR.} $\uparrow$
          & 25.83 & 25.51   & 25.22 & 24.17 & 23.94  & 24.02 & 25.45 & 29.13 & \textbf{25.41} \\
          & {\bf SSIM} $\uparrow$
          & 0.77 & 0.79     & 0.86 & 0.83 & 0.87 & 0.88 & 0.89 & 0.90 & \textbf{0.85} \\
          & {\bf LPIPS} $\downarrow$
          & 0.09 & 0.07     & 0.07 & 0.17 & 0.22 & 0.08 & 0.07 & 0.07 & \textbf{0.11} \\
      \midrule
      \multirow{4}{*}{\makecell{\textbf{\ours{}}}}
          & {\bf Depth L1.} [cm] $\downarrow$
          & 1.68 & 1.57     & 2.37  & 7.73  & 6.60  & 2.50  & 2.30 & 1.85 & \textbf{3.33} \\
          & {\bf PSNR.} $\uparrow$
          & 25.23 & 25.27   & 24.31 & 23.78 & 23.59 & 23.10 & 23.83 & 27.91 & 24.63 \\
          & {\bf SSIM} $\uparrow$
          & 0.77 & 0.78     & 0.85  & 0.84  & 0.88  & 0.88  & 0.88 & 0.89 & \textbf{0.85} \\
          & {\bf LPIPS} $\downarrow$
          & 0.09 & 0.07     & 0.08  & 0.16  & 0.23  & 0.07  & 0.08 & 0.07 & \textbf{0.11} \\
      \bottomrule
    \end{tabular}%
    \caption{2D novel view synthesis rendering results on the Replica dataset.}
    \label{tab:replica_eval_2D}
\end{table}

\newpage
\begin{figure*}[ht!]
  \centering
  \small
  \setlength{\tabcolsep}{0.1em}
  \begin{tabular}{>{\centering}m{0.06\textwidth} >{\centering\arraybackslash}m{0.93\textwidth}}
    GT &
    \includegraphics[width=0.33\linewidth]{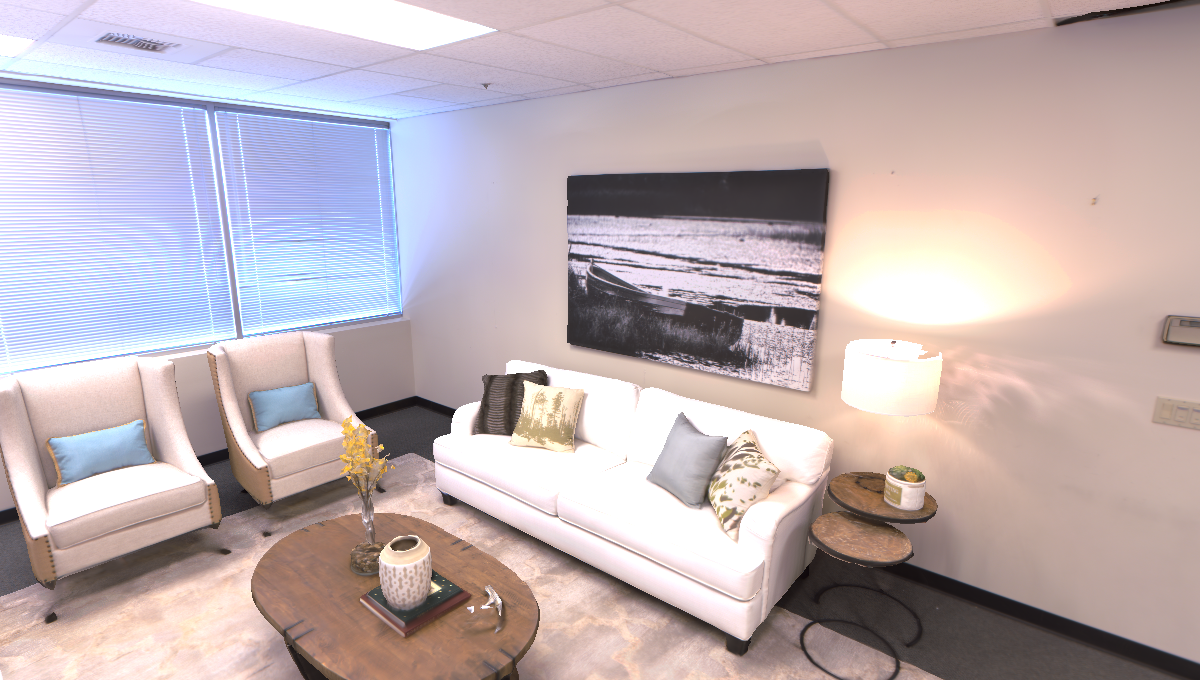}\hfill
    \includegraphics[width=0.33\linewidth]{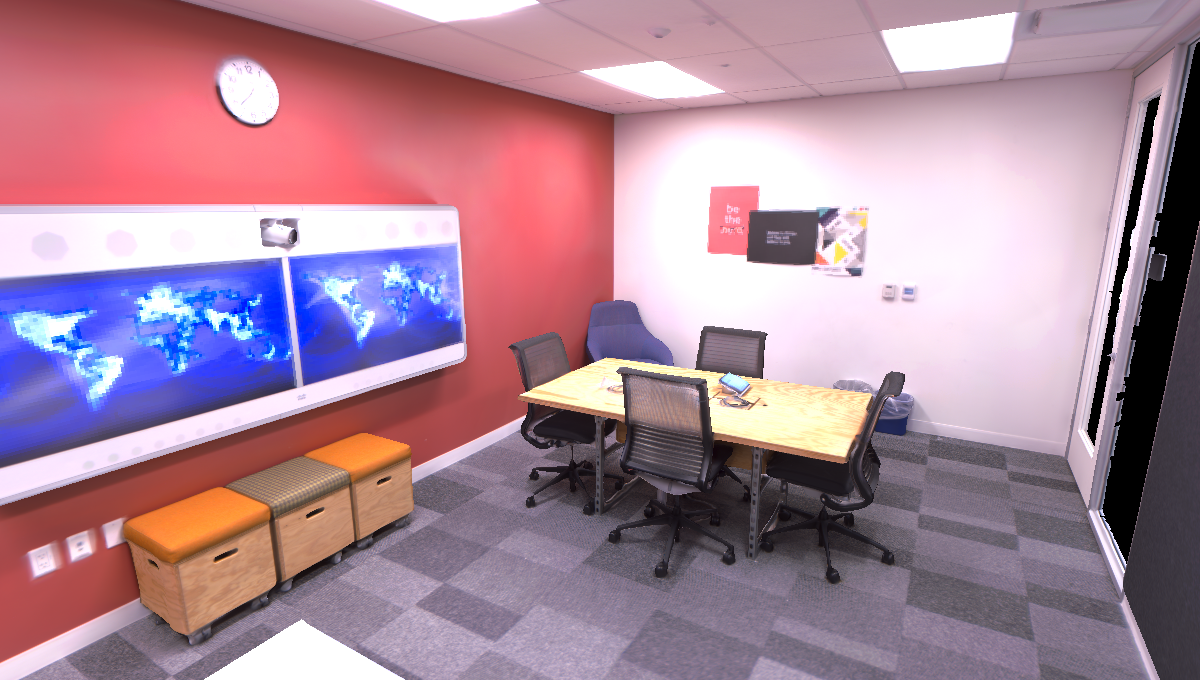}\hfill
    \includegraphics[width=0.33\linewidth]{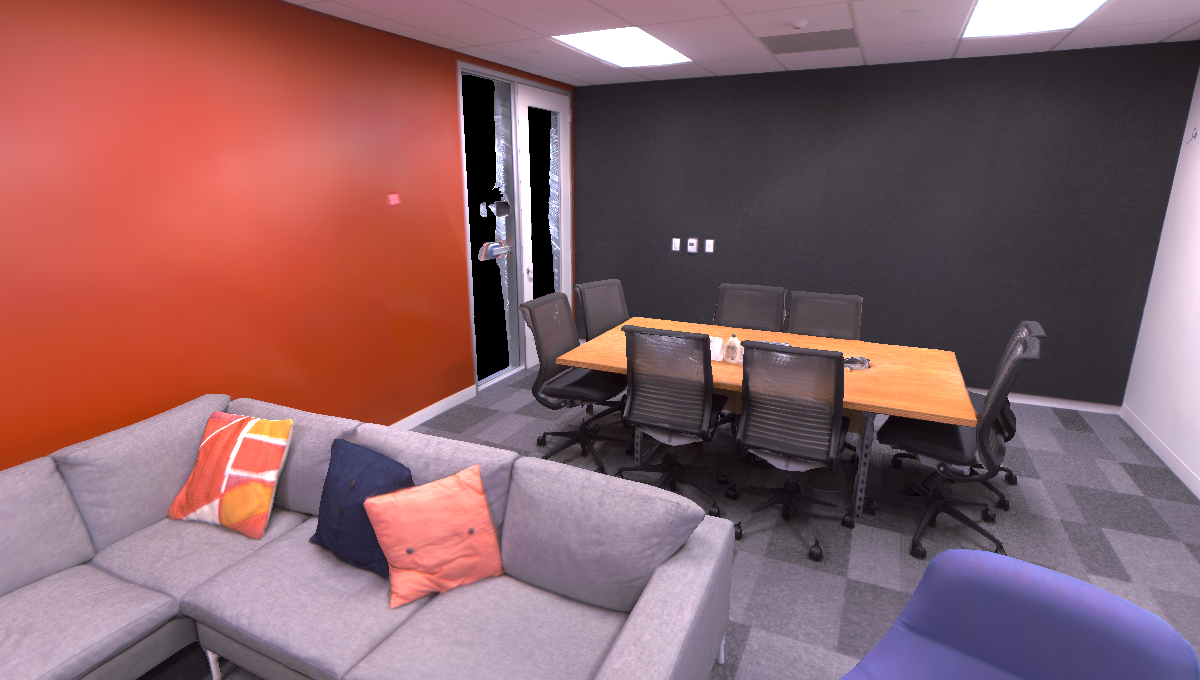}
    \\
    NICE \\ SLAM$^\ast$ &
    \includegraphics[width=0.33\linewidth]{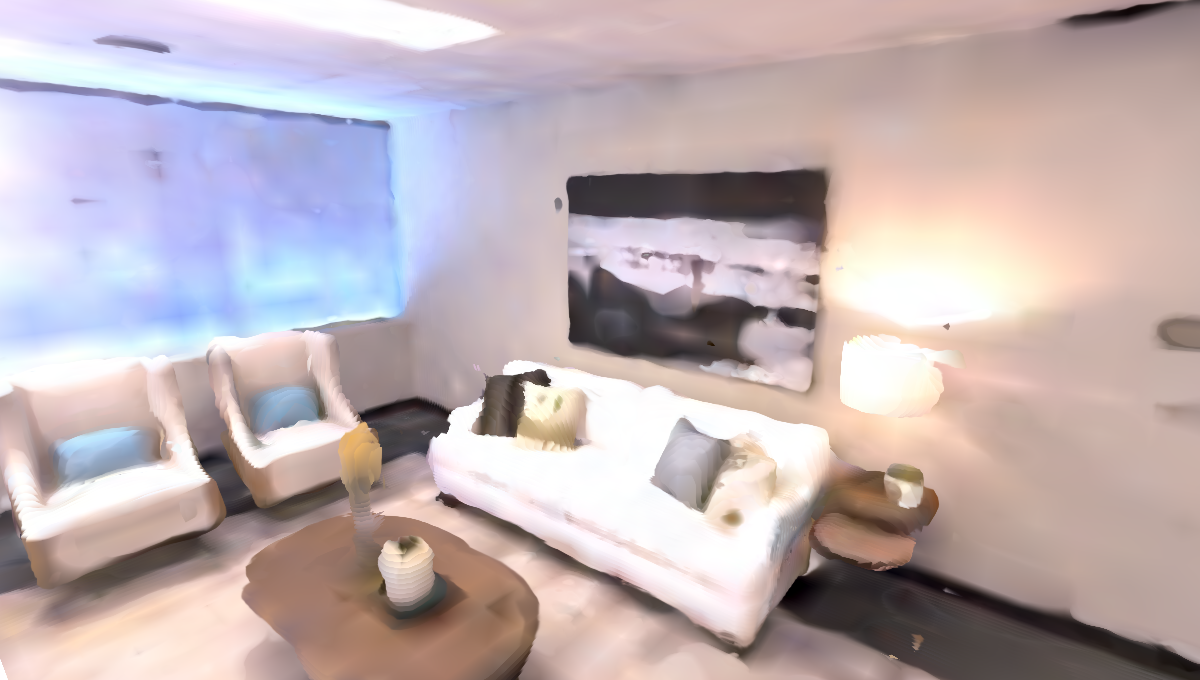}\hfill
    \includegraphics[width=0.33\linewidth]{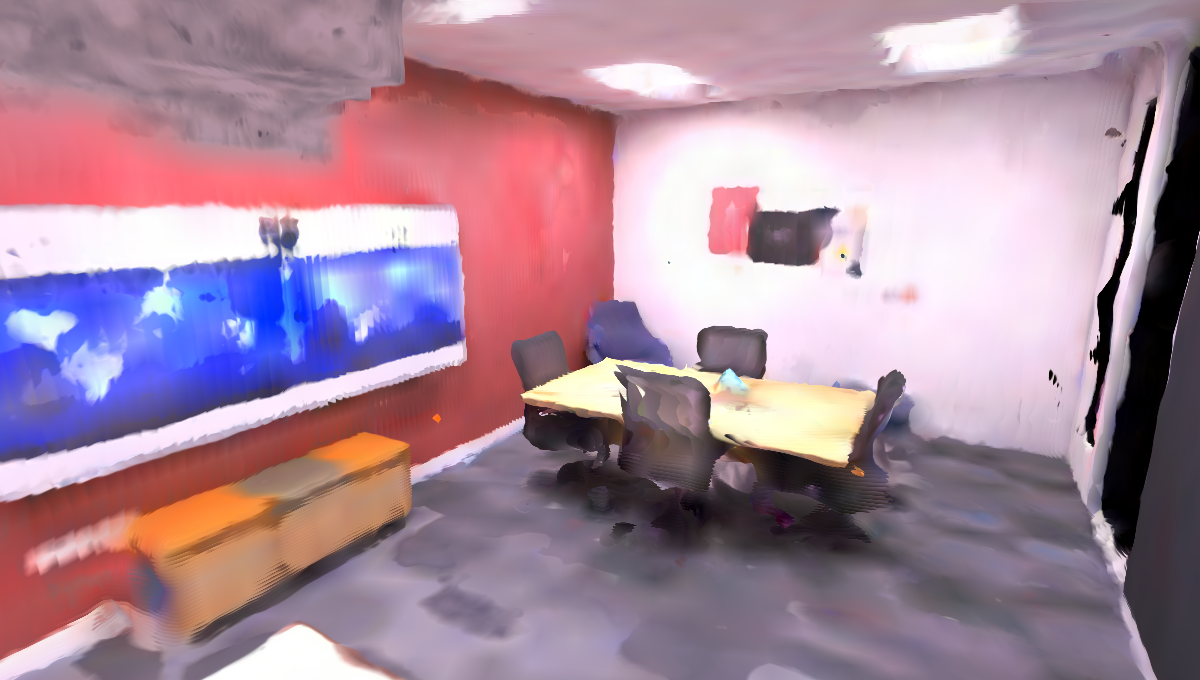}\hfill
    \includegraphics[width=0.33\linewidth]{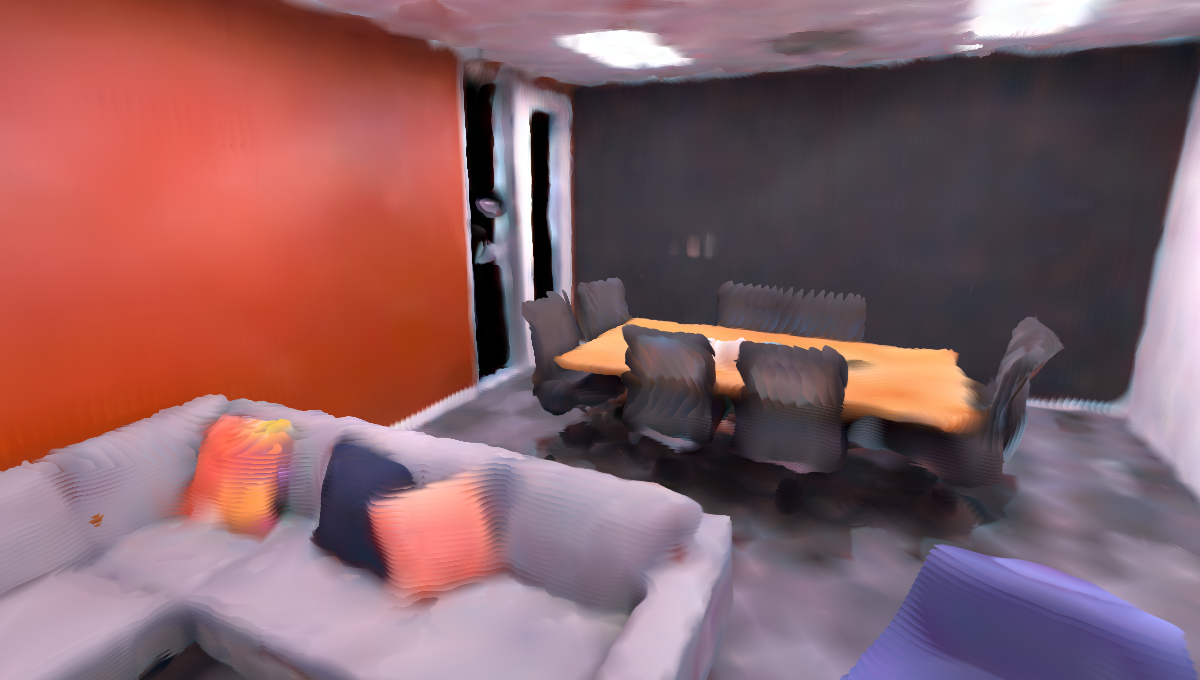}
    \\
    iMAP$^\ast$ &
    \includegraphics[width=0.33\linewidth]{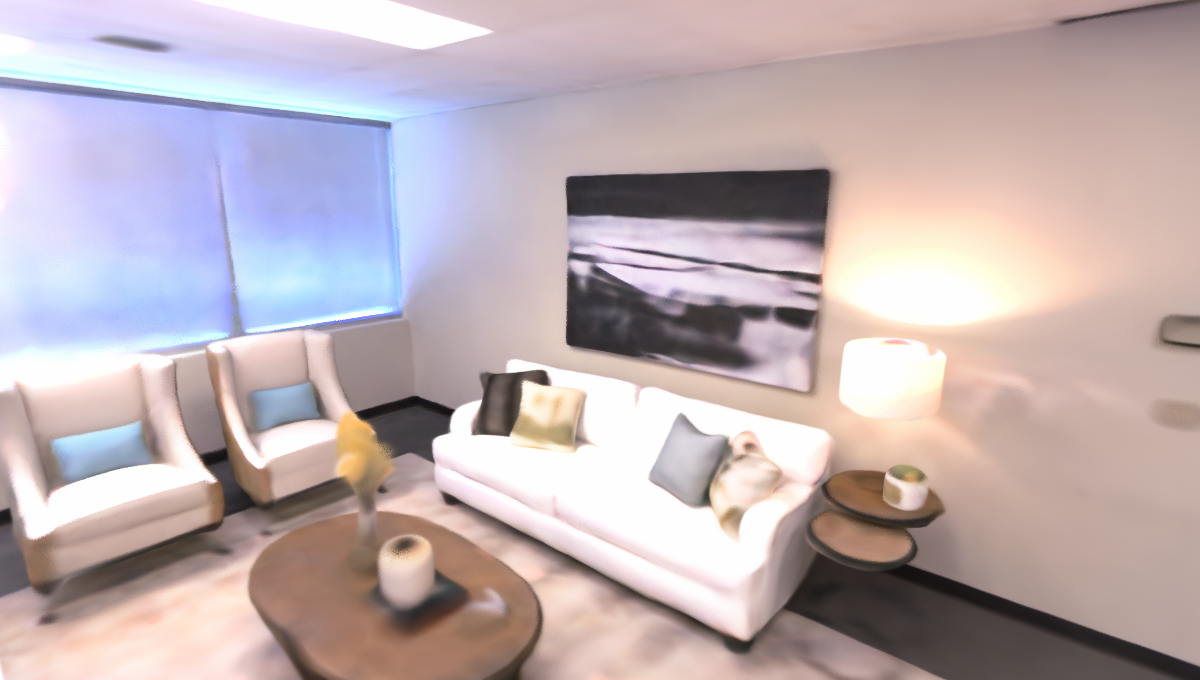}\hfill
    \includegraphics[width=0.33\linewidth]{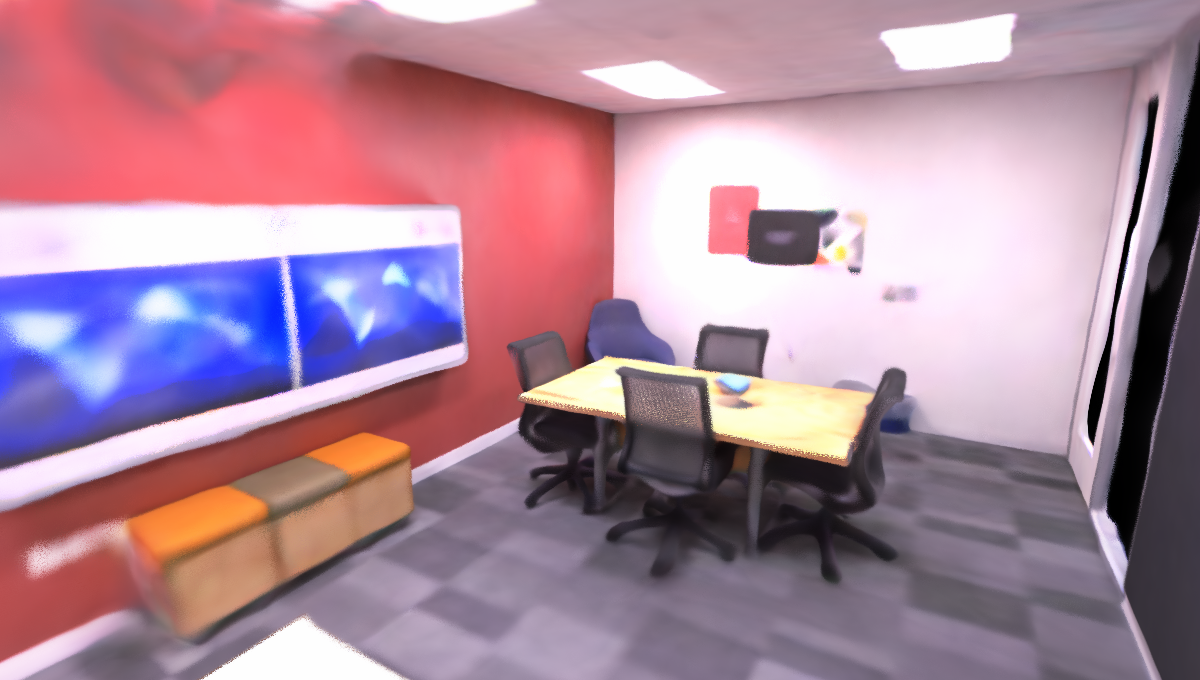}\hfill
    \includegraphics[width=0.33\linewidth]{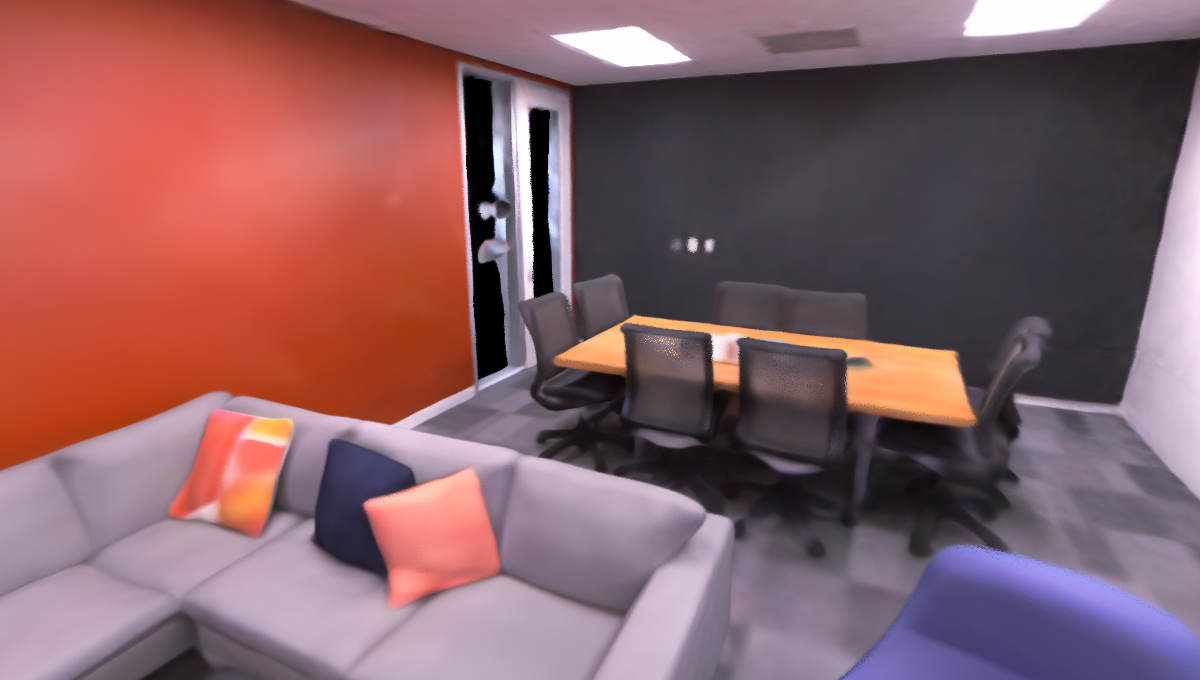}
    \\
     \makecell{\ours{}} &
    \includegraphics[width=0.33\linewidth]{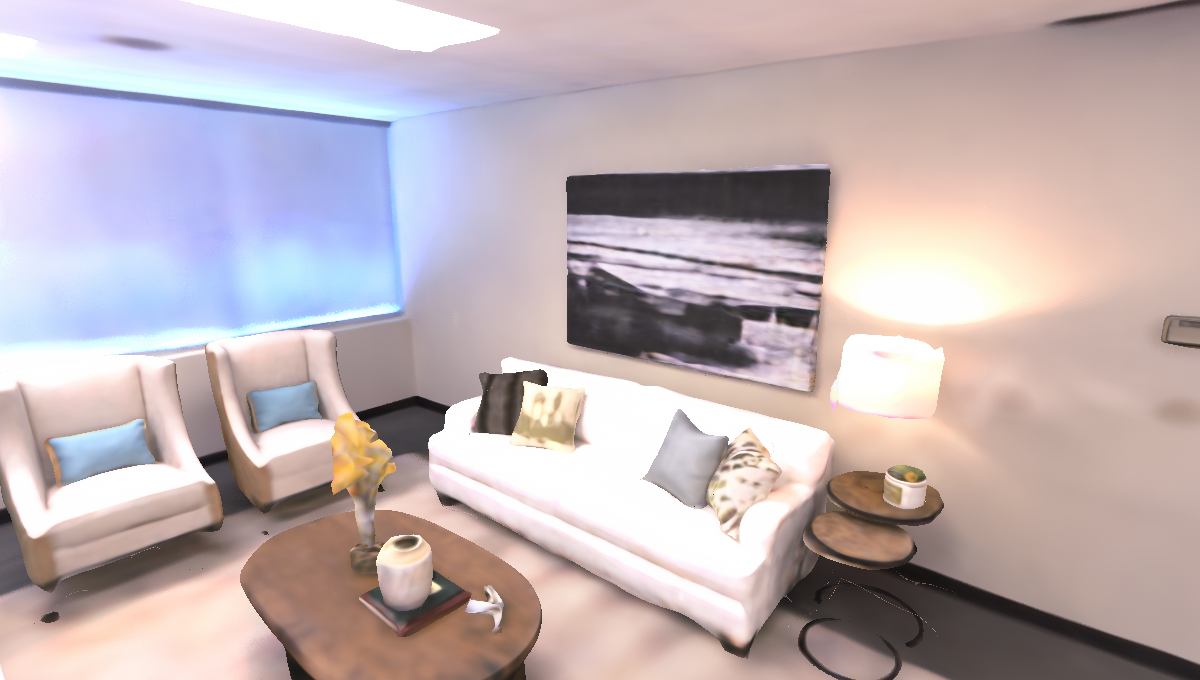}\hfill
    \includegraphics[width=0.33\linewidth]{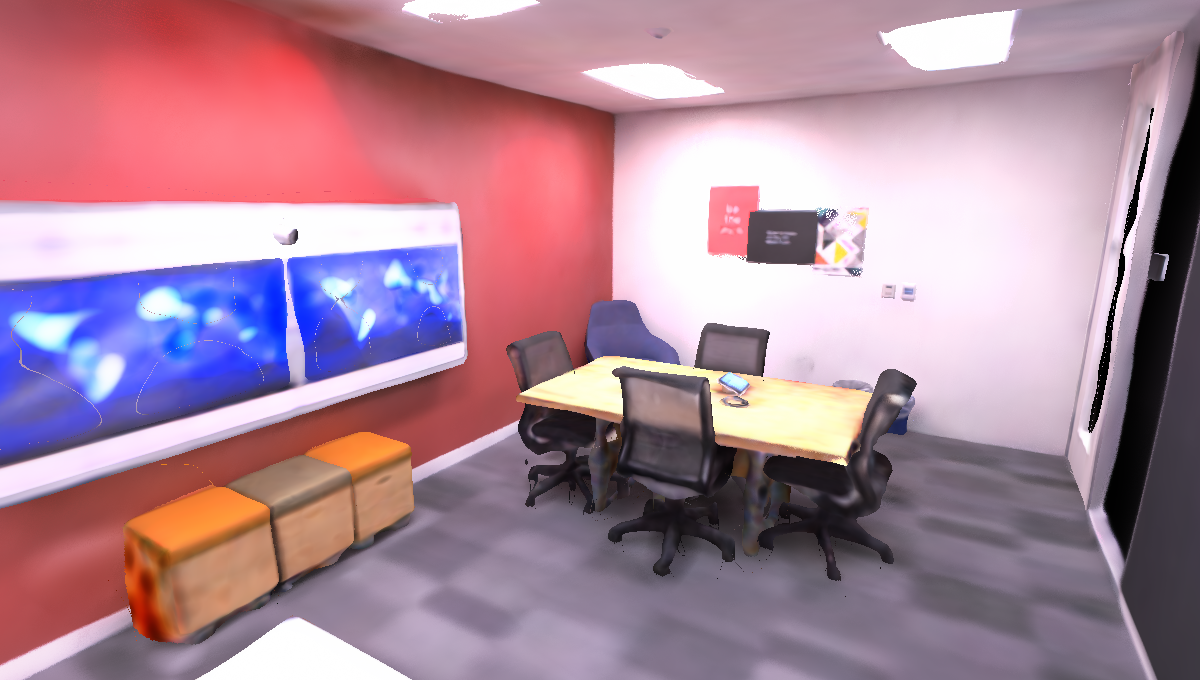}\hfill
    \includegraphics[width=0.33\linewidth]{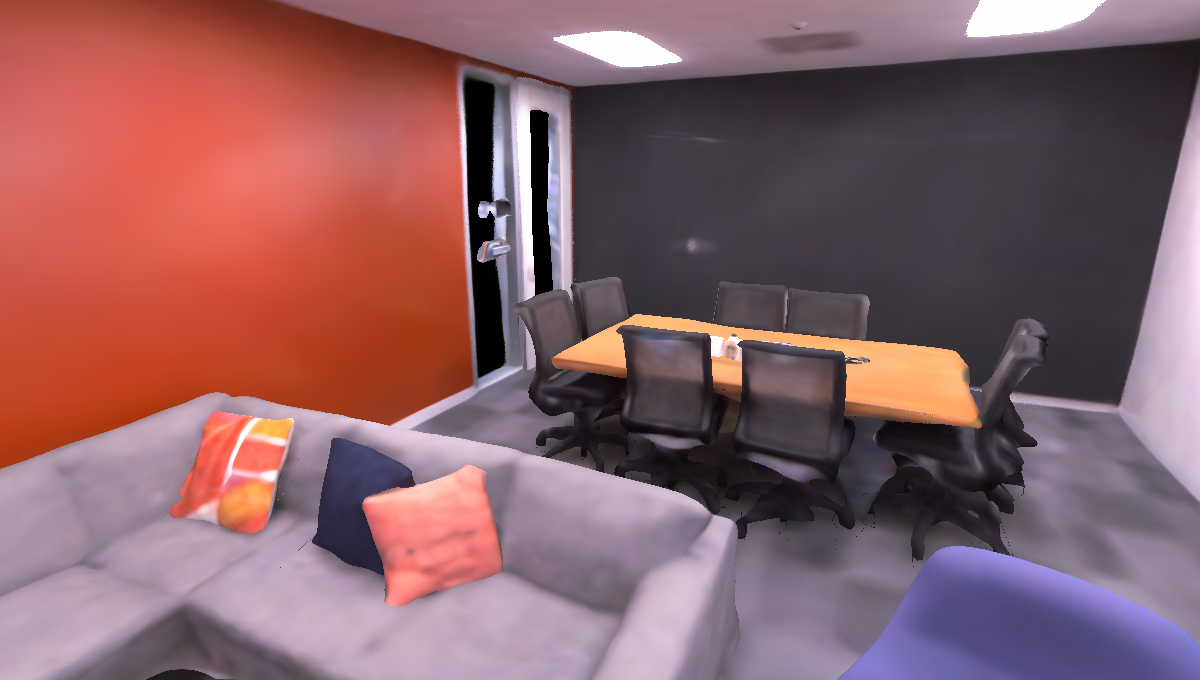}
  \end{tabular}
  \caption{Visualisation of 2D novel view synthesis rendering results on the Replica dataset, better when zoomed. }
  \label{fig:2D_novel_view}
\end{figure*}

\newpage
\section{Visualisation of Object-level Hole-filling}
\begin{figure*}[ht!]
  \centering
  \small
  \setlength{\tabcolsep}{0.2em}
  \begin{tabular}{C{0.24\linewidth}C{0.24\linewidth}C{0.24\linewidth}C{0.24\linewidth}}
    GT &  TSDF-Fusion$^\ast$ & iMAP$^\ast$ & \ours{} \\
    \includegraphics[trim={13cm 4cm 13cm 4cm}, clip, width=\linewidth]{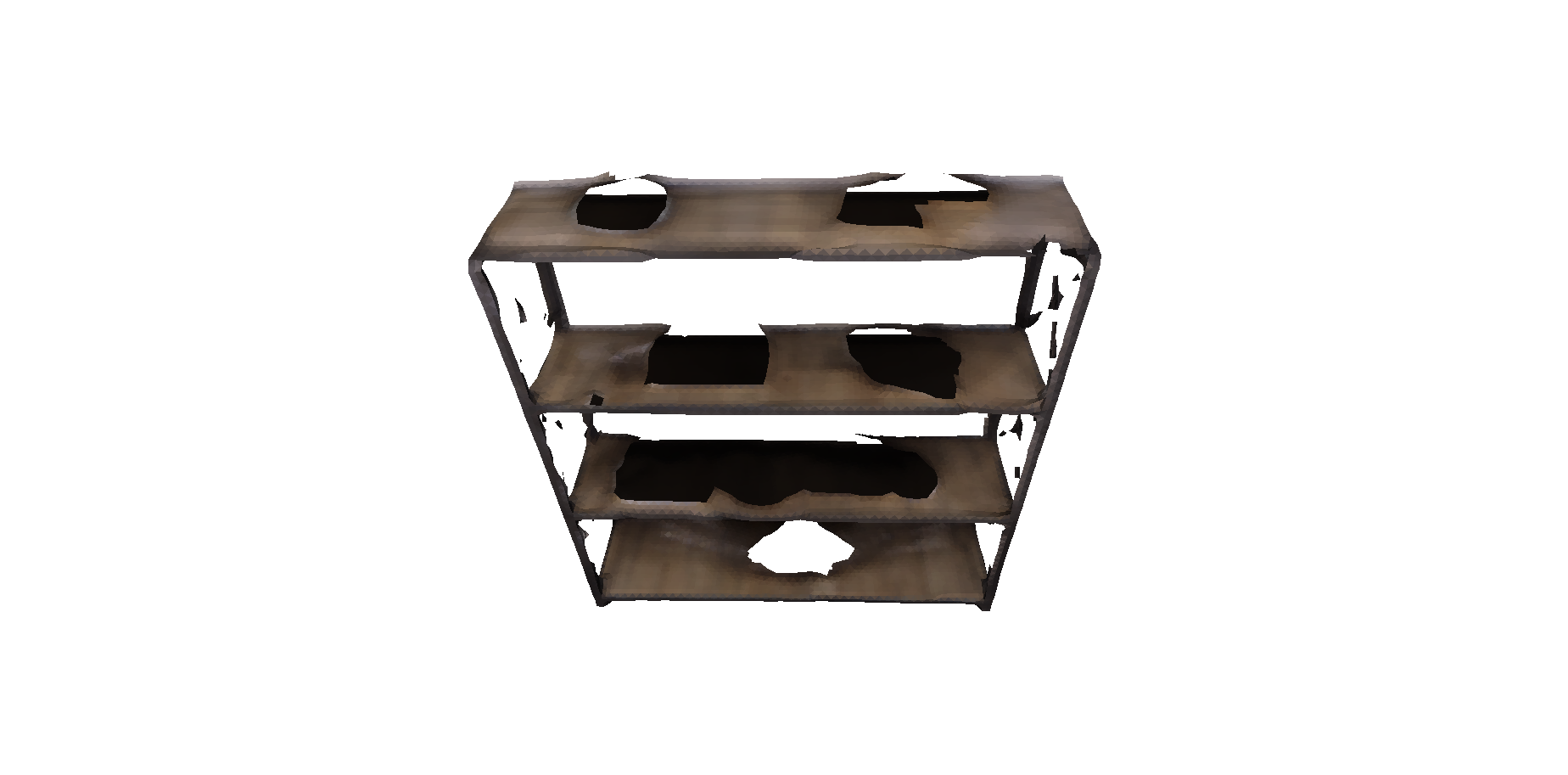} &
    \includegraphics[trim={13cm 4cm 13cm 4cm}, clip, width=\linewidth]{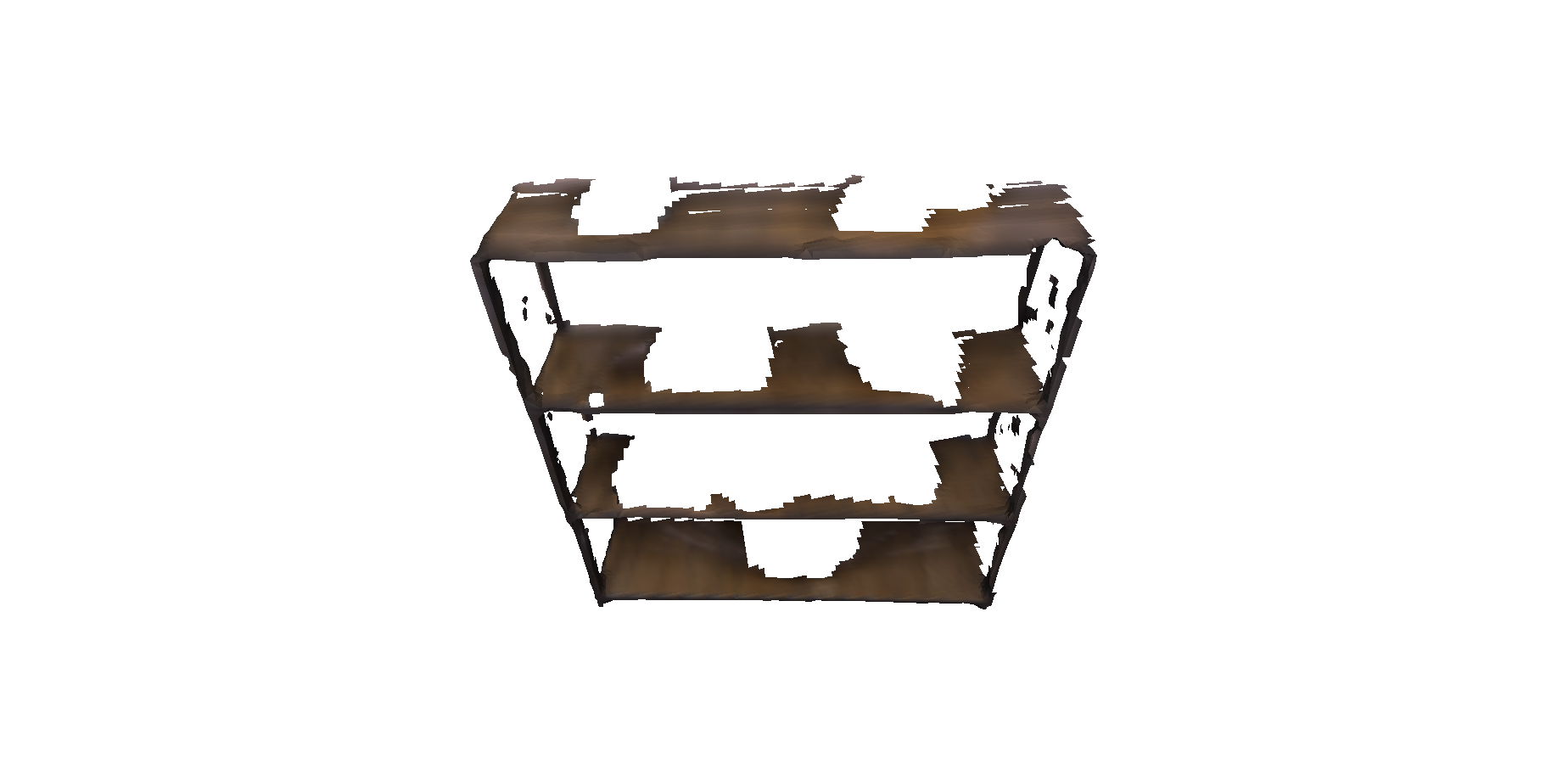} &
    \includegraphics[trim={13cm 4cm 13cm 4cm}, clip, width=\linewidth]{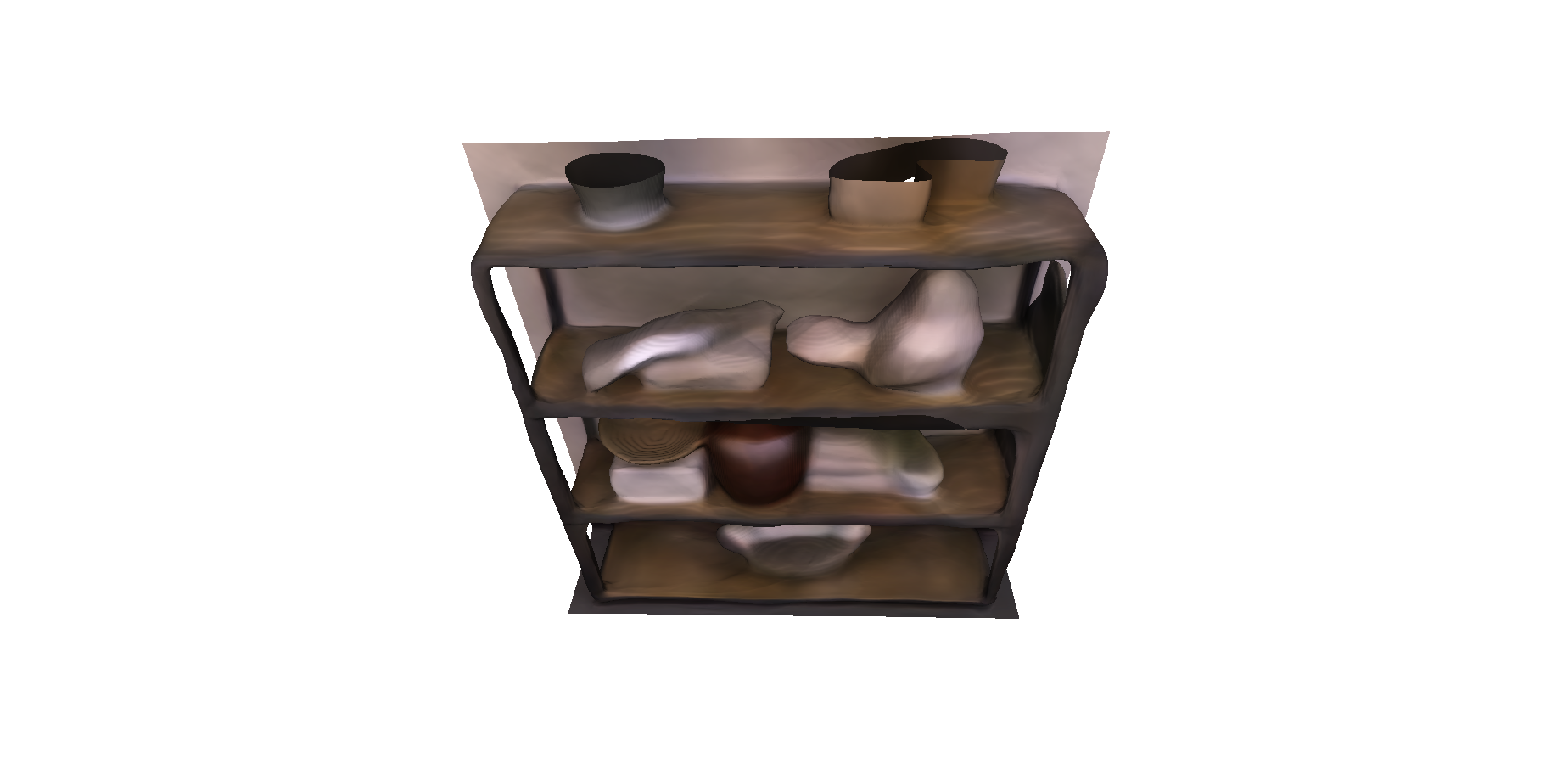} &
    \includegraphics[trim={13cm 4cm 13cm 4cm}, clip, width=\linewidth]{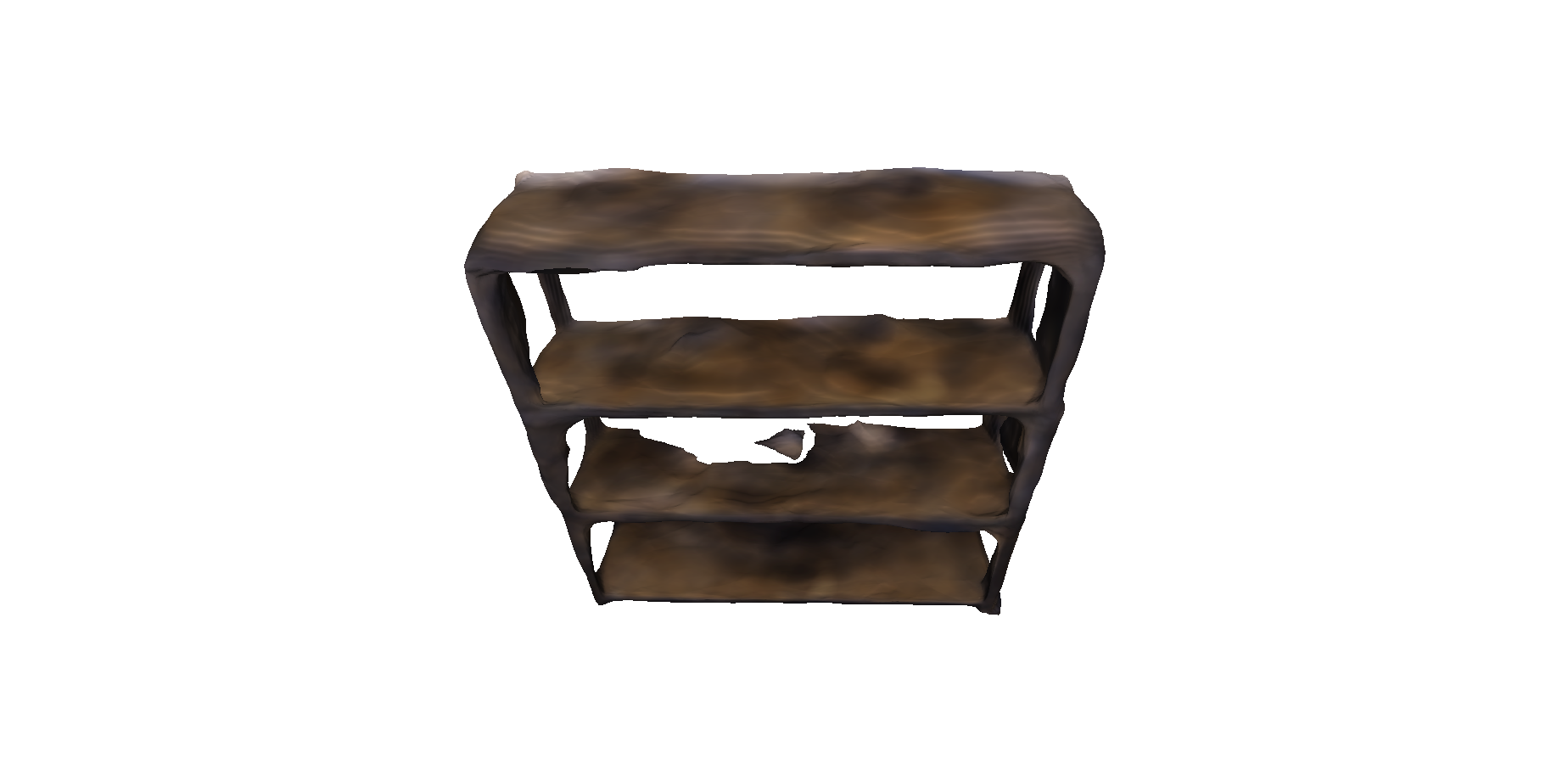}  \\
    \includegraphics[trim={6cm 3cm 6cm 3cm}, clip, width=\linewidth]{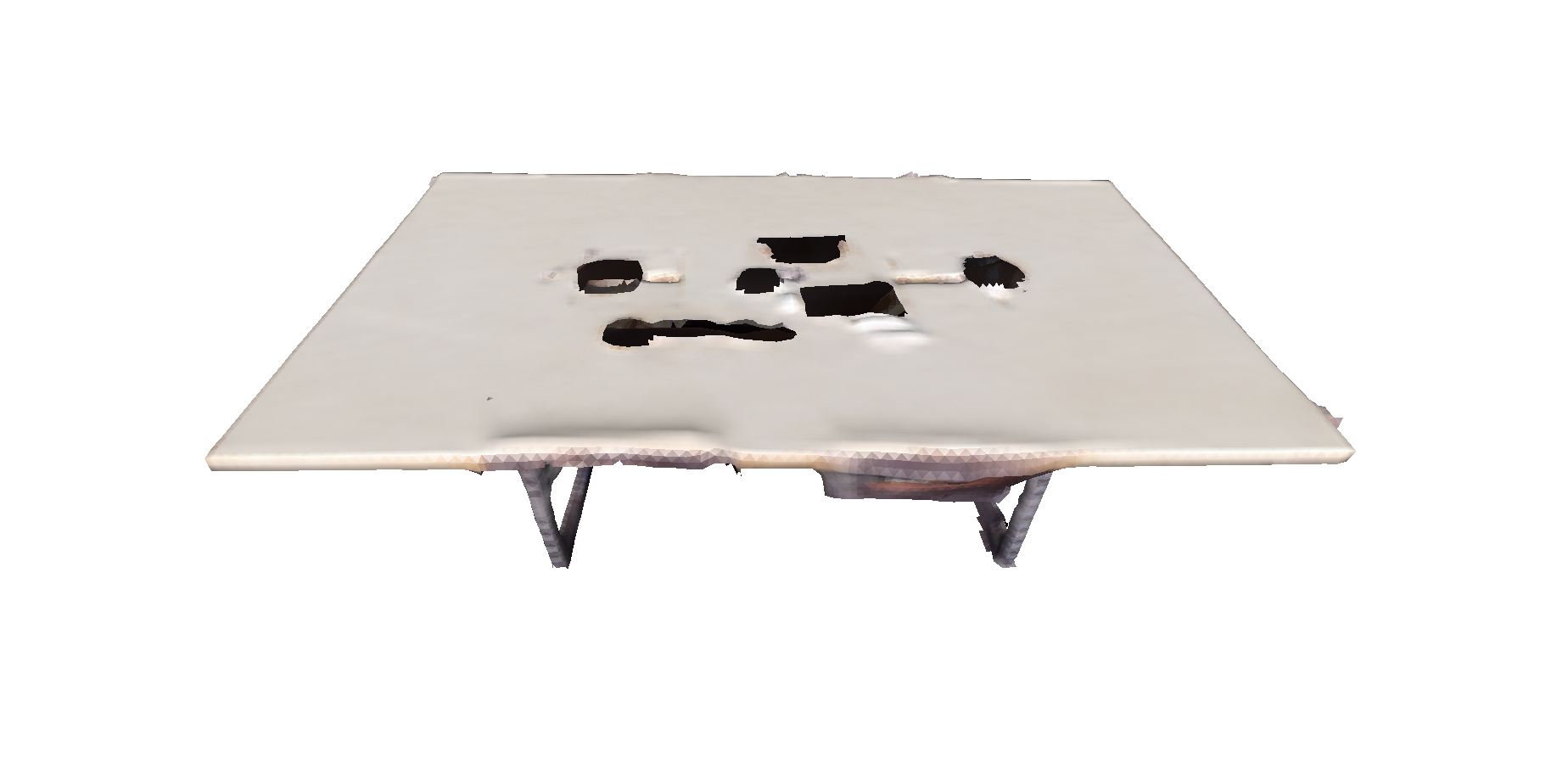} &
    \includegraphics[trim={6cm 3cm 6cm 3cm}, clip, width=\linewidth]{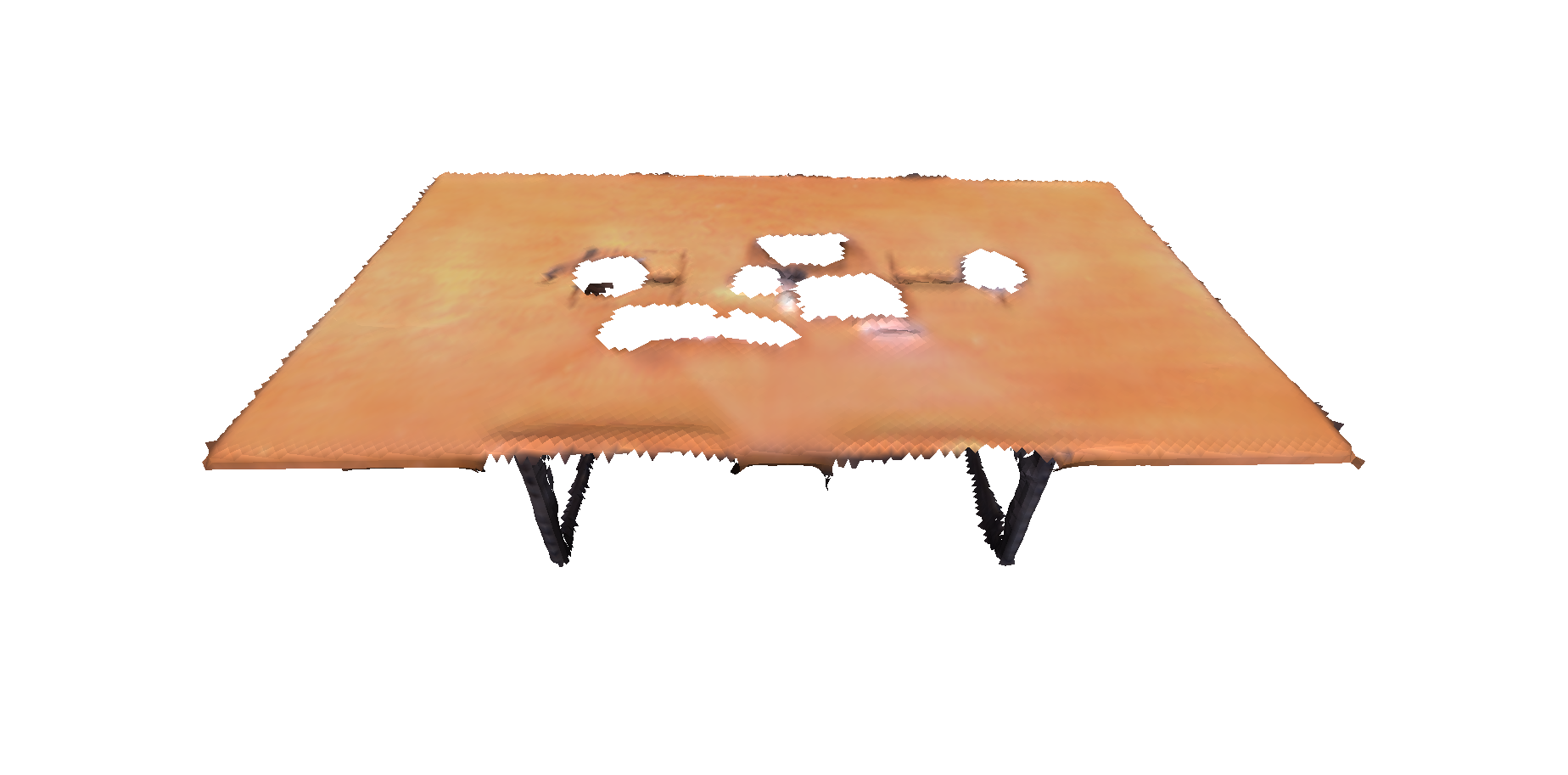} &
    \includegraphics[trim={6cm 3cm 6cm 3cm}, clip, width=\linewidth]{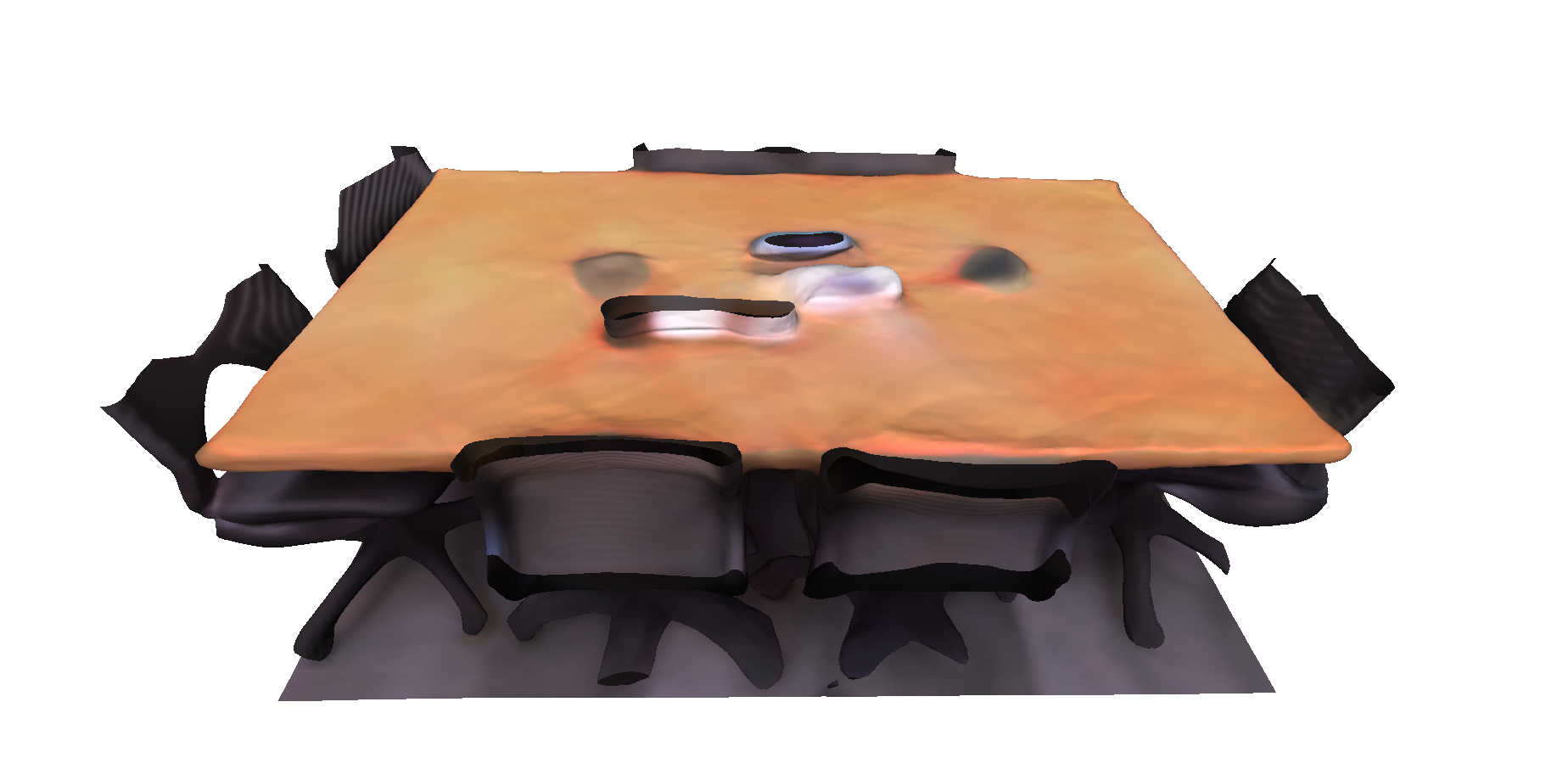} &
    \includegraphics[trim={6cm 3cm 6cm 3cm}, clip, width=\linewidth]{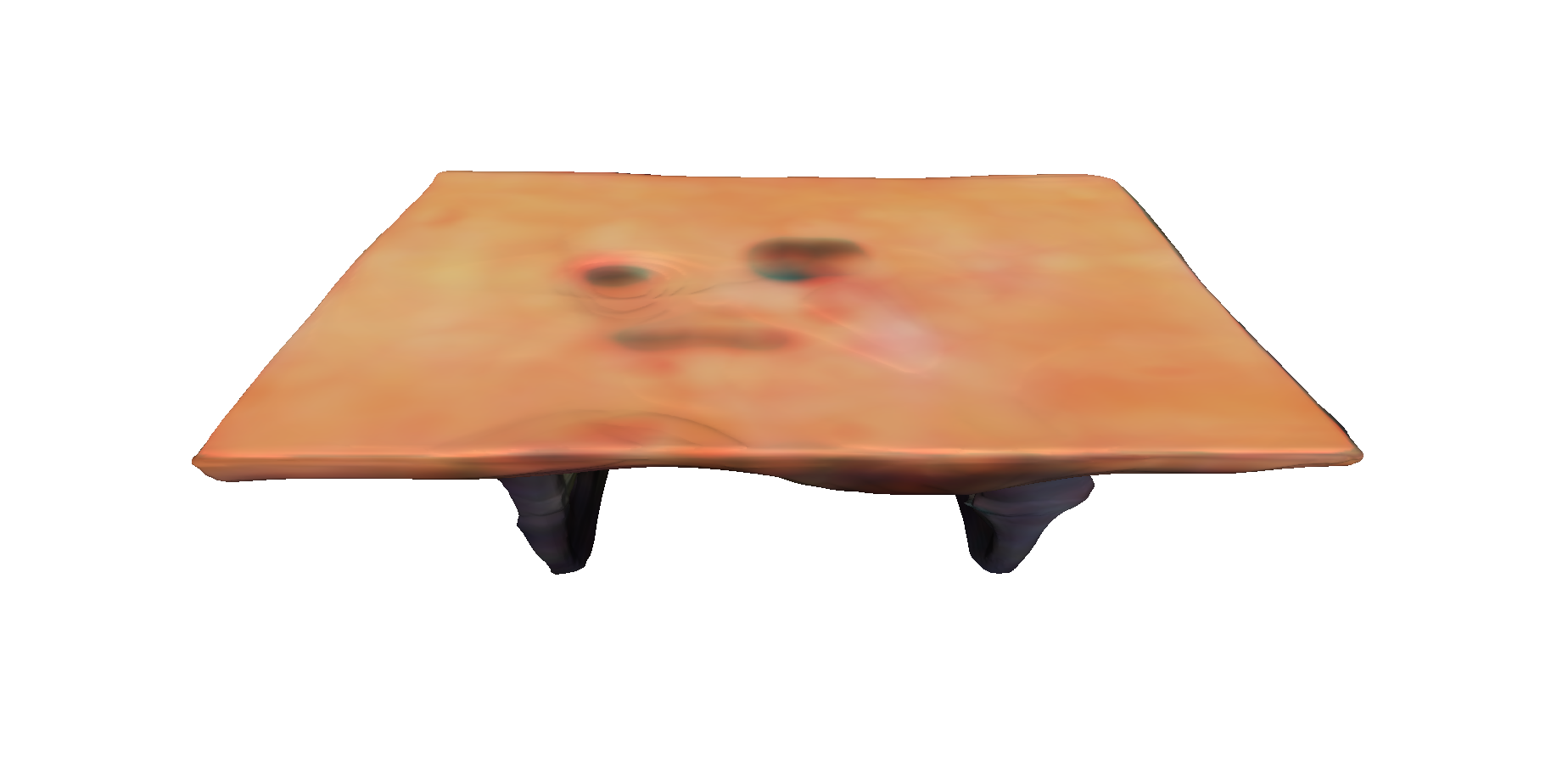}  \\
    \includegraphics[trim={14cm 2cm 14cm 2cm}, clip, width=\linewidth]{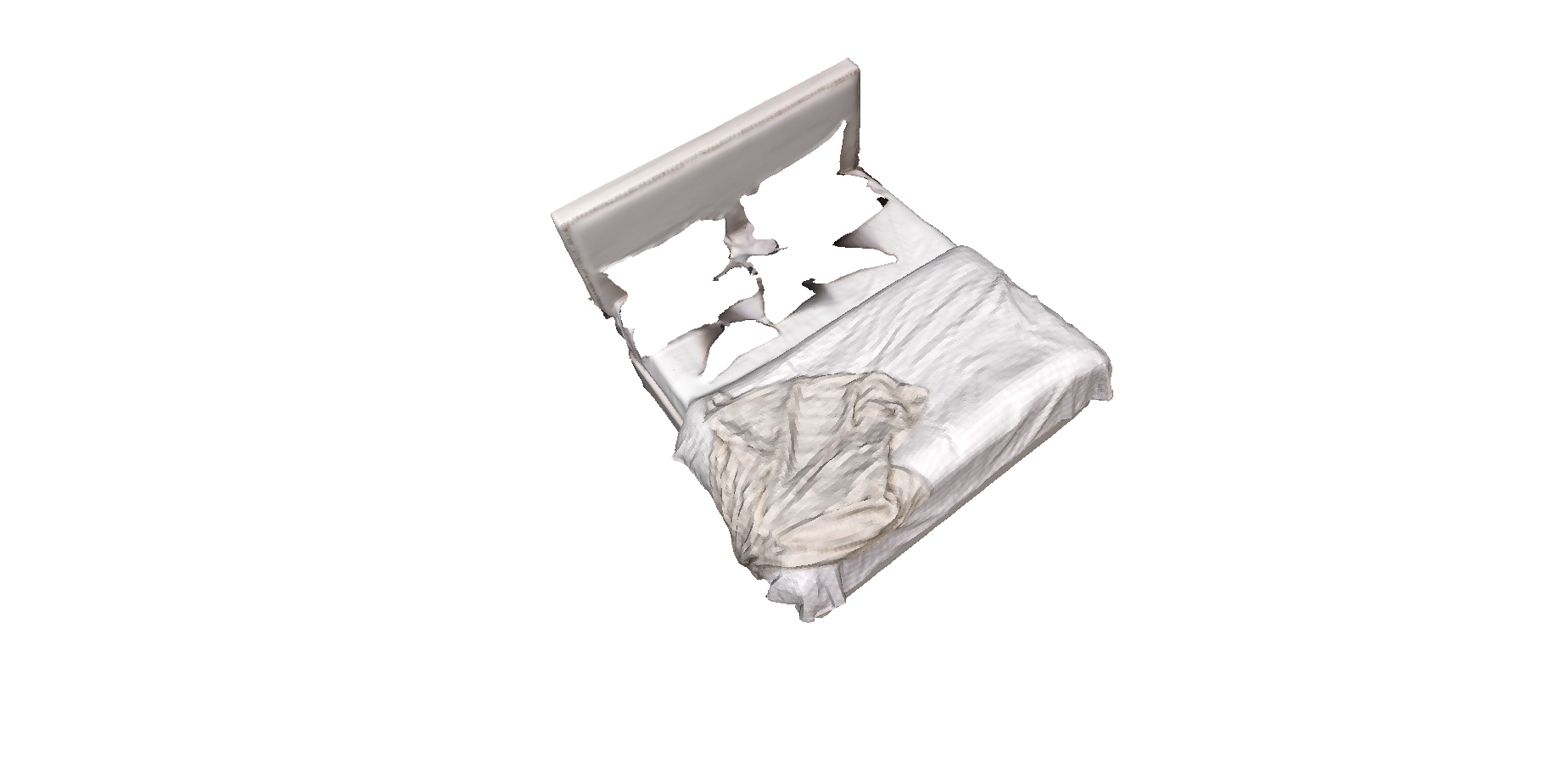} &
    \includegraphics[trim={14cm 2cm 14cm 2cm}, clip, width=\linewidth]{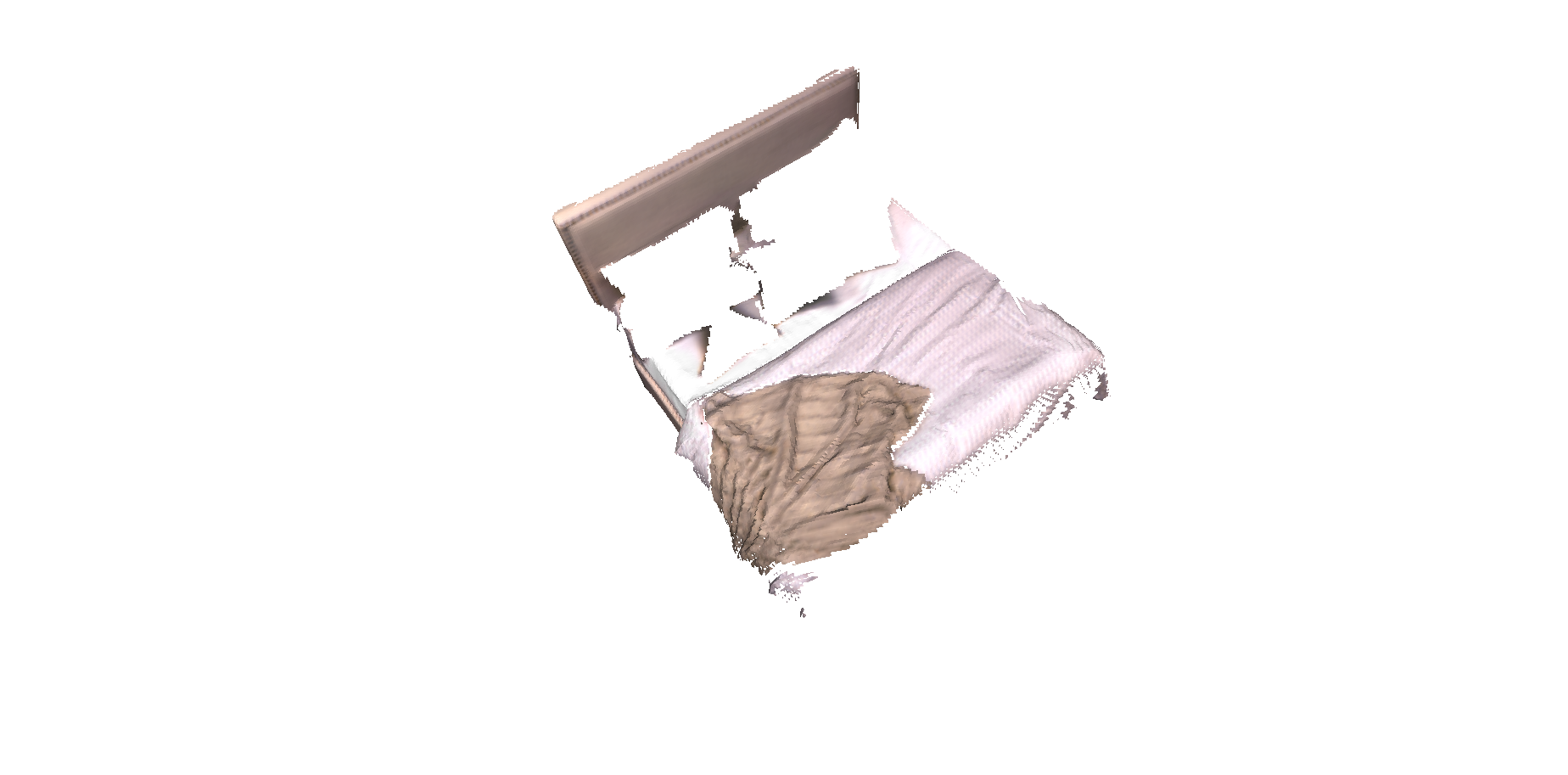} &
    \includegraphics[trim={14cm 2cm 14cm 2cm}, clip, width=\linewidth]{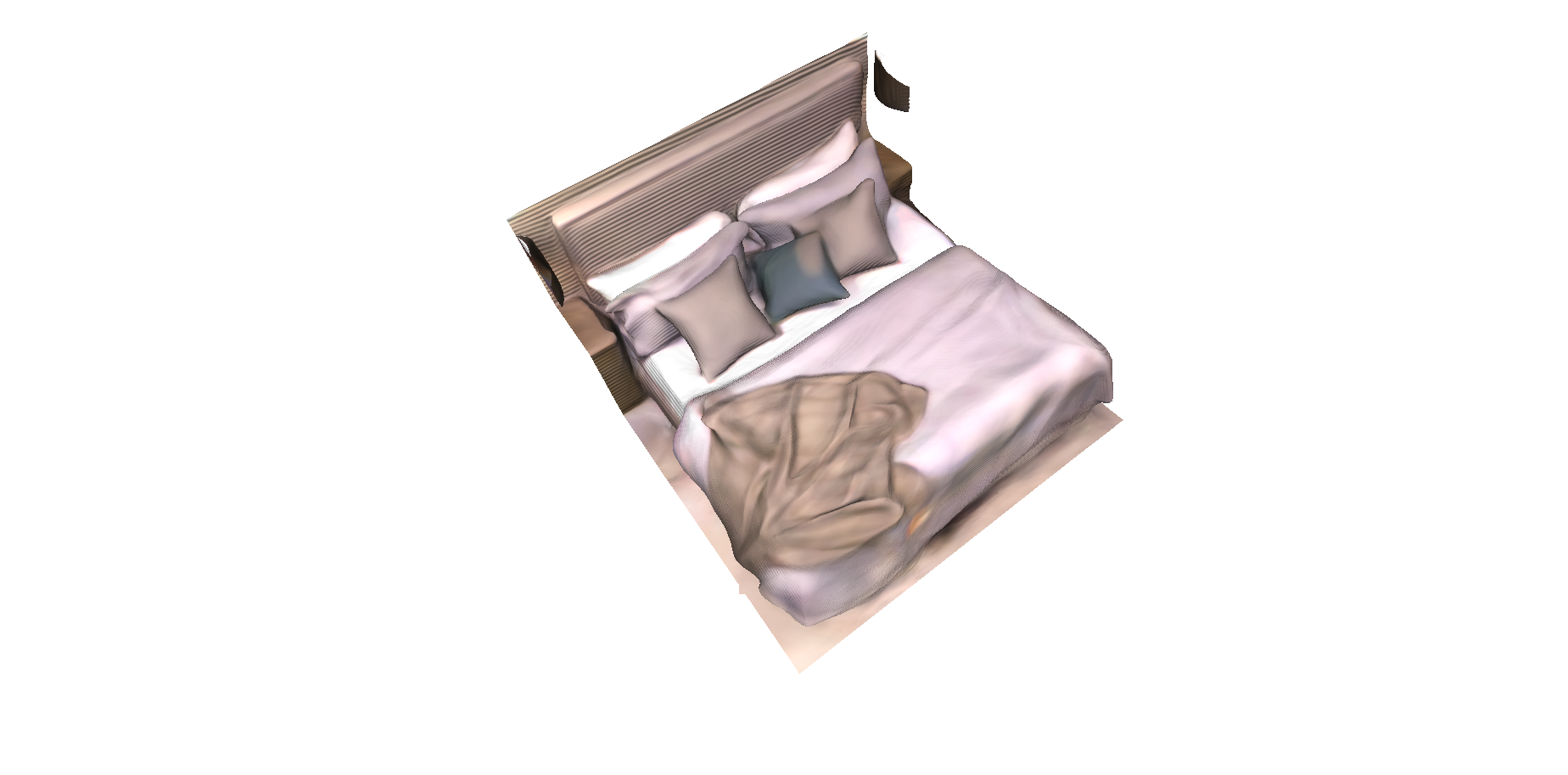} &
    \includegraphics[trim={14cm 2cm 14cm 2cm}, clip, width=\linewidth]{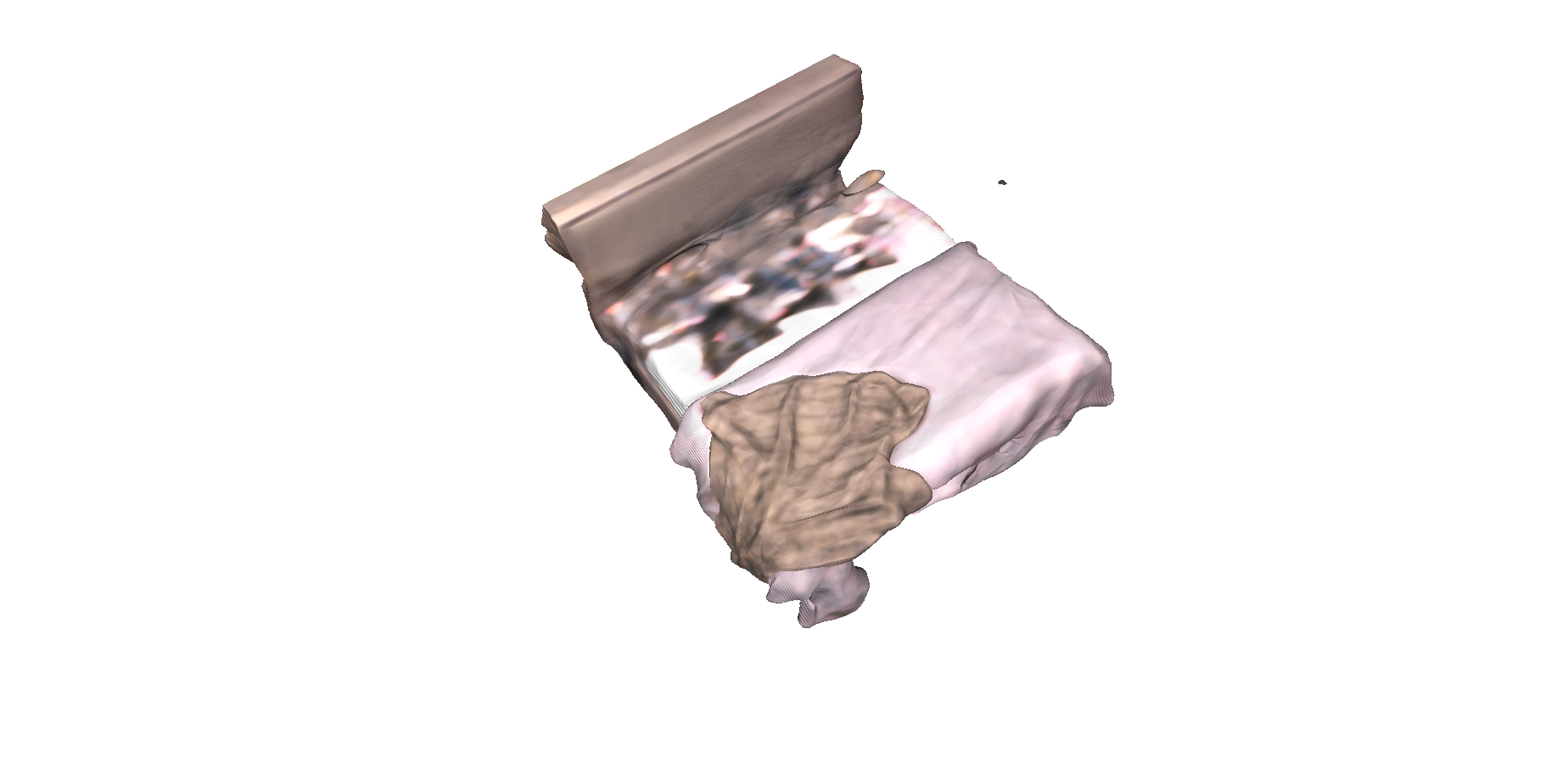}  \\
    \includegraphics[trim={0cm 0cm 0cm 2cm}, clip, width=\linewidth]{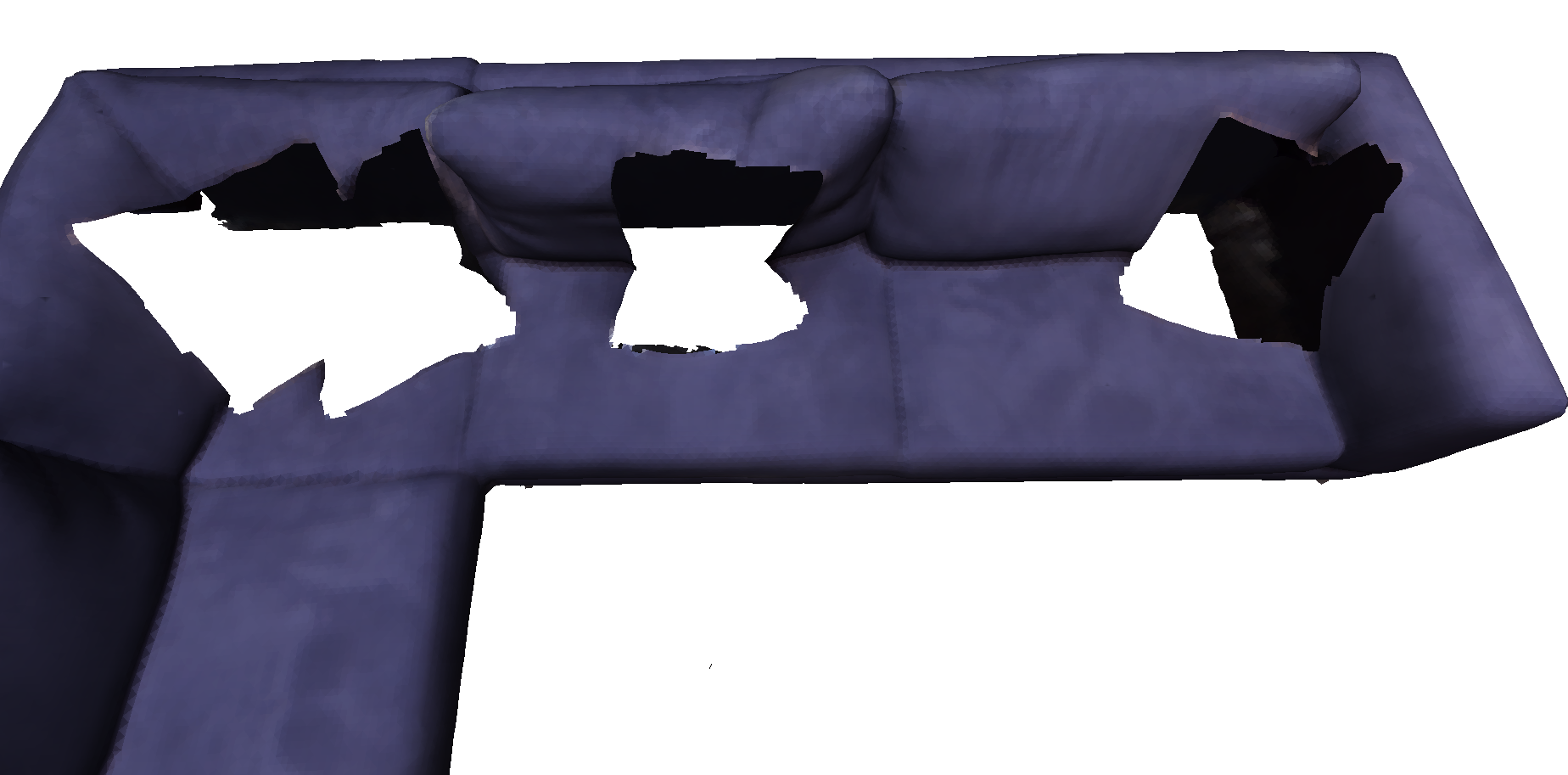} &
    \includegraphics[trim={0cm 0cm 0cm 2cm}, clip, width=\linewidth]{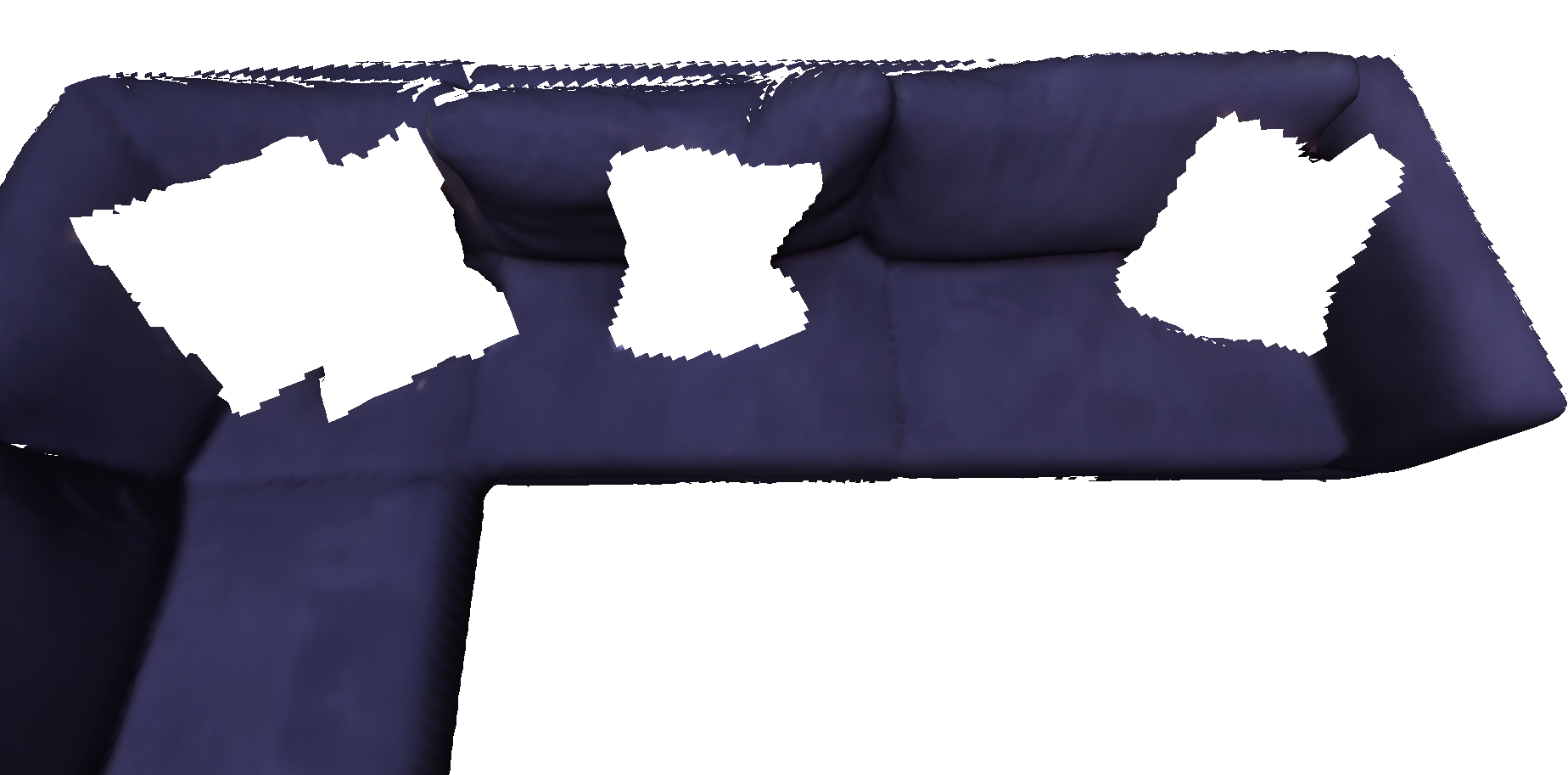} &
    \includegraphics[trim={0cm 0cm 0cm 2cm}, clip, width=\linewidth]{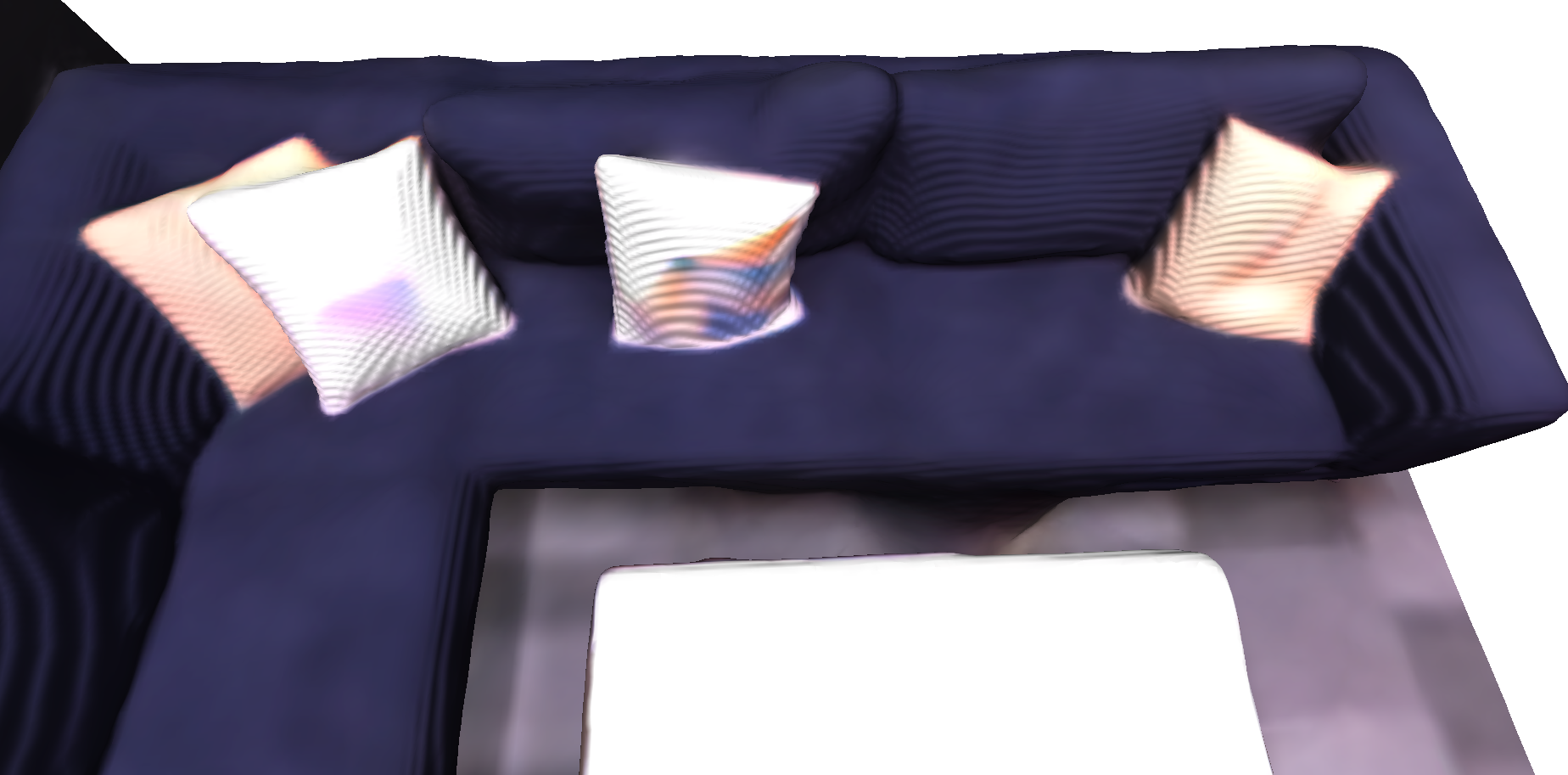} &
    \includegraphics[trim={0cm 0cm 0cm 2cm}, clip, width=\linewidth]{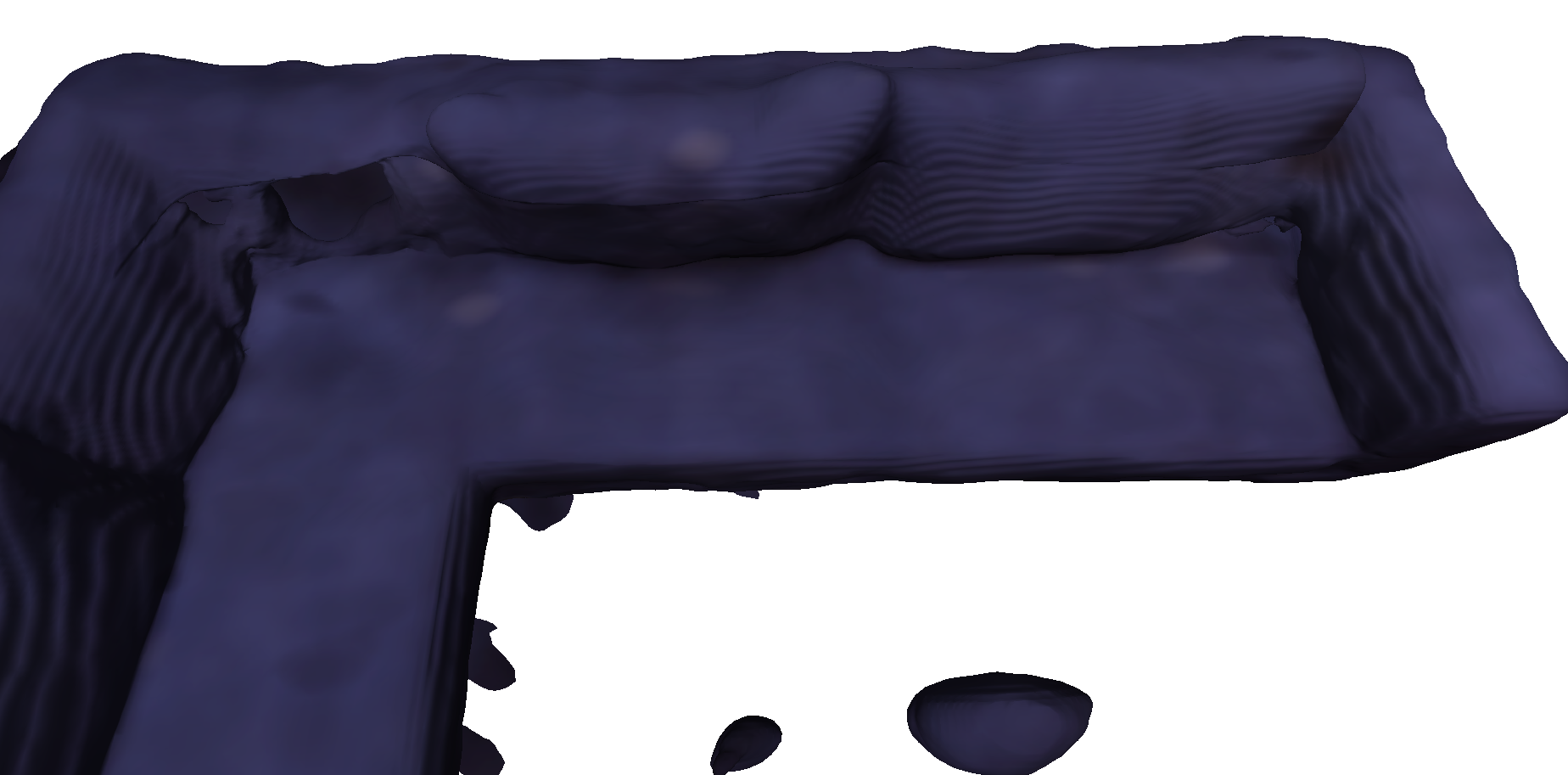}  \\
  \end{tabular}
  \caption{Visualisation of object-level hole-filling.}
  \label{fig:holefilling}
\end{figure*}

Compared to iMAP and NICE-SLAM, \ours{} shows significantly better hole-filling capability in unobserved regions with visual consistency, thanks to the disentangled object representation design. As shown in Fig.~\ref{fig:holefilling}, \ours{} is able to generate smooth and natural geometries without requiring any other priors.

\end{document}